\documentclass[journal]{IEEEtran}
\usepackage{graphicx}
\usepackage{amsmath}
\usepackage{amssymb}
\usepackage{makeidx}
\usepackage{subfig}
\usepackage{float}
\usepackage{tabularx}
\usepackage{makecell}
\usepackage{multirow}
\usepackage{algorithm}
\usepackage{algpseudocode}
\usepackage{hyperref}
\usepackage[justification=centering]{caption}
\usepackage{nth}
\usepackage{balance}

\begin{document}
%\doublespace
\title{Unsupervised Learning for Color Constancy}
\author{{%
Nikola Bani\'{c}, Karlo Ko\v{s}\v{c}evi\'{c}, and Sven Lon\v{c}ari\'{c}}%
}

\maketitle

\renewcommand\footnoterule{{\hrule height 0pt}}

\renewcommand\footnoterule{{\hrule height 0pt}}
\let\thefootnote\relax\footnotetext{

This work has been supported by the Croatian Science Foundation under Project IP-06-2016-2092.

Copyright (c) 2017 IEEE. Personal use of this material is permitted. However, permission to use this material for any other purposes must be obtained from the IEEE by sending a request to pubs-permissions@ieee.org.

The authors are with the Image Processing Group, Department of Electronic Systems and Information Processing, Faculty of Electrical Engineering and Computing, University of Zagreb, 10000 Zagreb, Croatia (email: nikola.banic@fer.hr; karlo.koscevic@fer.hr; sven.loncaric@fer.hr).

This manuscript is extended from the conference paper version~\cite{banic2018unsupervised}.
}

\begin{abstract}
Most digital camera pipelines use color constancy methods to reduce the influence of illumination and camera sensor on the colors of scene objects. The highest accuracy of color correction is obtained with learning-based color constancy methods, but they require a significant amount of calibrated training images with known ground-truth illumination. Such calibration is time consuming, preferably done for each sensor individually, and therefore a major bottleneck in acquiring high color constancy accuracy. Statistics-based methods do not require calibrated training images, but they are less accurate. In this paper an unsupervised learning-based method is proposed that learns its parameter values after approximating the unknown ground-truth illumination of the training images, thus avoiding calibration. In terms of accuracy the proposed method outperforms all statistics-based and many learning-based methods. An extension of the method is also proposed, which learns the needed parameters from non-calibrated images taken with one sensor and which can then be successfully applied to images taken with another sensor. This effectively enables inter-camera unsupervised learning for color constancy. Additionally, a new high quality color constancy benchmark dataset with $1707$ calibrated images is created, used for testing, and made publicly available. The results are presented and discussed. The source code and the dataset are available at \url{http://www.fer.unizg.hr/ipg/resources/color_constancy/}.
\end{abstract}


\IEEEpeerreviewmaketitle

\section{Introduction}
\label{sec:introduction}

\IEEEPARstart{B}{eside} other abilities the human visual system~(HVS) can recognize colors of scene objects even under various illumination. This ability is known as color constancy~\cite{ebner2007color} and most digital cameras have computational color constancy implemented in their image processing pipelines~\cite{kim2012new}. The task of computational color constancy is to get an accurate illumination estimation and then use it to chromatically adapt the image in order to remove the influence of the illumination on colors. The most commonly used image $\mathbf{f}$ formation model for this problem with included Lambertian assumption is~\cite{gijsenij2011computational}
\begin{equation}
\label{eq:image}
f_c(\mathbf{x})=\int^{}_{\omega} I(\lambda, \mathbf{x}) R(\lambda, \mathbf{x}) \rho_c (\lambda) d\lambda
\end{equation}
where $c\in\{R, G, B\}$ is a color channel, $\mathbf{x}$ is a given image pixel, $\lambda$ is the wavelength of the light, $\omega$ is the visible spectrum, $I(\lambda, \mathbf{x})$ is the spectral distribution of the light source, $R(\lambda, \mathbf{x})$ is the surface reflectance, and $\rho_c(\lambda)$ is the camera sensitivity of color channel $c$. To make the problem simpler, uniform illumination is usually assumed and by removing $\mathbf{x}$ from $I(\lambda, \mathbf{x})$, the observed light source color is given as
\begin{equation}
\label{eq:e}
\mathbf{e}=\left(\begin{array}{c}e_R\\e_G\\e_B\end{array}\right)=\int^{}_{\omega} I(\lambda)\boldsymbol{\rho}(\lambda)d\lambda.
\end{equation}

By knowing only the direction of $\mathbf{e}$, an image can be successfully chromatically adapted~\cite{barnard2002comparison}. With only image pixel values $\mathbf{f}$ given and both $I(\lambda)$ and $\boldsymbol{\rho}(\lambda)$ unknown, calculating $\mathbf{e}$ is an ill-posed problem, which needs additional assumptions to be solved. Many illumination estimation methods with different assumptions have been proposed. In the first of two main groups of illumination estimation methods are low-level statistics-based methods such as White-patch~\cite{land1977retinex, funt2010rehabilitation} and its improvements \cite{banic2013using, banic2014color, banic2014improving}, Gray-world~\cite{buchsbaum1980spatial}, Shades-of-Gray~\cite{finlayson2004shades}, Grey-Edge~(\nth{1} and \nth{2} order)~\cite{van2007edge}, Weighted Gray-Edge~\cite{gijsenij2012improving}, using bright pixels~\cite{joze2012role}, using bright and dark colors~\cite{cheng2014illuminant}, exploiting illumination color statistics perception~\cite{banic2019blue}, using gray pixels~\cite{quian2019revisiting}, exploiting expected illumination statistics~\cite{banic2018green}. The second main group consists of learning-based methods, all of which are supervised, like gamut mapping~(pixel, edge, and intersection based)~\cite{barnard2000improvements,finlayson2006gamut}, using neural networks~\cite{cardei2002estimating}, using high-level visual information~\cite{van2007using}, natural image statistics~\cite{gijsenij2007color}, Bayesian learning~\cite{gehler2008bayesian}, spatio-spectral learning~(maximum likelihood estimate, and with gen. prior)~\cite{chakrabarti2012color}, simplifying the illumination solution space~\cite{banic2015color, banic2015using, banic2015acolor}, using color/edge moments~\cite{finlayson2013corrected}, using regression trees with simple features from color distribution statistics~\cite{cheng2015effective}, performing various kinds of spatial localizations~\cite{barron2015convolutional, barron2017fast}, using convolutional neural networks~\cite{bianco2015color, shi2016deep, hu2017fc4, oh2017approaching}, using genetic algorithms~\cite{koscevic2019color}, modelling colour constancy by using the overlapping asymmetric Gaussin kernels with surround pixel contrast based sizes~\cite{akbarinia2018colour}, finding paths for the longest dichromatic line produced by specular pixels~\cite{woo2018improving}, detecting grey pixels with specific illuminant-invariant measuse in logarithmic space~\cite{yang2015efficient}, channel-wise pooling the responses double-oponnecy cells in LMS color space~\cite{gao2015color}. Statistics-based methods are characterized by a relatively high speed, simplicity, and usually lower accuracy, while learning-based methods are slower, but have higher accuracy. However, several recently proposed learning-based methods are not only highly accurate, but also as fast as statistics-based methods to the level of outperforming some of them~\cite{banic2015acolor, cheng2015effective}. This trend will likely continue and it will bring more accurate real-time color constancy to digital cameras.

Nevertheless, since all well-known learning-based methods are supervised, a major obstacle for their application is that for a given sensor, despite proposed workarounds~\cite{gao2017improving}, supervised learning-based methods have to be trained on calibrated images taken by preferably the same sensor~\cite{aytekin2017dataset}. To calibrate the images, a calibration object has to be placed in the scenes of these images and later segmented to extract the ground-truth illumination. Careful image acquisition and the amount of manual work required for calibration is the main bottleneck in enabling highly accurate color constancy for a given sensor.

To try to avoid such calibration, in this paper an unsupervised learning-based method is proposed that learns its parameter values from non-calibrated images with unknown ground-truth illumination. Such learning is possible by clustering the approximated ground-truth illuminations of images from the training set and then extracting information useful for illumination estimation on future new images. The method is fast, hardware-friendly, and it outperforms many state-of-the-art methods in terms of accuracy. To the best of the authors' knowledge this is the first unsupervised learning-based color constancy method with high accuracy on well-known and widely used benchmark datasets and therefore it represents a potential contribution to the color constancy philosophy.

Besides being a clear proof-of-concept that using unsupervised learning for highly accurate color constancy is possible, the proposed method opens another valuable and useful possibility, namely that of automatic online learning and adjustment of parameter values when a camera is used in special illumination conditions for a prolonged period of time.

An extension of the method is also proposed, which learns the needed parameters from non-calibrated images taken with one sensor and which can then be successfully applied to images taken with another sensor. This effectively enables inter-camera unsupervised learning for color constancy.

Additionally, a new high quality color constancy benchmark dataset with $1707$ calibrated high-quality images is created, used to test the proposed method, and made publicly available.

In short, the contributions of the paper are as follows:
\begin{itemize}
	\item a method for unsupervised learning for illumination estimation that was already published in a conference paper in~\cite{banic2018unsupervised}, but with more experimental results in this paper;
	\item a method for unsupervised learning for inter-camera illumination estimation, which simultaneously has the advantage of camera model and ground-truth independence;
	\item and finally a new large color constancy benchmark dataset with images of various illuminations taken both indoor and outdoor in several countries during day and night.
\end{itemize}

The paper is structured as follows: Section~\ref{sec:motivation} lays out the motivation for the proposed method, Section~\ref{sec:method} describes the method, Section~\ref{sec:inter} extends the proposed method to perform inter-camera learning, the general applicability of the proposed method is shown in Section~\ref{sec:applicability}, in Section~\ref{sec:results} the newly created dataset and the experimental results are presented and discussed, and, finally, Section~\ref{sec:conclusions} concludes the paper.

\begin{figure}[htb]
    \centering
    
	\includegraphics[width=\linewidth]{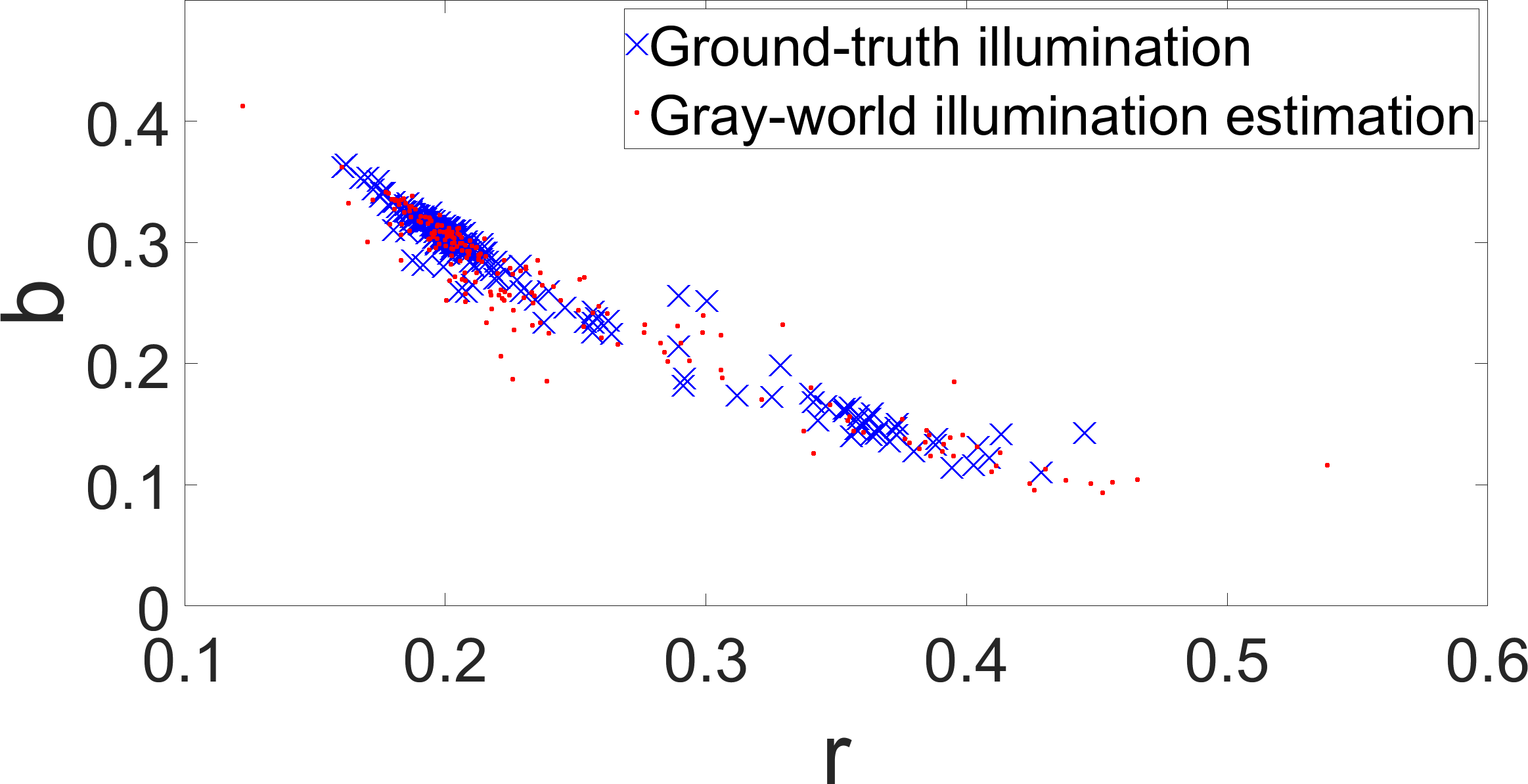}
	
    \caption{The $rb$-chromaticities of the ground-truth illuminations and Gray-world illumination estimations for images of the Samsung benchmark dataset~\cite{cheng2014illuminant}.}
	\label{fig:gw}
    
\end{figure}

%------------------------------------------------------------------------

\section{Motivation}
\label{sec:motivation}

Ground-truth illumination of training images for supervised learning-based methods is extracted from calibration objects placed in the image scenes. As explained in the introduction, obtaining the ground-truth illumination is time consuming, but it enables supervised learning and high illumination estimation accuracy. To reduce the amount of required time, usage of calibration objects has to be dropped out. Then in place of the real ground-truth illumination, some kind of its approximation has to be used instead, e.g. illumination estimations obtained by means of statistics-based methods that require no previous learning. But since they are usually less accurate than learning-based methods, using their estimations as the ground-truth illumination may be counterproductive. However, instead of only image-based illumination estimation, there are other kinds of information that such methods provide. Namely, even illumination estimations of the simplest statistics-based methods appear ''to correlate \textit{roughly} with the actual illuminant''~\cite{finlayson2013corrected} as shown in Fig~\ref{fig:gw} i.e. they occupy roughly the same region in the chromaticity plane. To have a better insight into this phenomenon, some additional numerical analysis is required.

\begin{figure}[htb]
    \centering
    
	\includegraphics[width=\linewidth]{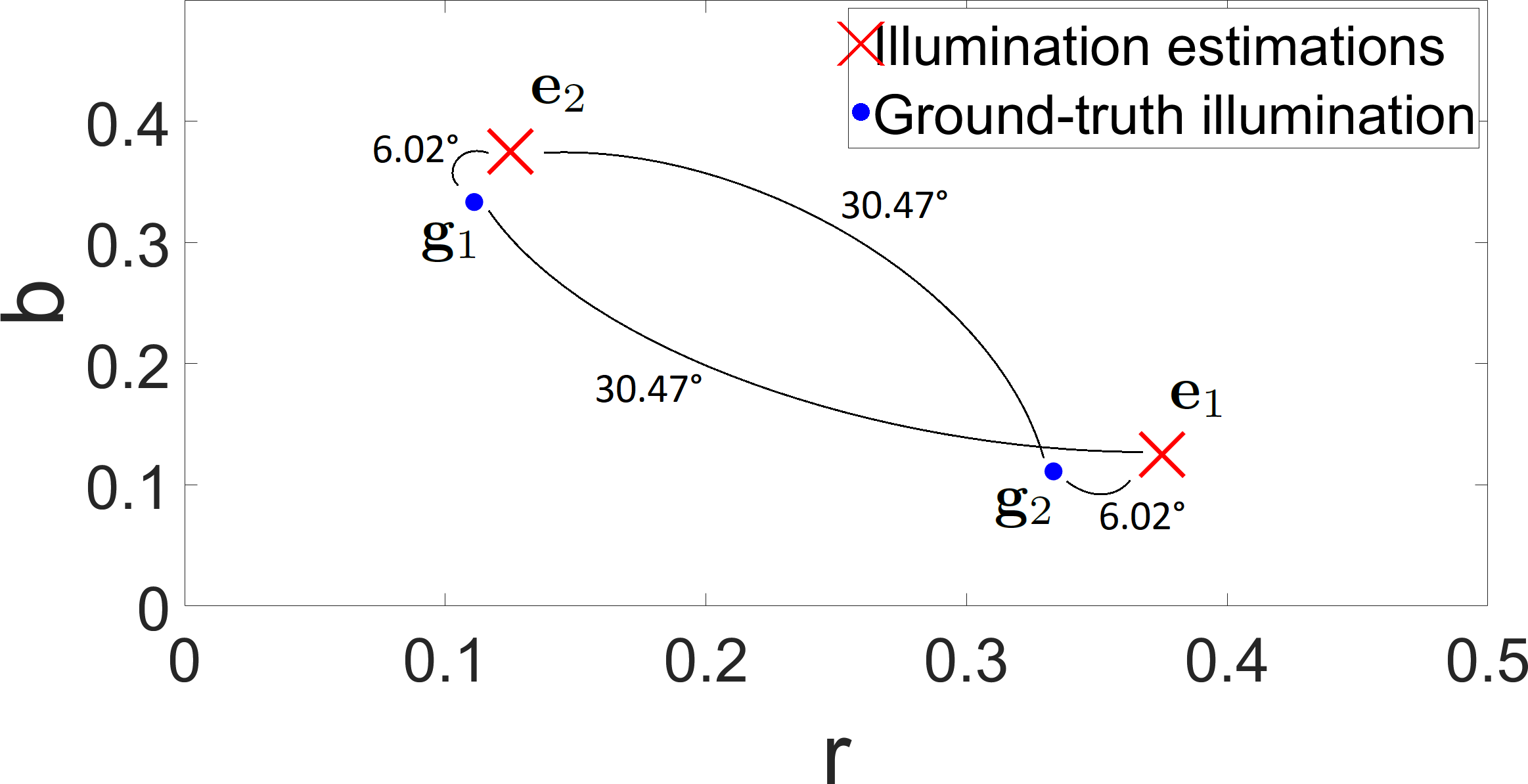}
	
    \caption{Illumination estimations for two images that are highly inaccurate, but have a significantly lower SAE.}
	\label{fig:sae}
    
\end{figure}

As described in more detail later in Section~\ref{subsec:setup}, the error measure for accuracy of illumination estimation is the angular error i.e. the angle between the vectors of ground-truth illumination and illumination estimation. One way to see how well a set of illumination estimations numerically resembles the set of ground-truth illuminations in terms of occupying the same region in the chromaticity space is to rearrange the existing illumination estimations between images in order to minimize the sum of overall angular errors obtained for such rearranged illumination estimations. More formally, if there are $M$ images, $\mathbf{g}_i$ is the ground-truth illumination for the $i$-th image, $\mathbf{e}_i$ is the illumination estimation for the $i$-th image, $a_{i,j}$ is the angle between $\mathbf{g}_i$ and $\mathbf{e}_j$ i.e. $a_{i,j}=\angle \left(\mathbf{g}_i,\mathbf{e}_j\right)$, $\{r_{i,j}\}_{M\times M}$ is a binary matrix where $r_{i,j}=1$ if and only if the ground-truth of the $i$-th image is assigned to the $j$-th illumination estimation, then the goal is to minimize the mean angular error $\frac{1}{M}\sum^{M}_{i=1}\sum^{M}_{j=1}r_{i,j}a_{i,j}$ under the constraints $\sum^{M}_{j=1}r_{i,j}=1, \forall i\in\{1\dots M\}$ and $\sum^{M}_{i=1}r_{i,j}=1, \forall j\in\{1\dots M\}$. For the sake of simplicity, from now on this minimal possible mean angular error for pairs $\left(\mathbf{g}_i,\mathbf{e}_j\right)$ for which $r_{i,j}=1$ for a given set of ground-truth illuminations and a given set of illumination estimations will be denoted as Sets' Angular Error~(SAE). Effectively, calculating SAE boils down to solving the optimal assignment problem~\cite{burkard2012assignment}. It must be clearly stressed here that a low SAE does not implicate an accurate method; an inaccurate method can under certain conditions produce estimations with a low SAE as shown in Fig.~\ref{fig:sae}. There the angular error for the ground-truth and illumination estimation in the case of both the first and the second image is $30.47^{\circ}$. However, if only the overall unordered positions of all ground-truths and illumination estimations are considered, they occupy roughly the same places and the angle between the members of pairs $\left(\mathbf{g}_1,\mathbf{e}_2\right)$ and $\left(\mathbf{g}_2,\mathbf{e}_1\right)$ obtained when calculating SAE is $6.02^{\circ}$.

\begin{figure}[htb]
    \centering
    
	\includegraphics[width=\linewidth]{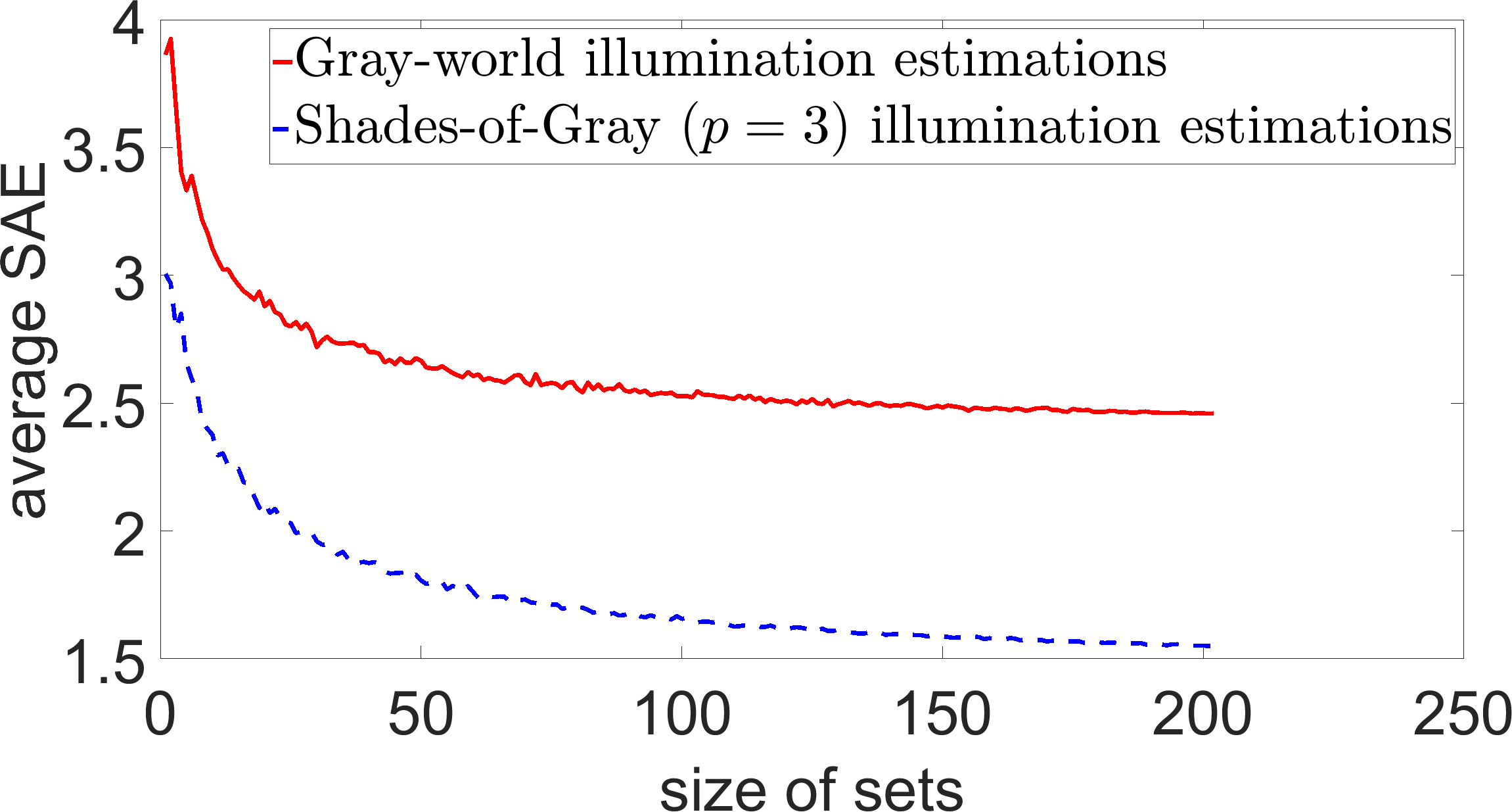}
	
    \caption{Values of SAE averaged over $1000$ random subsets of the Sony benchmark dataset~\cite{cheng2014illuminant} for various subset sizes; $p$ is the Minkowski norm power used by Shades-of-Gray~\cite{finlayson2004shades}.}
	\label{fig:saes}
    
\end{figure}

As the number of points in the sets grows, SAE should decrease since every point will have more pairing opportunities. This is shown in Fig.~\ref{fig:saes} where the values of SAE averaged over $1000$ random subsets of the Sony benchmark dataset~\cite{cheng2014illuminant} decrease as the size of the used subsets increases. Based on the obtained empirical evidence, including the results shown in Fig.~\ref{fig:saes}, it can be concluded that the SAE is seemingly more influenced by the method choice then by the set size.

These results show that by applying well chosen methods to a sufficient number of given images it is possible to obtain a low SAE, which is a proof of concept that a relatively accurate approximation of the set of unknown ground-truth illuminations for these images is feasible. This definitely motivates to exploit the demonstrated concept further, but to have a practical use of it, at least two questions need to be answered: first, what other information useful for a more accurate illumination estimation can be extracted from a set of ground-truth illumination approximations, and second, how to obtain such approximated sets that have a low SAE?

As for the first question, the ground-truth illuminations or their approximations for many images can reveal in which chromaticity space regions are future illumination estimations of new images most likely to appear. There are several methods that rely on such kind of information~\cite{banic2015color, mazin2015estimation, cheng2015effective, banic2015acolor} with probably the least demanding one being the Color Dog method~\cite{banic2015acolor}. During the training phase it clusters the ground-truth illuminations by using the $k$-means clustering~\cite{vassilvitskii2007k} with the angular instead of the Euclidean distance. The cluster centers obtained in this process become the only illumination estimations that the method will ever produce when used later in production. When applied to a new image, Color Dog first runs the White-patch~\cite{funt2010rehabilitation} and Gray-world methods~\cite{buchsbaum1980spatial}. Under the Gray-world assumption the average scene reflectance is achromatic and the illumination estimation is calculated as
\begin{equation}
\frac{\int\textbf{f}(x)dx}{\int dx}=\mathbf{e}_{GW}.
\label{eq:gw}
\end{equation}
The White-patch method assumes that the illumination can be recovered from the maximum intensities of color channels as
\begin{equation}
\max_{\mathbf{x}}f_c\left(\mathbf{x}\right)=e_{WP, c}.
\label{eq:wp}
\end{equation}
Both Gray-world and White-patch have low accuracy, but they have no parameters, which means that they do not require any kind of tuning, and they are simple and practical to implement. While some other methods could give higher accuracy, their choice would require additional tuning and probably higher computation time. After the Color Dog method applies them, the angular distances between their illumination estimations and the learned cluster centers are used as weighted votes to determine which center should represent the illumination on the given image. If $\mathbb{C}$ is the set of learned centers, then this is
\begin{equation}
\label{eq:voting}
\mathbf{e}=\underset{\mathbf{c}_i \in \mathbb{C}}{\operatorname{\arg\max}} \left( \frac{\mathbf{c}_i \cdot \mathbf{e}_{GW}}{||\mathbf{c}_i || \cdot || \mathbf{e}_{GW}||} + \frac{\mathbf{c}_i \cdot \mathbf{e}_{WP}}{||\mathbf{c}_i || \cdot || \mathbf{e}_{WP}||} \right).
\end{equation}
The votes for a given learned center are calculated as the sum of cosine values of angles between the center and each of the estimations of the Gray-world and White-patch methods i.e. Eq.~\eqref{eq:voting} will pick the center that is closer to both estimations. Similar effect is achieved by minimizing the sum of the angles. Well positioned centers in the chromaticity plane result in relatively small errors~\cite{banic2015acolor} so despite its simplicity, Color Dog is highly accurate. It must be stressed again that it is the discrete simplification of the solution space that enables higher accuracy by significantly stabilizing the otherwise relatively inaccurate results of the Gray-world and White-patch methods, even when combined. The centers and their number are learned through nested cross-validation~\cite{japkowicz2011evaluating}. Since accurate ground-truth illuminations are needed for such learning, using approximations gives poor results, but the main idea of Color Dog can be the basis for a method that learns from approximations. Such a new method is proposed in the following section.

The reason for choosing Color Dog as the basis for the new method is its simplicity. Namely, the requirement of learning only a few centers should be highly appropriate for the targeted conditions of missing ground-truth where all assumptions should be carefully chosen. To find more details about the Color Dog method, its properties, and more about the theory behind it, the reader is referred to~\cite{banic2015acolor} since going into more details about the assumptions and theory behind the Color Dog method would be out of the scope of this paper.

%------------------------------------------------------------------------

\section{The proposed method}
\label{sec:method}

Nested cross-validation can be circumvented by simply fixing the number of centers. The simplest solution is to use a single center, but the detrimental effect of this has already been shown in~\cite{banic2015perceptual}. Using more centers increases the upper limit for accuracy because of the finer chromaticity space representation, but it also poses a harder classification problem for which the upper accuracy limit may be rarely reached. Empirical results have shown that using more than two centers mostly leads to lower accuracy. Thus the new method proposed here uses only two centers and assumes that most images can be classified as having either a warmer i.e. reddish or a cooler i.e. blueish illumination, which is effectively a simplification of the Planckian locus~\cite{schanda2007colorimetry} that has already been used for illumination estimation in several methods~\cite{banic2015color, mazin2015estimation}. A somewhat similar rough division to an indoor and outdoor type illumination has been successfully used for a slightly different purpose in~\cite{cheng2016two}. As stated in Section~\ref{sec:introduction}, assumptions are needed to tackle the ill-posed nature of color constancy and in the rest of the paper the described assumption will be denoted as \textbf{the two illuminations assumption}. In Section~\ref{sec:results} this assumption is generally shown to be effectively valid and it is shown what to do if it does not hold for a training set.

\begin{figure*}[htb]
    \centering
    
	\subfloat[]{
		\parbox{0.32\linewidth}{
			\includegraphics[width=1\linewidth]{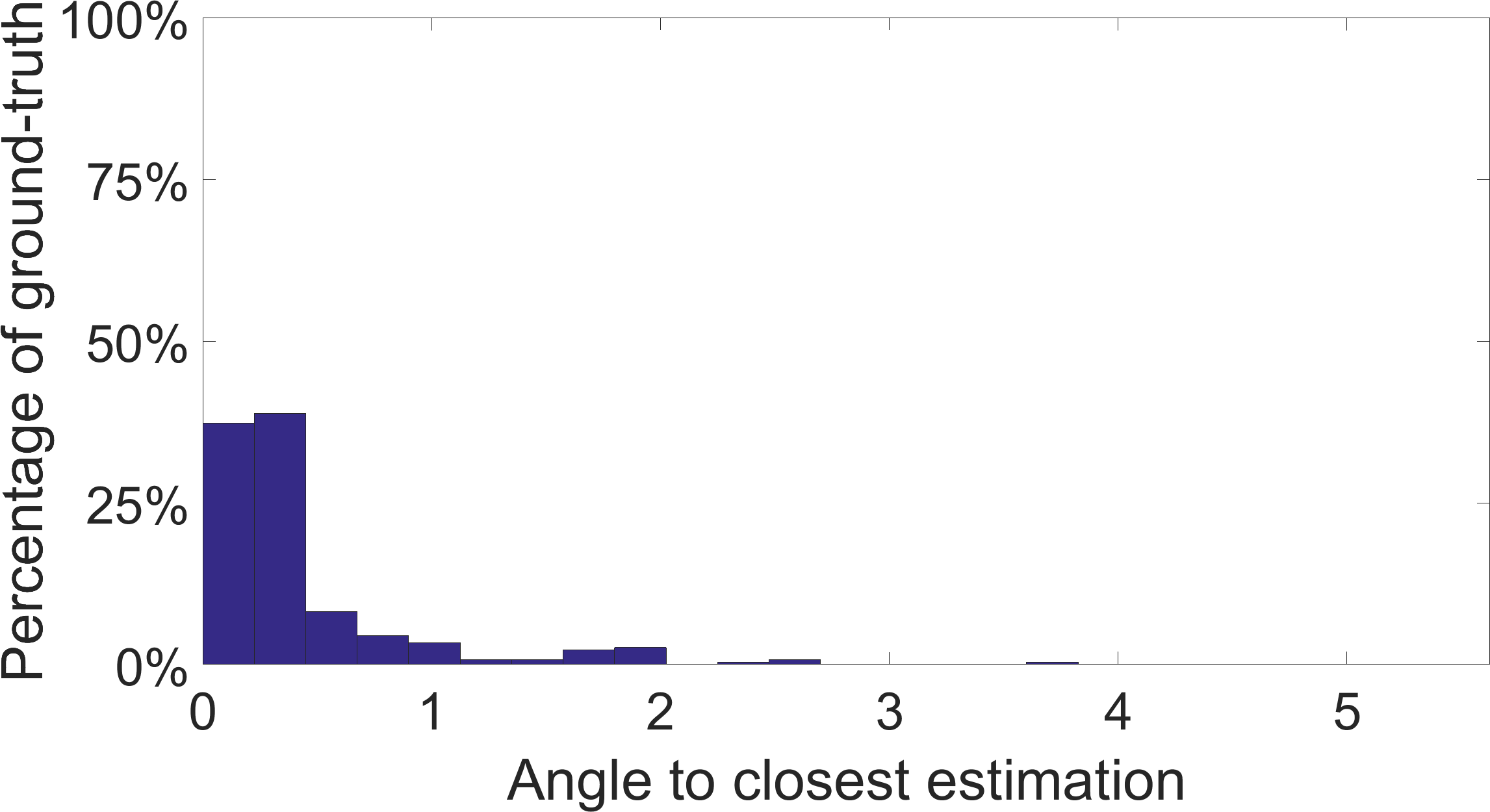}\\\\
			\includegraphics[width=1\linewidth]{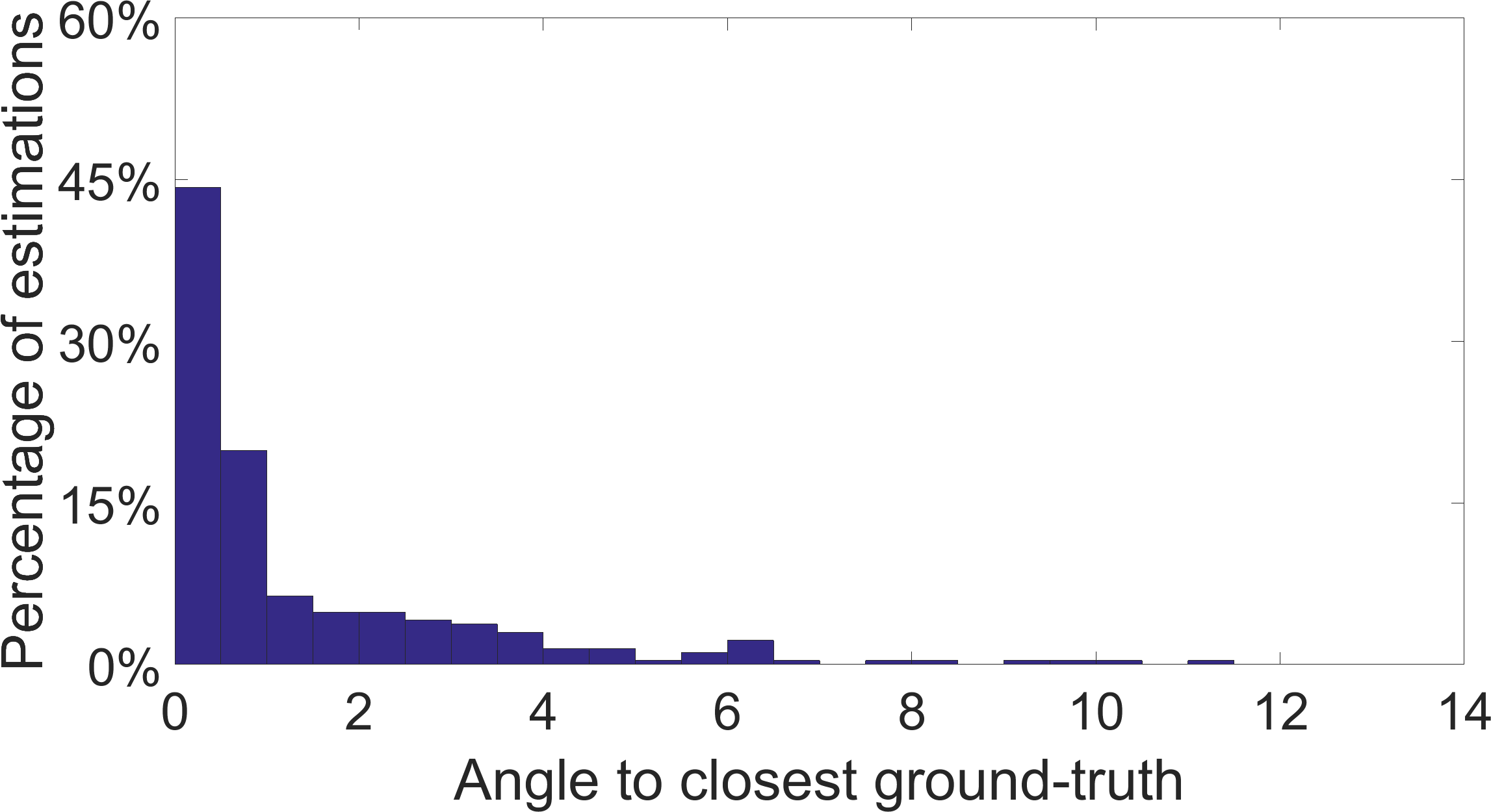}
			}
	\label{fig:n_1}
	}%
	~%
	\subfloat[]{
		\parbox{0.32\linewidth}{
			\includegraphics[width=1\linewidth]{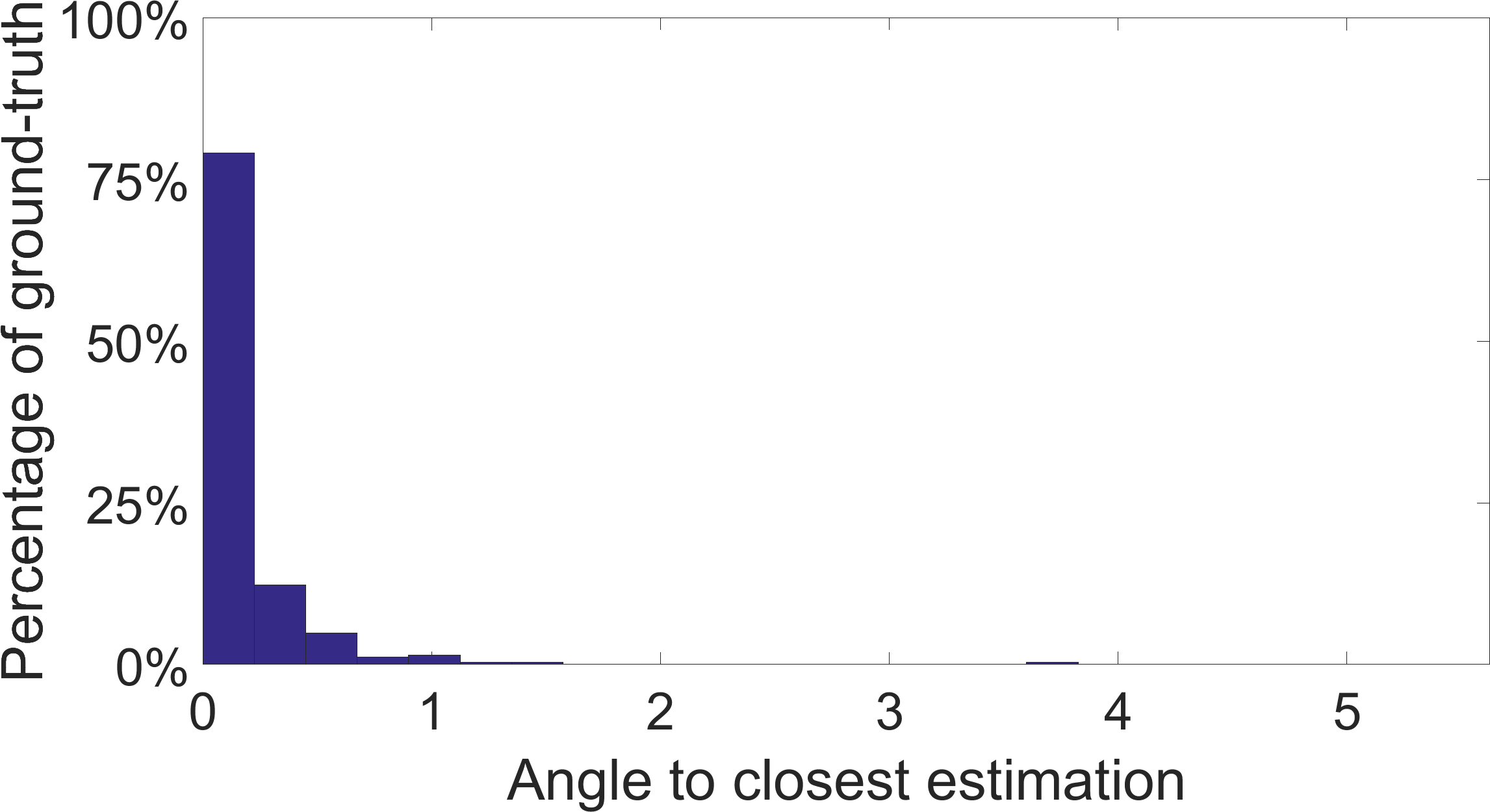}\\\\
			\includegraphics[width=1\linewidth]{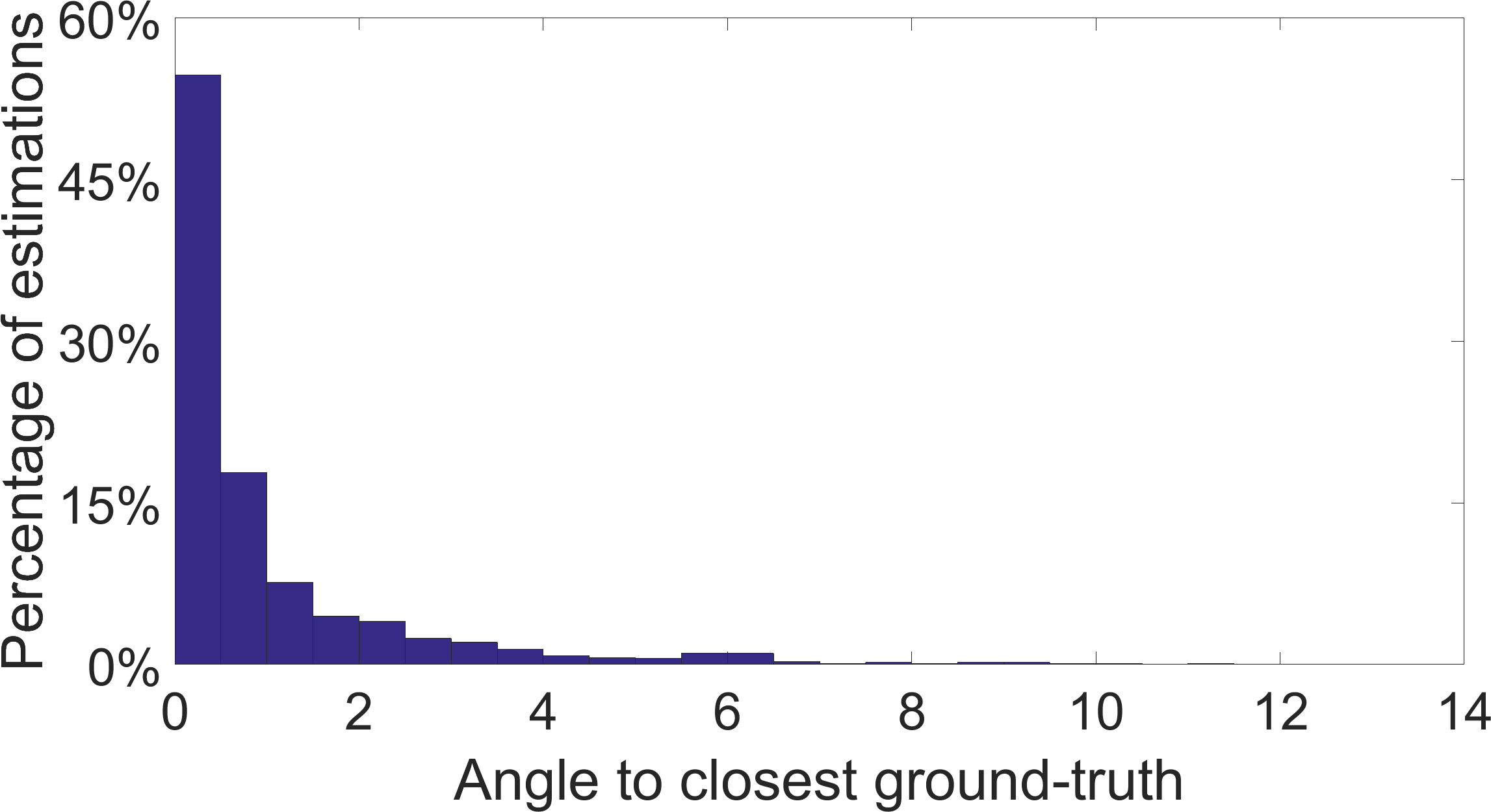}
			}
	\label{fig:n_4}
	}%
	~%
	\subfloat[]{
		\parbox{0.32\linewidth}{
			\includegraphics[width=1\linewidth]{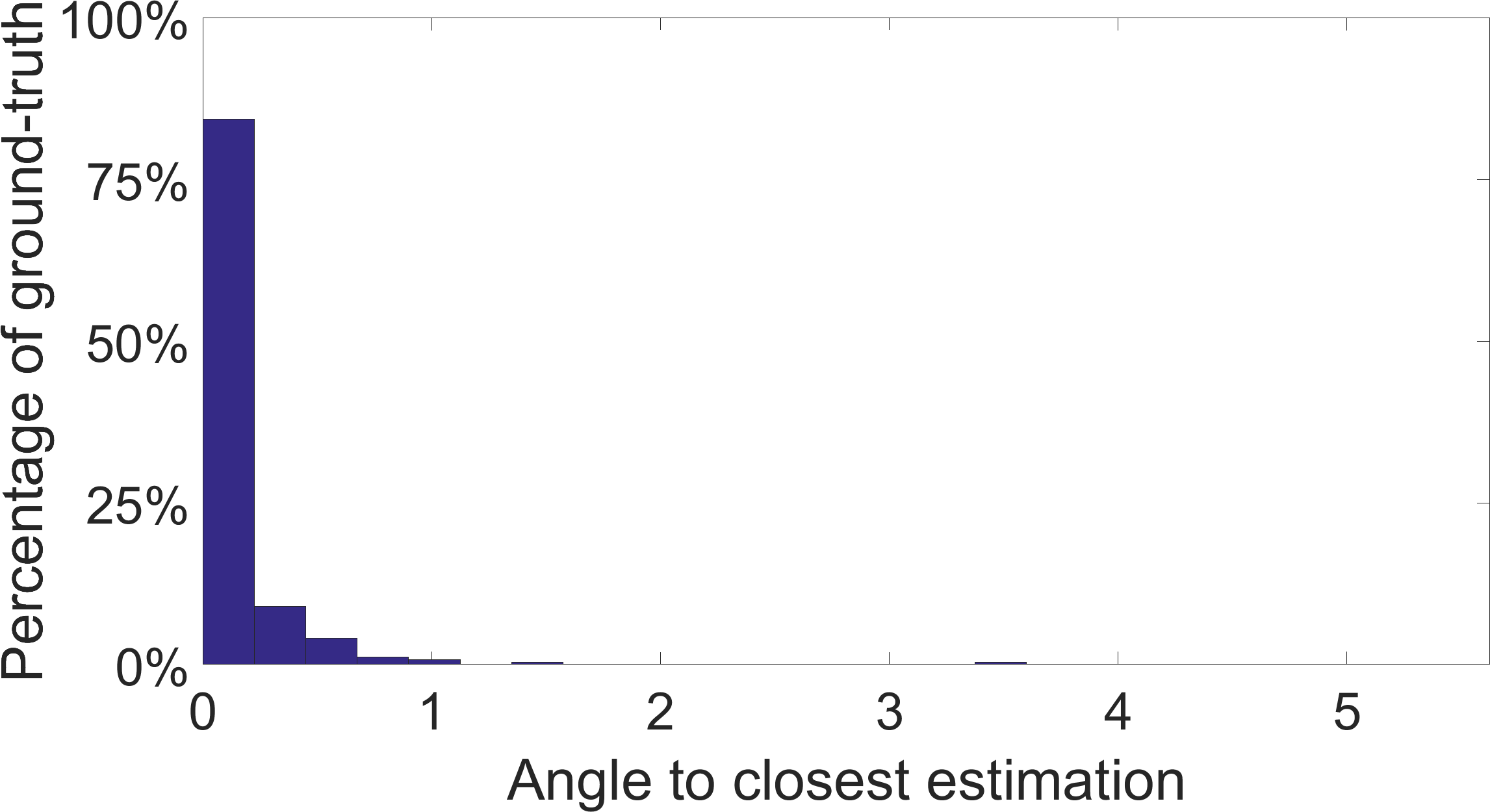}\\\\
			\includegraphics[width=1\linewidth]{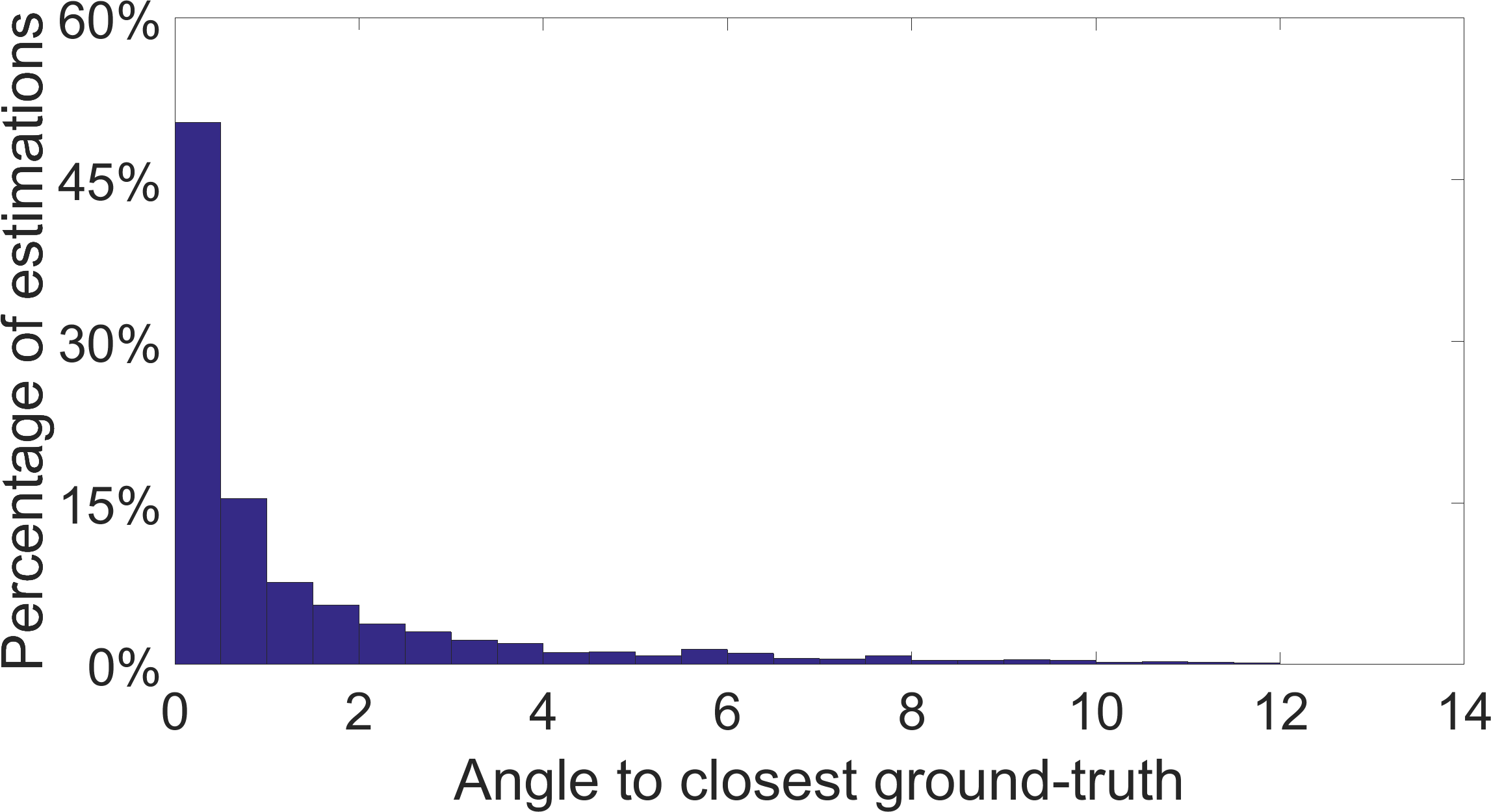}
			}
	\label{fig:n_8}
	}
	
    \caption{Percentage of ground-truth with specified angle to the closest estimation (first row) and vice versa (second row) obtained with Shades-of-Gray on images of the Sony benchmark dataset~\cite{cheng2014illuminant} for (a)~$n=1$, (b)~$n=4$, and (c)~$n=8$. Taken from~\cite{banic2018unsupervised}.}
	\label{fig:closest}
    
\end{figure*}

With the answer to the first question from the previous section proposed, it remains to resolve the second one i.e. which illumination estimations should be clustered to get centers that are well positioned among the ground-truth illumination? A single statistics-based method with fixed parameter values may achieve a relatively low SAE, but with unknown ground-truth illuminations, it cannot be said which parameter values will result in minimal SAE. To solve this problem, it can be assumed that for any set of parameter values of a statistics-based method in most cases there will be a number of training images for which the method's illumination estimations will be accurate. Other parameter values should again give accurate estimations for some other images. By repeating the illumination estimation for more sets of parameter values and combining the results, the region with the actual ground-truth illumination should be more densely filled with illumination estimations than other regions. Such behaviour can also be observed for the Shades-of-Gray~(SoG)~\cite{finlayson2004shades} method, which uses the Minkowski norm $p$ for illumination estimation 
\begin{equation}
\left(\int\left(f_c(\mathbf{x})\right)^p d\mathbf{x}\right)^\frac{1}{p}=e_c.
\label{eq:sog}
\end{equation}
SoG already offers a diversity of illumination estimations by only changing the value of its single parameter. While other statistics-based methods like Gray-Edge may be more accurate, this holds only if their multiple parameters are well chosen. In order to avoid possible problems related to parameter value tuning, the proposed method clusters combined SoG illumination estimations for $p\in\{1,2,...,n\}$. Since in such combination there are several illumination estimations per image, SAE cannot be used because of its definition. An alternative for measuring how well such combined illumination estimations occupy the space around the ground-truth illuminations is to check the histograms of angles between the illumination estimations and their closest ground-truth illuminations and vice versa. Such histograms also provide more information then a single number such as SAE. Fig.~\ref{fig:closest} shows the influence of $n$ on the mentioned histograms. It can be observed that using combined SoG estimations for various values of $p$ can indeed result in a more accurate coverage of the chromaticity plane regions populated with ground-truth illuminations. Theoretically this should also improve the representation accuracy of obtained clustering centers.

\begin{figure*}[htb]
    \centering
    
	\subfloat[]{
	\includegraphics[width=0.48\linewidth]{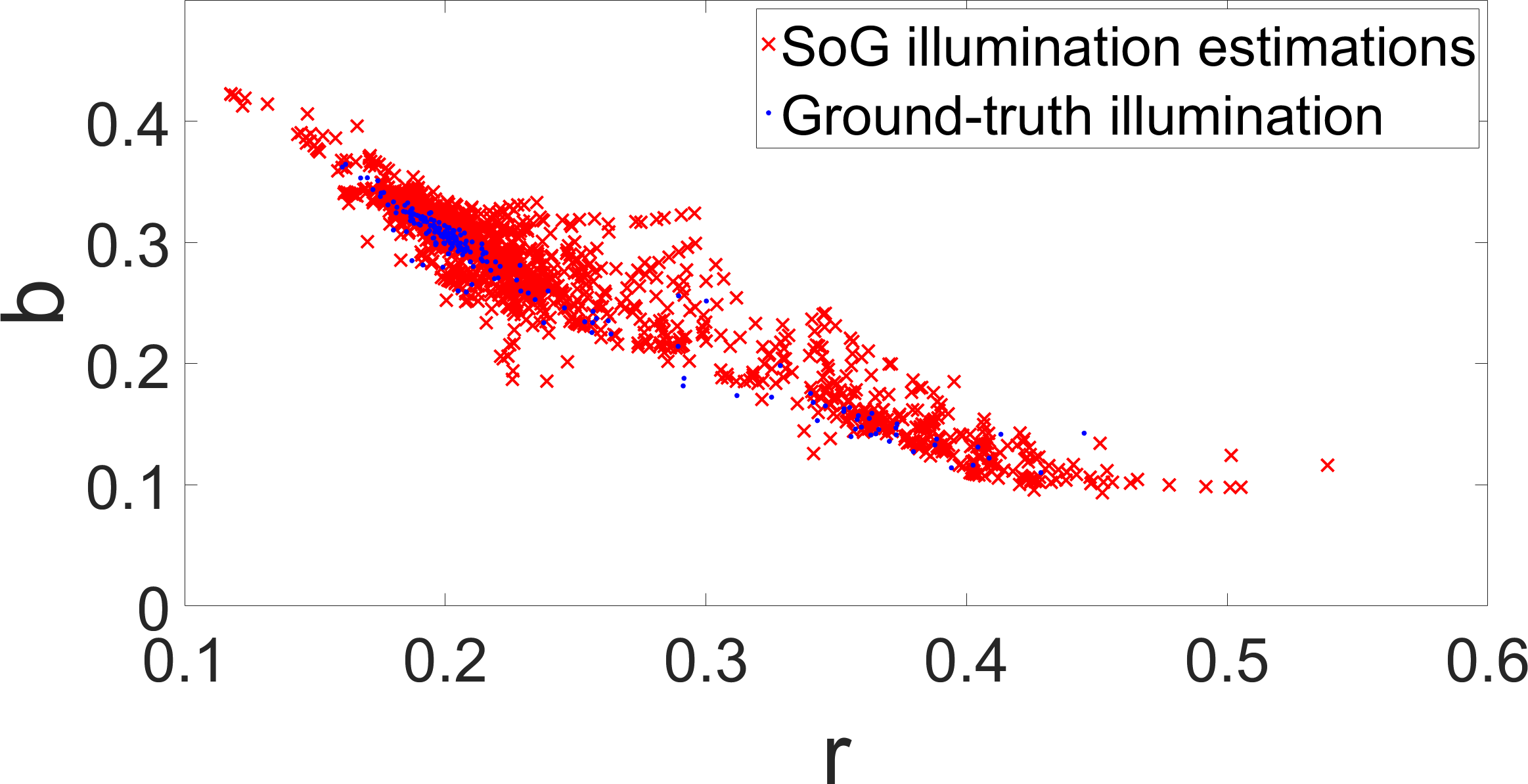}
	\label{fig:sogs}
	}%
	~%
	\subfloat[]{
	\includegraphics[width=0.48\linewidth]{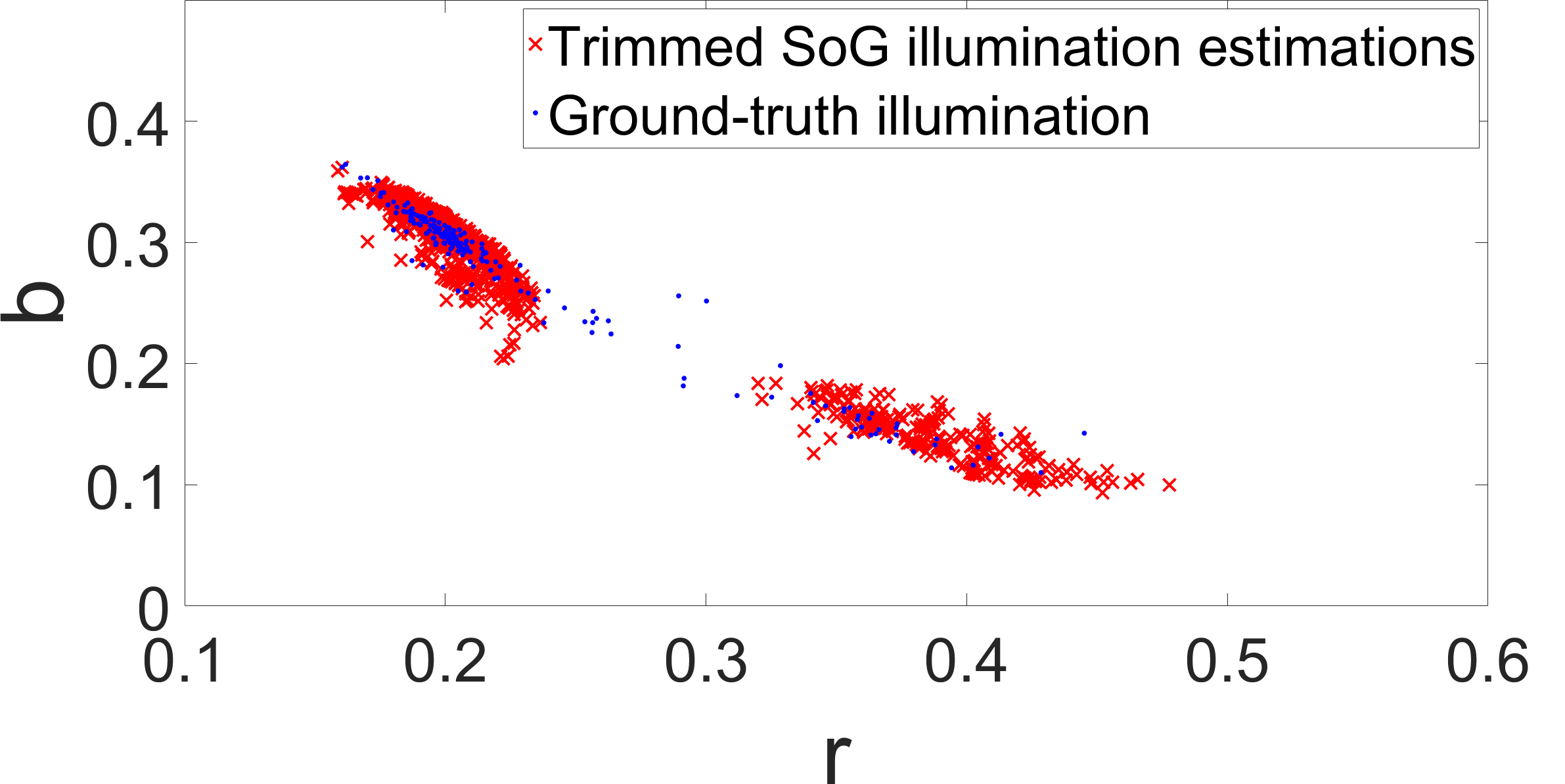}
	\label{fig:trimmed_sogs}
	}
	
    \caption{The $rb$-chromaticities of the ground-truth illuminations and SoG illumination estimations for $n=8$ on images of the Samsung benchmark dataset~\cite{cheng2014illuminant} (a) before and (b) after trimming with $t=0.3$ (best viewed in color). Taken from~\cite{banic2018unsupervised}.}
	\label{fig:trimming}
    
\end{figure*}

\begin{figure}[htb]
    \centering
    
	\includegraphics[width=\linewidth]{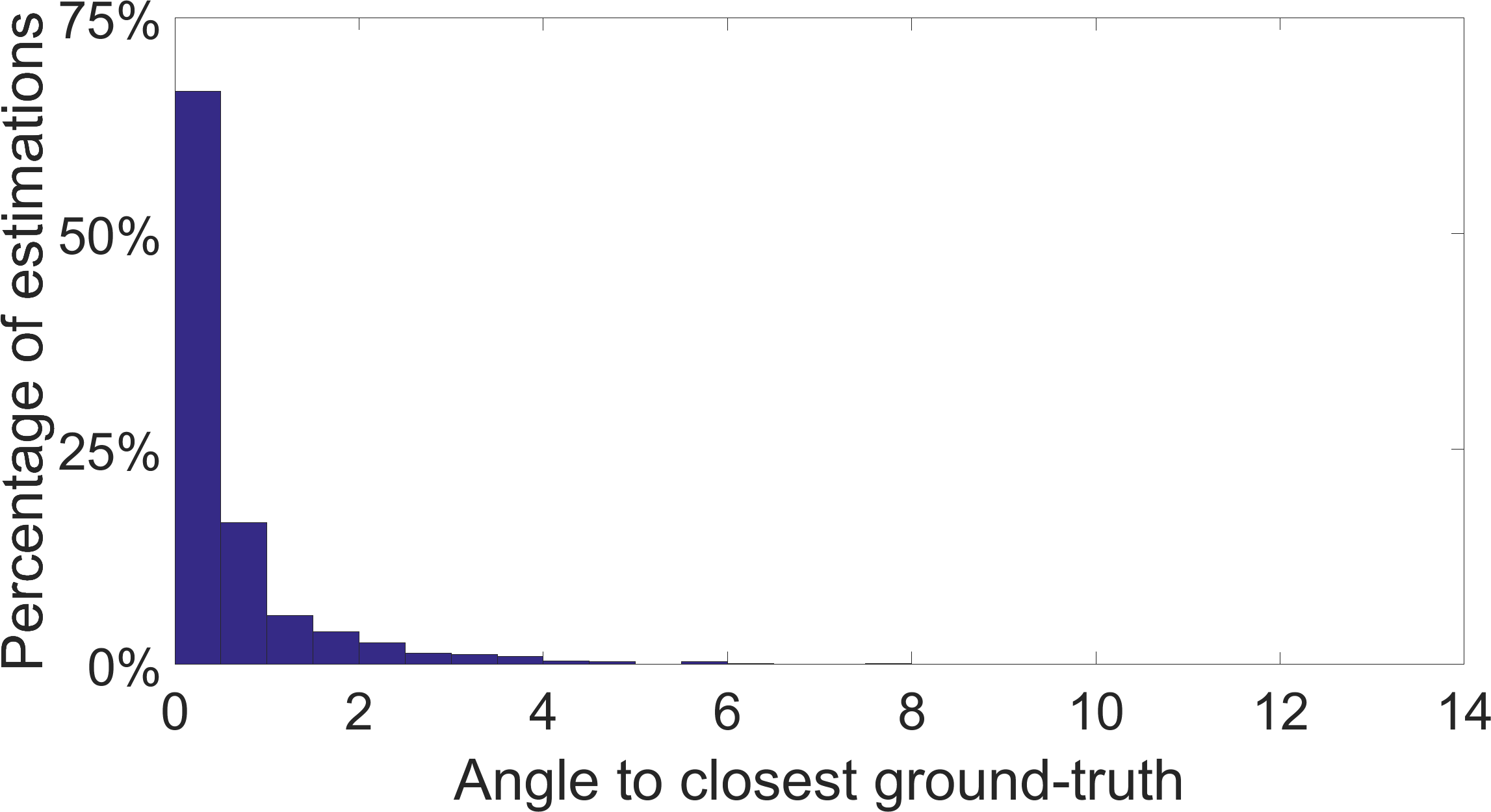}
	
    \caption{Percentage of SoG estimations for $n=8$ with specified angle to the closest ground-truth for the Sony benchmark dataset~\cite{cheng2014illuminant}.}
	\label{fig:trimmed_angles}
    
\end{figure}

However, besides putting more points around the actual chromaticity plane region with the ground-truth, combining estimations for several values of $p$ also introduces a lot of estimations that are far away from all ground-truth illuminations and represent noise. Under the used assumption such estimations should be scattered and less dense than the estimations closer to the ground-truth region and this could be used to reduce their influence. A direct solution would be to use clustering techniques that consider outliers and simply ignore them with one example being DBSCAN~\cite{ester1996density}. However, since DBSCAN and some other similar methods determine the number of centers on their own and additionally the problem here does not involve some arbitrarily shaped clusters, another solution is proposed. After the initial clustering with $k$-means, for each cluster center $100\cdot t$\% of its furthest estimations are removed i.e. trimmed and then clustering is repeated only on the remaining estimations to obtain the final cluster centers. This trimming procedure is summarized in Algorithm~\ref{alg:trimming}. Fig.~\ref{fig:trimming} shows an example of such an outlier removal. The numerical effect of it can be observed when comparing the lower right histogram in Fig.~\ref{fig:closest} and the histogram in Fig.~\ref{fig:trimmed_angles}, which shows that after trimming the remaining illuminations are much closer to the ground-truth. Default parameter values are set to $n=8$ and $t=0.3$ since they were empirically found to work well. For the experimental results in Section~\ref{sec:results} these values have been used for all benchmark datasets. Tuning them for each dataset individually would result in a significantly higher accuracy. However, that would defeat the whole purpose of unsupervised learning because ground-truth illumination, which is supposed to be unknown to the proposed method, would be needed for such fine tuning of parameter values.

For simpler notation in the experimental results and because the proposed method learns the values of its parameters from images obtained in the wild without knowing their ground-truth illumination, it is named Color Tiger~(CT). Now that the whole theoretical background with all required assumptions has been explained, Color Tiger's training procedure can be simply described as learning the centers of two clusters from a specifically trimmed set of illumination estimations obtained by applying Shades-of-Gray to training images for every $p\in\{1,2,...,8\}$. This is additionally summarized in Algorithm~\ref{alg:training}. The illumination estimation for new images resembles the one of the Color Dog method and it is described in Algorithm~\ref{alg:application}.
\begin{algorithm}
\caption{Trimming}
\label{alg:trimming}
\hspace*{\algorithmicindent}\textbf{Input:} data $\mathbb{D}$, number of centers $k$, threshold $t$\\
\hspace*{\algorithmicindent}\textbf{Output:} trimmed data $\mathbb{T}$
\begin{algorithmic}[1]

\State $\mathbb{C}=kmeans(\mathbb{D}, k)$ \Comment Use angular distance

\State $\mathbb{T}\gets\{\}$

\State $r=1-t$

\For{$\mathbf{c}_i\in\mathbb{C}$}
	\State $\mathbb{D}_{\mathbf{c}_i}=\{\mathbf{d}\in\mathbb{D} \mid \mathbf{c}_i=\underset{\mathbf{c}_j \in \mathbb{C}}{\operatorname{\arg\min}}\angle\left(\mathbf{c}_j, \mathbf{d}\right)\}$
	\State $r'=\lfloor 100\cdot r\rfloor$-th percentile of $\{\angle\left(\mathbf{c}_i, \mathbf{d}\right) \mid \mathbf{d}\in\mathbb{D}_{\mathbf{c}_i}\}$
	\State $\mathbb{D}'_{\mathbf{c}_i}=\{\mathbf{d} \mid \angle\left(\mathbf{c}_i, \mathbf{d}\right) \leq r' \}$
	\State $\mathbb{T}\gets\mathbb{T}\cup\mathbb{D}'_{\mathbf{c}_i}$
\EndFor

%\State $SetOutputImage(R)$

\end{algorithmic}

\end{algorithm}
\begin{algorithm}
\caption{Color Tiger Training}
\label{alg:training}
\hspace*{\algorithmicindent}\textbf{Input:} images $\mathbb{I}$, SoG upper power $n$, trimming $t$\\
\hspace*{\algorithmicindent}\textbf{Output:} set of two centers $\mathbb{C}$
\begin{algorithmic}[1]

\State $\mathbb{E}\gets\{\}$

\For{$\mathbf{I}\in\mathbb{I}$}
	\For{$p\in\{1, 2, ..., n\}$}
		\State $\mathbf{e}=ShadesOfGray(\mathbf{I}, p)$ \Comment Apply Eq.~\eqref{eq:sog}
		\State $\mathbb{E}\gets\mathbb{E}\cup\{\mathbf{e}\}$
	\EndFor
\EndFor

\State $\mathbb{E}'=Trimming(\mathbb{E}, 2, t)$ \Comment Algorithm~\ref{alg:trimming}
\State $\mathbb{C}=kmeans(\mathbb{E}', 2)$ \Comment Use angular distance

%\State $SetOutputImage(R)$

\end{algorithmic}

\end{algorithm}
\begin{algorithm}
\caption{Color Tiger Application}
\label{alg:application}
\hspace*{\algorithmicindent}\textbf{Input:} image $\mathbf{I}$, set of two centers $\mathbb{C}$\\
\hspace*{\algorithmicindent}\textbf{Output:} illumination estimation $\mathbf{e}$
\begin{algorithmic}[1]

\State $\mathbf{e}_{GW}=GrayWorld(\mathbf{I})$ \Comment Apply Eq.~\eqref{eq:gw}

\State $\mathbf{e}_{WP}=WhitePatch(\mathbf{I})$ \Comment Apply Eq.~\eqref{eq:wp}

\State $\mathbf{e}=\underset{\mathbf{c}_i \in \mathbb{C}}{\operatorname{\arg\max}} \left( \frac{\mathbf{c}_i \cdot \mathbf{e}_{GW}}{||\mathbf{c}_i || \cdot || \mathbf{e}_{GW}||} + \frac{\mathbf{c}_i \cdot \mathbf{e}_{WP}}{||\mathbf{c}_i || \cdot || \mathbf{e}_{WP}||} \right)$

%\State $SetOutputImage(R)$

\end{algorithmic}

\end{algorithm}

%------------------------------------------------------------------------

\section{Inter-camera learning}
\label{sec:inter}

The Color Tiger method is originally designed to be trained and used on images taken with the same camera sensor. The next step is to extend it so that it can train on images taken with one sensor and be used on images taken with another sensor. A solution to this problem has already been proposed~\cite{gao2017improving}, but it requires calibrated images and reflectance spectras to learn a $3\times 3$ sensor transformation matrix. Here, however, the goal is to use neither calibrated images nor any reflectance spectras in order for the method to be fully unsupervised.

\begin{figure}[htb]
    \centering
    
	\includegraphics[width=\linewidth]{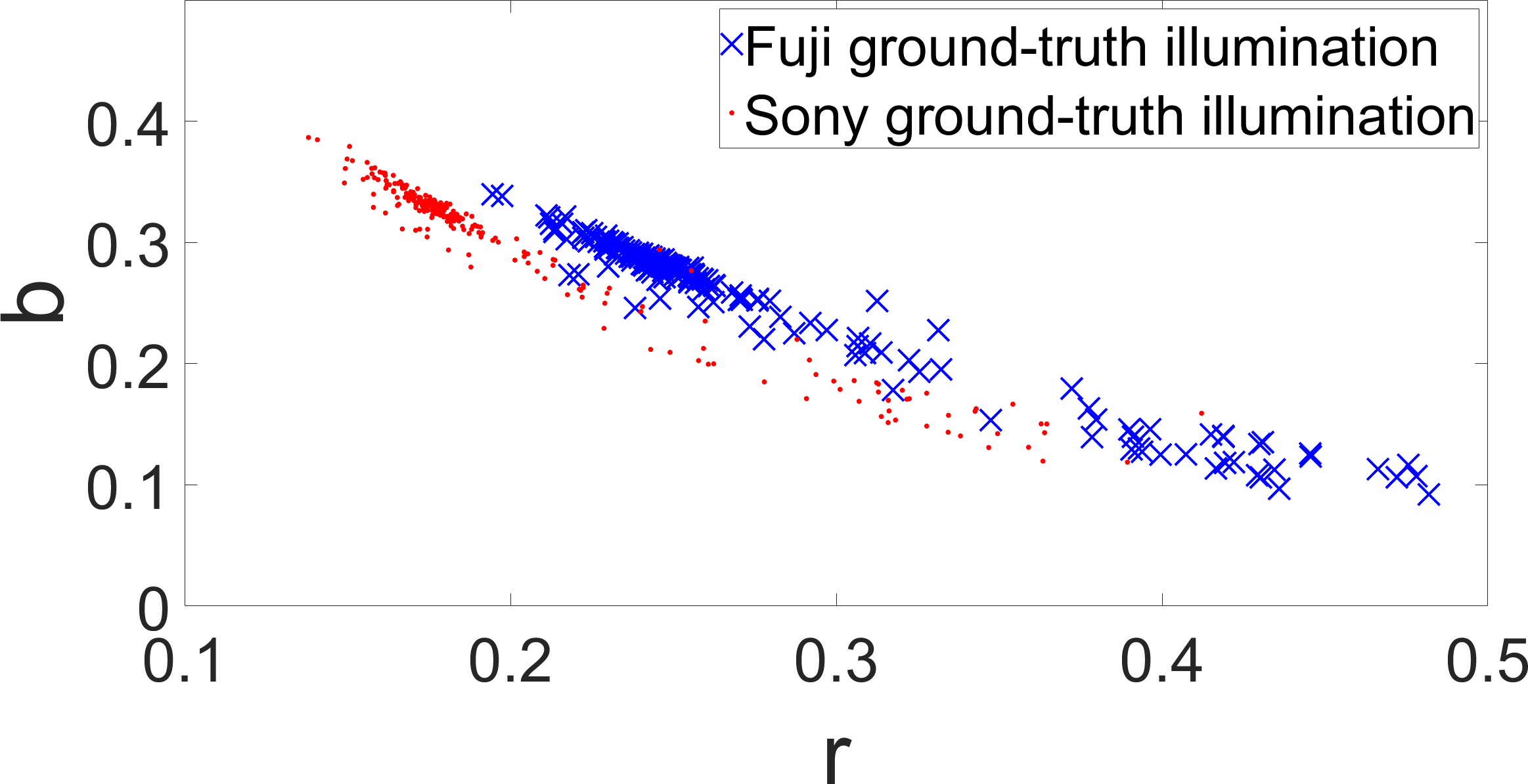}
	
    \caption{$rb$-chromaticities of ground-truth illuminations for Fuji and Sony datasets~\cite{cheng2014illuminant}.}
	\label{fig:fuji_sony}
    
\end{figure}

If the original Color Tiger method is trained and later used on images taken with a different sensor, its accuracy will most probably decrease. The reason is that camera spectral sensitivities affect the color domains of the images and illuminations~\cite{gao2017improving}. Because of this the ground-truth illuminations for real-world images taken with different cameras often occupy different regions in the chromaticity space as shown in Fig.~\ref{fig:fuji_sony}. Therefore, clustering centers learned on approximated ground-truths from one sensor will probably not be aligned with the modes of ground-truth illumination from the other sensor. For the Color Tiger method to work in such circumstances, the used colors from the training images and the used colors from the images to which the method is supposed to be applied have to be brought into the same colorspace. Transforming a raw pixel color $\mathbf{e}$ from an image taken with a given camera sensor that has its own linear RGB colorspace to pixel color $\mathbf{e}'$ in a different RGB colorspace can be modelled as~\cite{kim2012new}
\begin{equation}
\label{eq:transfer}
\mathbf{T}_s\mathbf{T}_w\mathbf{e}=\mathbf{e}'
\end{equation}
where $\mathbf{T}_w$ is a diagonal $3\times 3$ white balancing matrix of the \textit{von Kries model}~\cite{kries1902theoretische} that removes the illumination influence and $\mathbf{T}_s$ is a $3\times 3$ matrix that performs the transformation from the initial camera's RGB colorspace to another RGB colorspace, e.g. the linear sRGB colorspace. In practice, although the three simple multiplicative sensor gains $g_r, g_g, g_b$ for the red, green, and blue channel, respectively, are supposed to be incorporated into $\mathbf{T}_s$, because of their multiplicative nature they are by definition picked up by statistics-based illumination estimation methods such as Gray-world or Shades-of-Gray. For example, if an image of a prefectly white wall was taken under perfectly white illumination, it would not appear white, but most probably greenish because of the usual dominance of the green gain~\cite{kim2012new}. Therefore, if statistics-based methods are used, the sensor channel gains are already included in $\mathbf{T}_w$, which can then be written as $\mathbf{T}_w=\mathbf{T}'_w\mathbf{G}^{-1}=\mathbf{G}^{-1}\mathbf{T}'_w$ where $\mathbf{T}'_w$ now focuses only on illumination and $\mathbf{G}=\mathop{\mathrm{\mathbf{diag}}}(g_r, g_g, g_b)$. Since the gains are now excluded from $\mathbf{T}_s$, it is changed to $\mathbf{T}'_s=\mathbf{T}_s\mathbf{G}^{-1}$ so that $\mathbf{T}_s\mathbf{T}_w=\mathbf{T}'_s\mathbf{G}\mathbf{G}^{-1}\mathbf{T}'_w=\mathbf{T}'_s\mathbf{T}'_w$. Matrix $\mathbf{T}_s$ deals only with the differences between the three chromaticities of the red, green, and blue additive primaries of the camera RGB colorspace and the target RGB colorspace. Matrix $\mathbf{T}_s$ can be calculated by various techniques, but they require additional calibration. Eq.~\eqref{eq:transfer} can now be rewritten as
\begin{equation}
\label{eq:transfer2}
\mathbf{T}_s\mathbf{T}'_w\mathbf{G}^{-1}\mathbf{e}=\mathbf{e}'.
\end{equation}
In the case of Color Tiger, values of $\mathbf{e}$ that need to be transformed into $\mathbf{e}'$ are illumination estimations i.e. assumed white colors under the given scene illumination. Since this illumination information is needed for a successful clustering, it has to be preserved by avoiding white balancing. This can be achieved by simply dropping $\mathbf{T}'_w$ from the equation to get
\begin{equation}
\label{eq:transfer3}
\mathbf{T}_s\mathbf{G}^{-1}\mathbf{e}=\mathbf{e}^{*}
\end{equation}
where $\mathbf{e}^{*}$ is color $\mathbf{e}$ transformed to the target RGB colorspace, but without removing the illumination influence present in the original scene. Because white has equal values of red, green, and blue channels in any RGB colorspace, it is not affected by $\mathbf{T}_s$. Since the term $\mathbf{G}^{-1}\mathbf{e}$ represents the illumination color, it is not supposed to deviate significantly from white and under this assumption matrix $\mathbf{T}_s$ should not have a high impact on $\mathbf{G}^{-1}\mathbf{e}$. For this reason Eq.~\eqref{eq:transfer3} can be approximated as
\begin{equation}
\label{eq:transfer4}
\mathbf{G}^{-1}\mathbf{e}=\mathbf{e}^{+}\approx \mathbf{e}^{*}.
\end{equation}

If, on the other hand, $\mathbf{G}^{-1}\mathbf{e}$ is a color that significantly differs from white, using Eq.~\eqref{eq:transfer4} becomes inappropriate since the differences between cameras become high enough to e.g. resolve camera metamers~\cite{prasad2016strategies} or improvise hyperspectral imaging~\cite{wug2016yourself}. It is important to stress this in order to discourage potential use of Eq.~\eqref{eq:transfer4} in such and similar cases.

The only problem remaining is obtaining the value of $\mathbf{G}$ without any calibration or supervised learning. In the spirit of the Gray-world assumption, for the purpose of extracting $\mathbf{G}$ it is going to be assumed that the mean value of ground-truth illuminations of real-world images should be white i.e. that the reddish and blueish illumination biases should on average cancel out each other. If under this assumption the mean value differs from white, this can be directly attributed to sensor gains i.e. the mean contains the values of $g_r, g_g$, and $g_b$. However, instead of the ground-truth illuminations only their approximations i.e. illumination estimations are available for the images from the training set. In the particular case this can be the set of combined Shades-of-Gray illumination estimations obtained for $p\in\{1, 2, ..., n\}$ and denoted as $\mathbb{E}$ in Algorithm~\ref{alg:training}. Instead of using the trimmed set $\mathbb{E}'$, a better way for handling the outliers in $\mathbb{E}$ is to use median instead of mean. Therefore, the values of sensor gains are estimated as
\begin{equation}
\label{eq:gains_r}
g_r=\underset{\mathbf{e}^{(i)} \in \mathbb{E}}{\mathrm{median}} \{ e_r^{(i)} \},
\end{equation}
\begin{equation}
\label{eq:gains_g}
g_g=\underset{\mathbf{e}^{(i)} \in \mathbb{E}}{\mathrm{median}} \{ e_g^{(i)} \},
\end{equation}
\begin{equation}
\label{eq:gains_b}
g_b=\underset{\mathbf{e}^{(i)} \in \mathbb{E}}{\mathrm{median}} \{ e_b^{(i)} \}.
\end{equation}

\begin{figure}[htb]
    \centering
    
	\includegraphics[width=\linewidth]{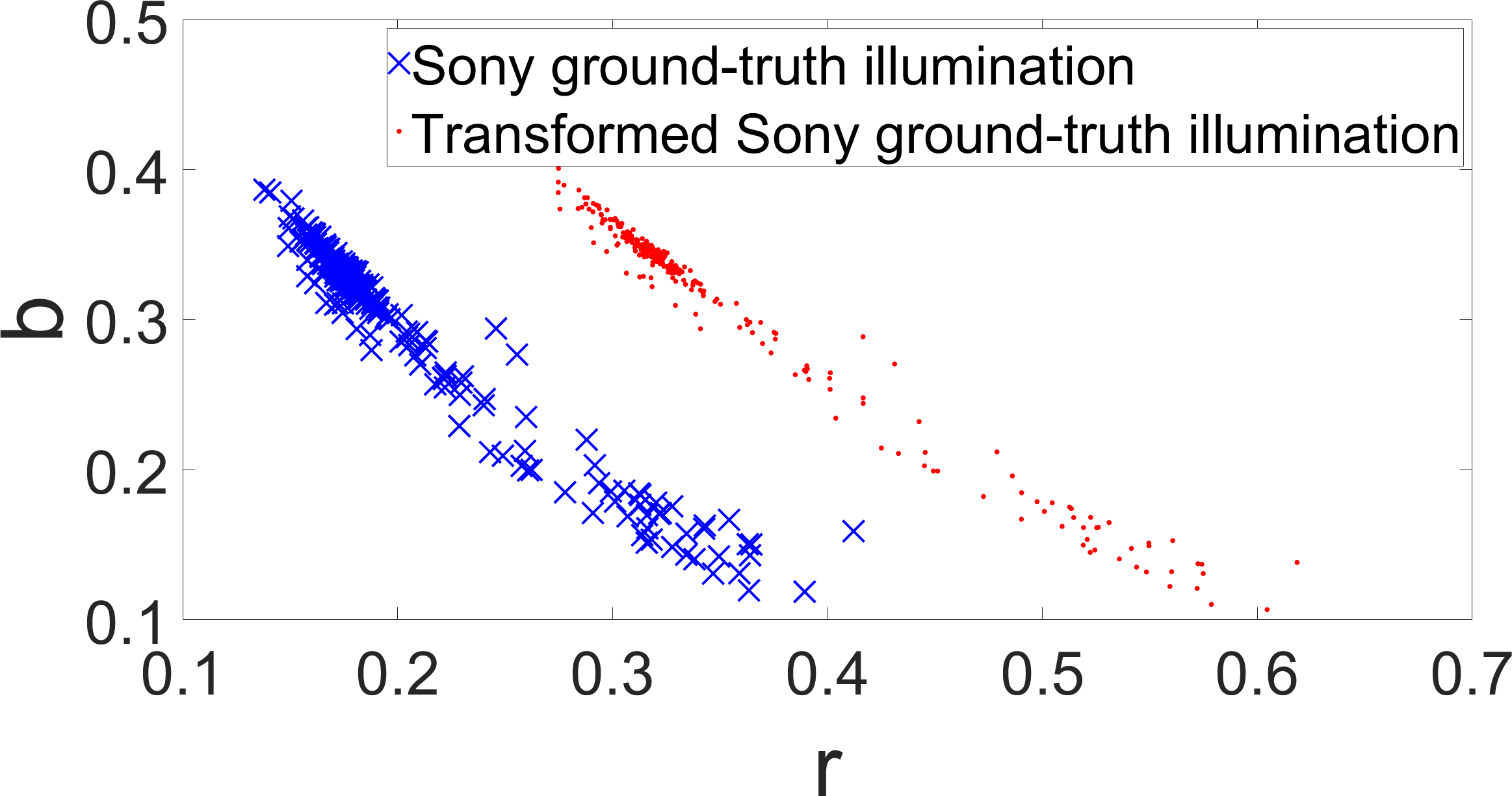}
	
    \caption{$rb$-chromaticities of the original and transformed ground-truth illuminations for Sony dataset~\cite{cheng2014illuminant}.}
	\label{fig:transformation}
    
\end{figure}

With these estimations of elements of $\mathbf{G}$, Eq.~\eqref{eq:transfer4} can now be used to transform colors of each $\mathbf{e}\in\mathbb{E}$ to a neutral RGB colorspace. An example of such transformation of real ground-truth illuminations is shown in Fig.~\ref{fig:transformation}. It can be seen that effectively the illumination chromaticities have been shifted in the chromaticity space without any other significant changes. The procedure for learning $\mathbf{G}$ is summarized in Algorithm~\ref{alg:g}.

\begin{algorithm}
\caption{Matrix $\mathbf{G}$ Learning}
\label{alg:g}
\hspace*{\algorithmicindent}\textbf{Input:} images $\mathbb{I}$, SoG upper power $n$\\
\hspace*{\algorithmicindent}\textbf{Output:} matrix $\mathbf{G}$
\begin{algorithmic}[1]

\State $\mathbb{E}\gets\{\}$

\For{$\mathbf{I}\in\mathbb{I}$}
	\For{$p\in\{1, 2, ..., n\}$}
		\State $\mathbf{e}=ShadesOfGray(\mathbf{I}, p)$ \Comment Apply Eq.~\eqref{eq:sog}
		\State $\mathbb{E}\gets\mathbb{E}\cup\{\mathbf{e}\}$
	\EndFor
\EndFor

\State $g_r=\underset{\mathbf{e}^{(i)} \in \mathbb{E}}{\mathrm{median}} \{ e_r^{(i)} \}$
\State $g_g=\underset{\mathbf{e}^{(i)} \in \mathbb{E}}{\mathrm{median}} \{ e_g^{(i)} \}$
\State $g_b=\underset{\mathbf{e}^{(i)} \in \mathbb{E}}{\mathrm{median}} \{ e_b^{(i)} \}$

\State $\mathbf{G}=\mathop{\mathrm{\mathbf{diag}}}(g_r, g_g, g_b)$

%\State $SetOutputImage(R)$

\end{algorithmic}

\end{algorithm}

Besides learning values of $\mathbf{G}$ for the sensor used to create the training image, another $\mathbf{G}'$ has to be learned for the target sensor used to create images to which the extended Color Tiger is applied in order to neutralize its sensor gains. This can be simply done by applying the procedure in Algorithm~\ref{alg:g} to a set of images taken with the target sensor. The size of this set is not supposed to be the same as the size of training images used to learn $\mathbf{G}$ and the clustering centers; otherwise it would be more efficient to simply apply the original Color Tiger instead of its extended version. The influence of this size on the overall accuracy is examined further in Section~\ref{sec:inter}.

Since the proposed extended Color Tiger method is more flexible and applicable to a wider set of data than the original Color Tiger method, it is named Color Bengal Tiger~(CBT). The procedures for training and applying Color Bengal Tiger are summarized in Algorithm~\ref{alg:training2} and~\ref{alg:application2}, respectively.

\begin{algorithm}
\caption{Color Bengal Tiger Training}
\label{alg:training2}
\hspace*{\algorithmicindent}\textbf{Input:} images $\mathbb{I}$, target sensor images $\mathbb{I}'$, SoG upper power $n$, trimming $t$\\
\hspace*{\algorithmicindent}\textbf{Output:} gain matrices $\mathbf{G}$ and $\mathbf{G}'$, set of two centers $\mathbb{C}$
\begin{algorithmic}[1]

\State $\mathbf{G}=MatrixGLearning(\mathbb{I}, n)$ \Comment Algorithm~\ref{alg:g}
\State $\mathbf{G}'=MatrixGLearning(\mathbb{I}', n)$ \Comment Algorithm~\ref{alg:g}
\State $\mathbb{E}=\{\}$

\For{$\mathbf{I}\in\mathbb{I}$}
	\For{$p\in\{1, 2, ..., n\}$}
		\State $\mathbf{e}=ShadesOfGray(\mathbf{I}, p)$ \Comment Apply Eq.~\eqref{eq:sog}
		\State $\mathbb{E}\gets\mathbb{E}\cup\{\mathbf{G}^{-1}\mathbf{e}\}$
	\EndFor
\EndFor

\State $\mathbb{E}'=Trimming(\mathbb{E}, 2, t)$ \Comment Algorithm~\ref{alg:trimming}
\State $\mathbb{C}=kmeans(\mathbb{E}', 2)$ \Comment Use angular distance

%\State $SetOutputImage(R)$

\end{algorithmic}

\end{algorithm}

\begin{algorithm}
\caption{Color Bengal Tiger Application}
\label{alg:application2}
\hspace*{\algorithmicindent}\textbf{Input:} image $\mathbf{I}$, gain matrix $\mathbf{G}'$, set of two centers $\mathbb{C}$\\
\hspace*{\algorithmicindent}\textbf{Output:} illumination estimation $\mathbf{e}$
\begin{algorithmic}[1]

\State $\mathbf{e}_{GW}=\mathbf{G}'^{-1}GrayWorld(\mathbf{I})$ \Comment Apply Eq.~\eqref{eq:gw}

\State $\mathbf{e}_{WP}=\mathbf{G}'^{-1}WhitePatch(\mathbf{I})$ \Comment Apply Eq.~\eqref{eq:wp}

\State $\mathbf{e}=\mathbf{G}'\underset{\mathbf{c}_i \in \mathbb{C}}{\operatorname{\arg\max}} \left( \frac{\mathbf{c}_i \cdot \mathbf{e}_{GW}}{||\mathbf{c}_i || \cdot || \mathbf{e}_{GW}||} + \frac{\mathbf{c}_i \cdot \mathbf{e}_{WP}}{||\mathbf{c}_i || \cdot || \mathbf{e}_{WP}||} \right)$

%\State $SetOutputImage(R)$

\end{algorithmic}

\end{algorithm}

It must be noted that obtaining accurate values for elements of $\mathbf{G}$ for a pair of camera sensors is also possible by simply using the information extracted after taking a single image per camera where each image contains a calibration object. Nevertheless, since the topic of this paper is unsupervised learning, one of the goals here is also to examine whether or not at to what degree $\mathbf{G}$ can be learned without any calibration.

\section{General applicability}
\label{sec:applicability}

A potential concern about the proposed methods is related to their general applicability to images with illuminations that considerably differ from the average illumination in the tested datasets. Namely, in both cases the proposed methods use only two colors to cover all possible illuminations. While this really limits the flexibility of the proposed methods, this is actually a desirable feature often used in the industry and not a bug.

\begin{figure}[htb]
    \centering
    
	\includegraphics[width=\linewidth]{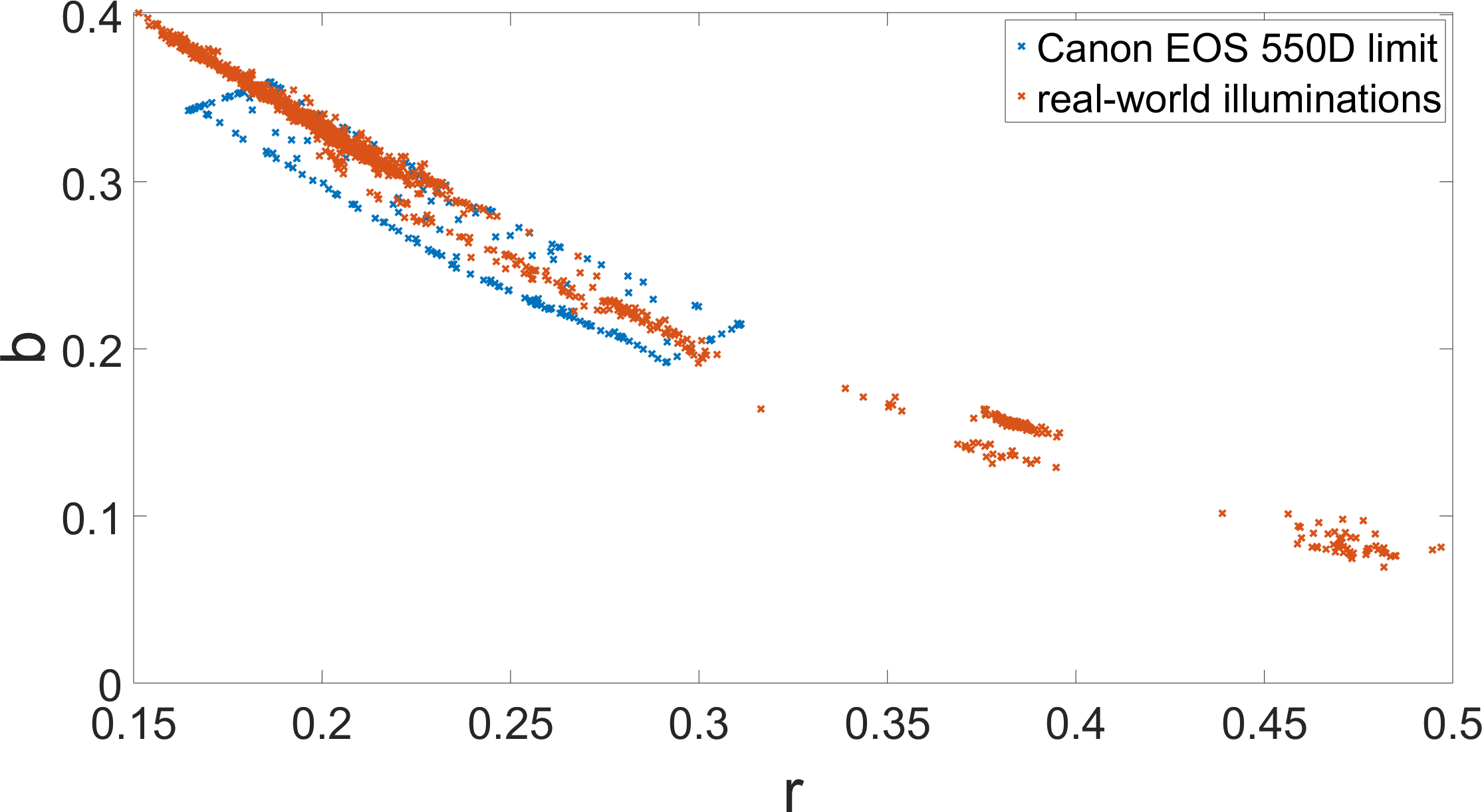}
	
    \caption{Comparison of the locations of real-world illuminations and the camera's built-in limitation of its illumination estimations in the $rb$-chromaticity plane.}
	\label{fig:limit}
    
\end{figure}

A good starting point to explain why this is so is to mention the experiment conducted in~\cite{koscevic2019color} where several Canon camera models were used to take images influenced by several hundreds illuminations. It has been shown how Canon cameras simply restrict most illumination colors to a polygon that encloses the most commonly observed illuminations. This means that even if a method in the camera would estimate the illumination to be e.g. pure red, the camera would in the chromaticity plane still put this estimation closer to the most commonly observed illuminations. An illustration of this is given in Fig.~\ref{fig:limit} where the ground-truth illuminations from the benchmark dataset described in Section~\ref{subsec:cube+} are plotted together with the polygon that limits the illumination estimations of the very same Canon EOS 550D camera that was used to create the mentioned benchmark dataset.

It can be seen that there the camera allows its illumination estimations to fall into only a relatively small part of the whole chromaticity plane that does not cover many of the relatively common illuminations from the real-world. There are at least two reasons that explain the benefits of such a behavior.

First, this prevents an illumination estimation to erroneously choose a highly unlikely illumination. For example, if an image consists only of purple pixels, due to the ill-posedness of the illumination estimation problems there are infinitely many explanations for such a scene. Two of them include e.g. a purple wall under the white illumination and white wall under the purple illumination. The white illumination occurs more often than the purple one thus also making it more likely to be influencing the mentioned scene. In practice, methods such as Gray-world would be heavily leaned towards the purple illumination estimation and this is where restrictions like the on in Canon cameras perform a certain damage control.

Second, even if the illumination estimation correctly predicts a highly saturated illumination color, the quality of fully correcting such an illumination is not necessary of the highest quality and thus also not always desirable. Namely, because of the model described by Eq.~\eqref{eq:image}, a highly saturated illumination color will result in reflectance of different color not to be fully brought to visibility. This means that fully correcting the colors in such cases can result in them looking washed out after the correction, which can justify the absence of full correction that occurs when the describes camera limitations are present.

Having only two illumination centers in the proposed methods also serves as a restriction that prevents the occurrence of highly unlikely illumination estimations that can happen with the Gray-world and White-patch methods. Therefore, having only two centers can be considered a kind of an advantage.

%------------------------------------------------------------------------

\begin{figure*}[htb]
	\captionsetup[subfloat]{labelformat=empty}
    \centering
    
	\begin{subfloat}[]{}
		\includegraphics[width=0.16\linewidth]{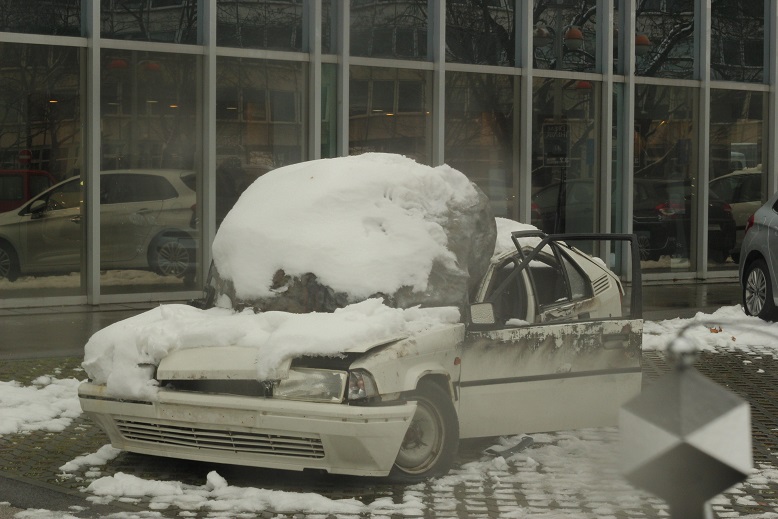}
		\includegraphics[width=0.16\linewidth]{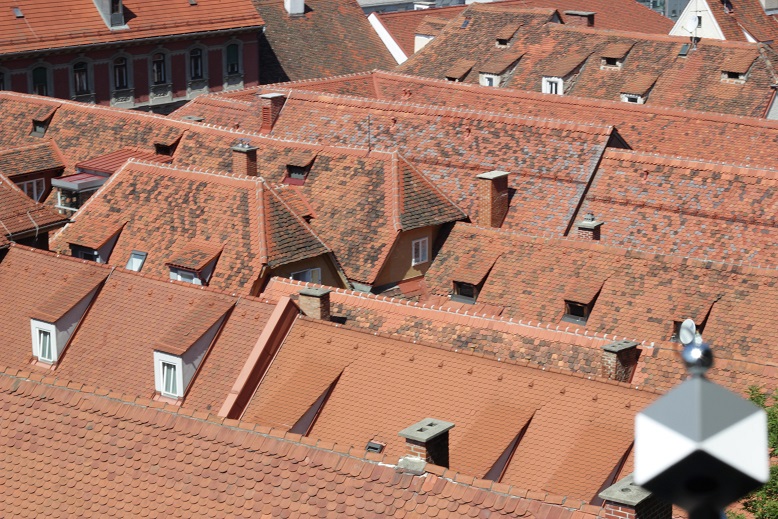}
		\includegraphics[width=0.16\linewidth]{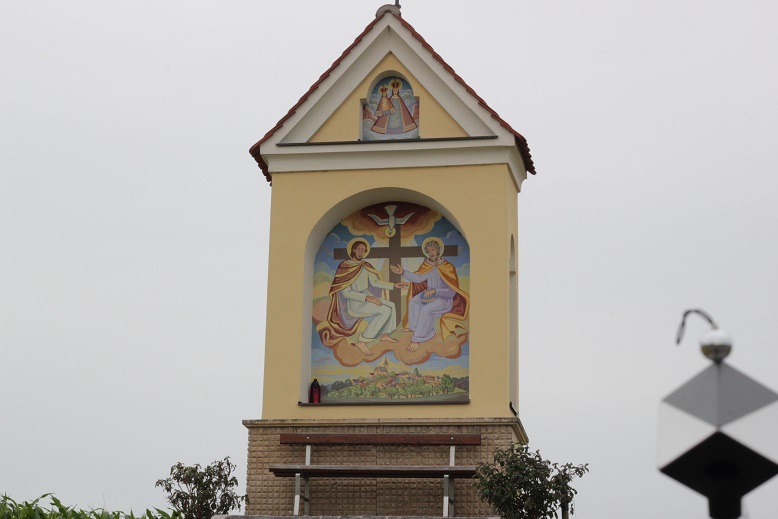}
		\includegraphics[width=0.16\linewidth]{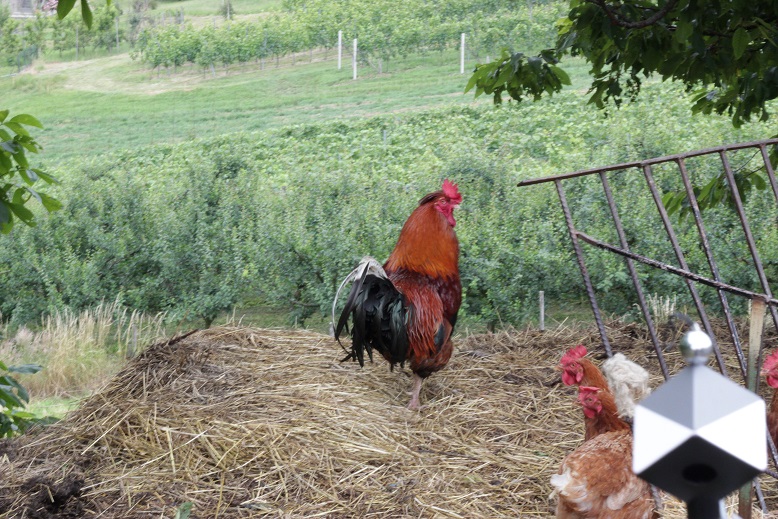}
		\includegraphics[width=0.16\linewidth]{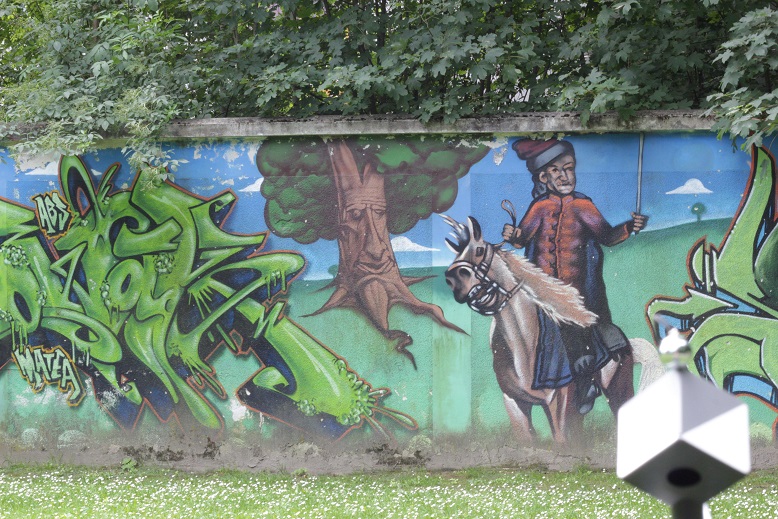}
		\includegraphics[width=0.16\linewidth]{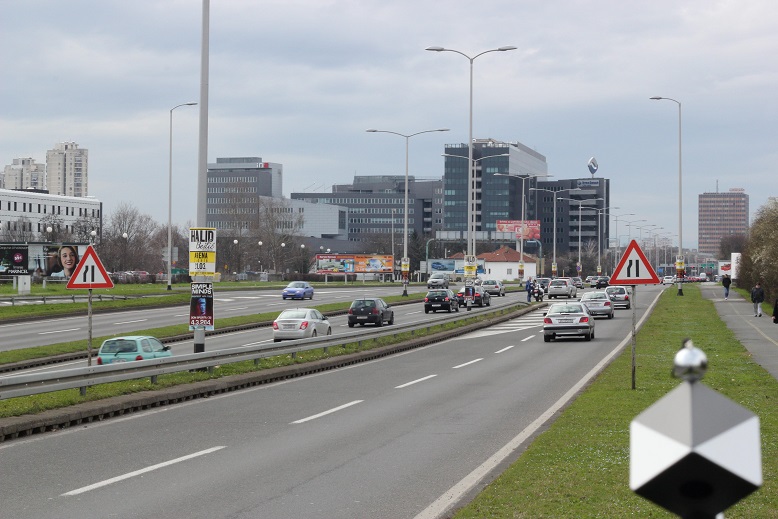}
	\end{subfloat}
	\begin{subfloat}[]{}
		\includegraphics[width=0.16\linewidth]{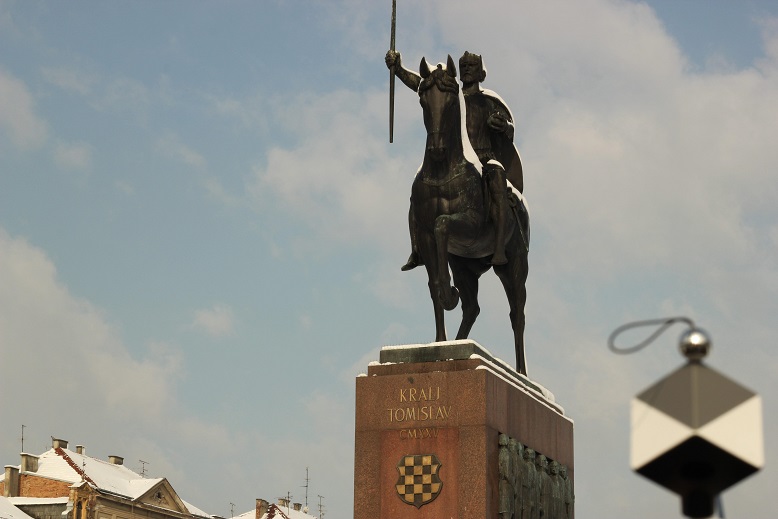}
		\includegraphics[width=0.16\linewidth]{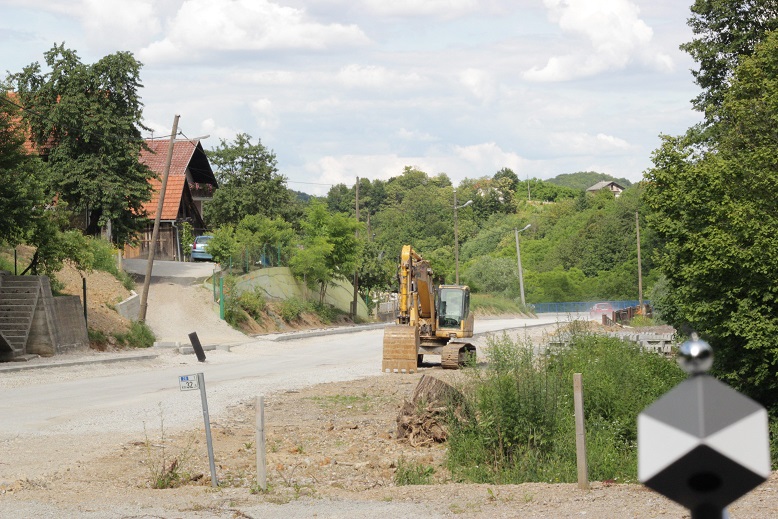}
		\includegraphics[width=0.16\linewidth]{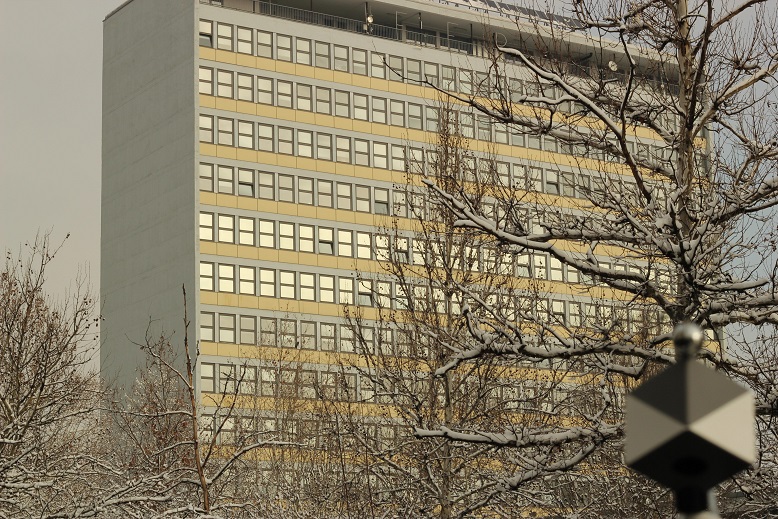}
		\includegraphics[width=0.16\linewidth]{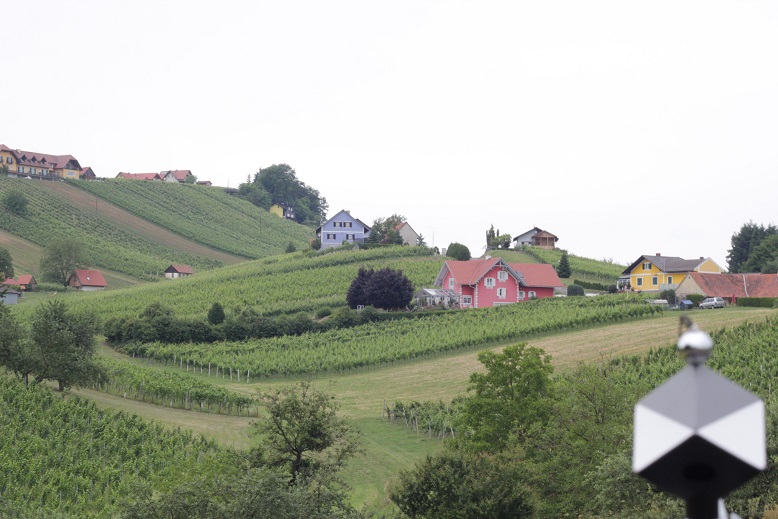}
		\includegraphics[width=0.16\linewidth]{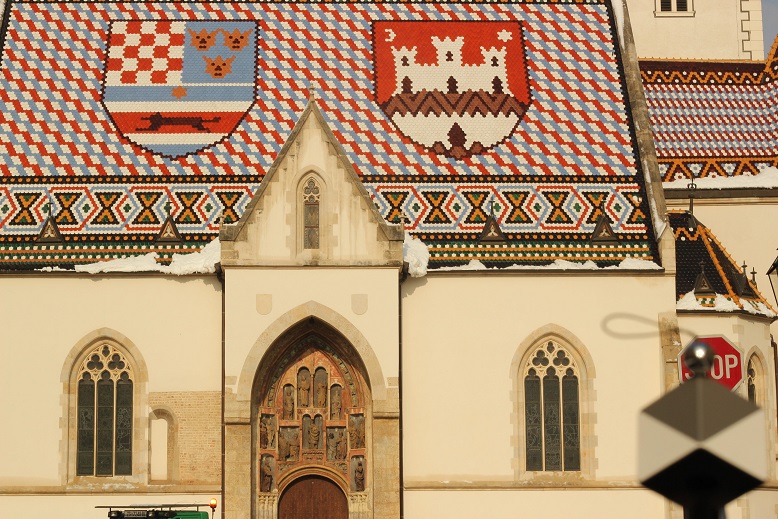}
		\includegraphics[width=0.16\linewidth]{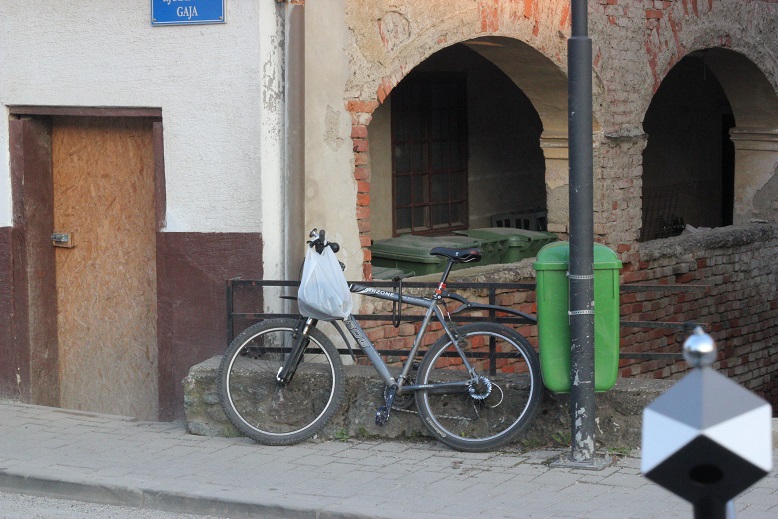}
	\end{subfloat}
	\begin{subfloat}[]{}
		\includegraphics[width=0.16\linewidth]{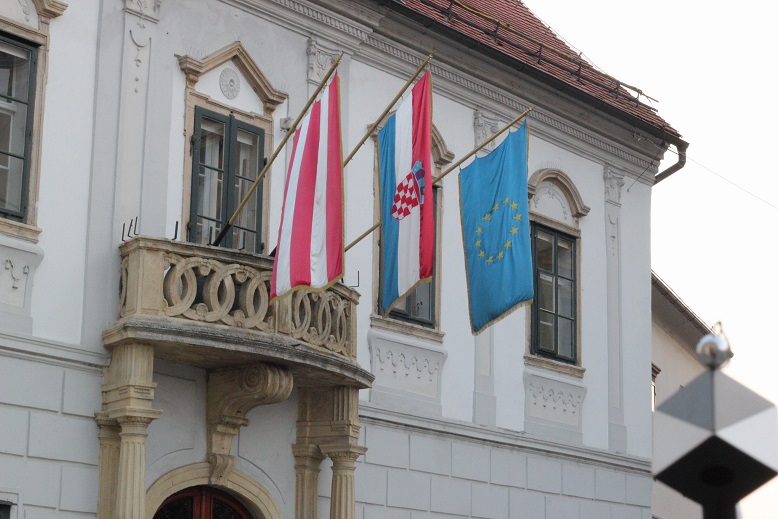}
		\includegraphics[width=0.16\linewidth]{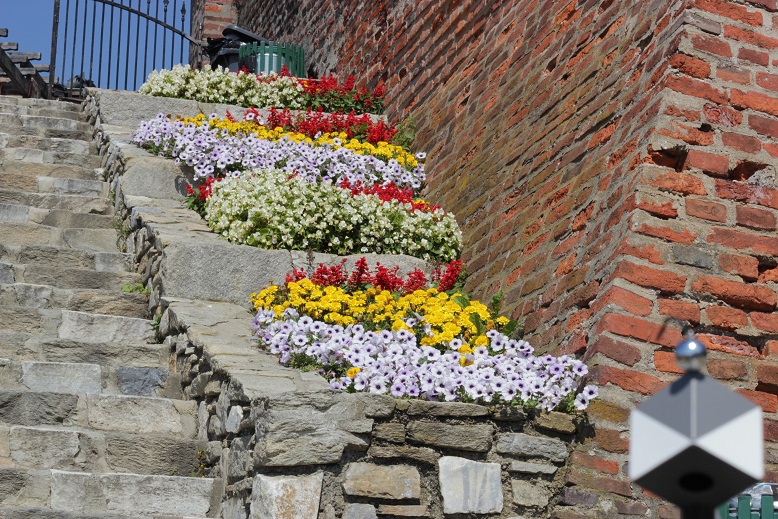}
		\includegraphics[width=0.16\linewidth]{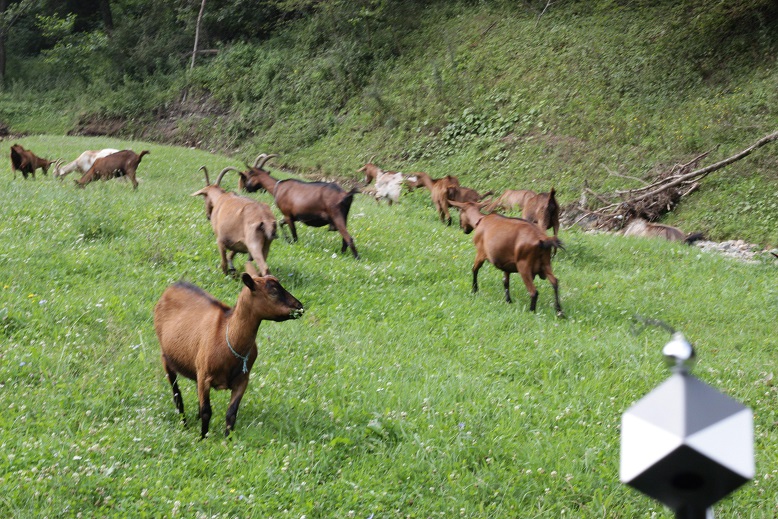}
		\includegraphics[width=0.16\linewidth]{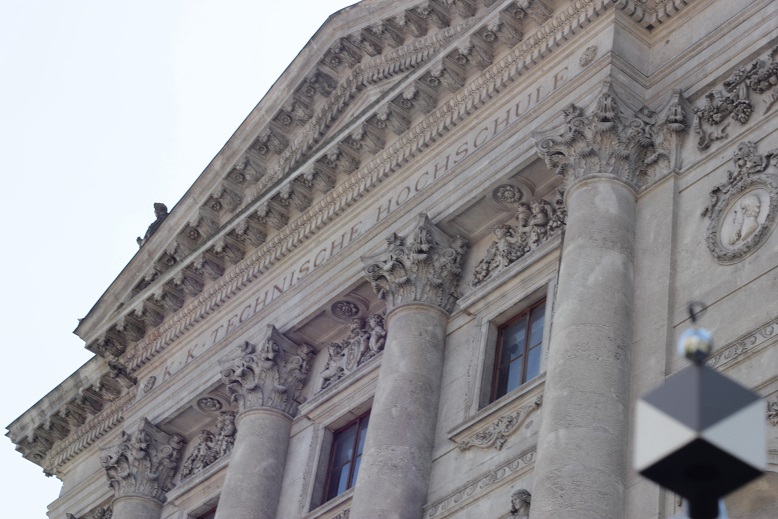}
		\includegraphics[width=0.16\linewidth]{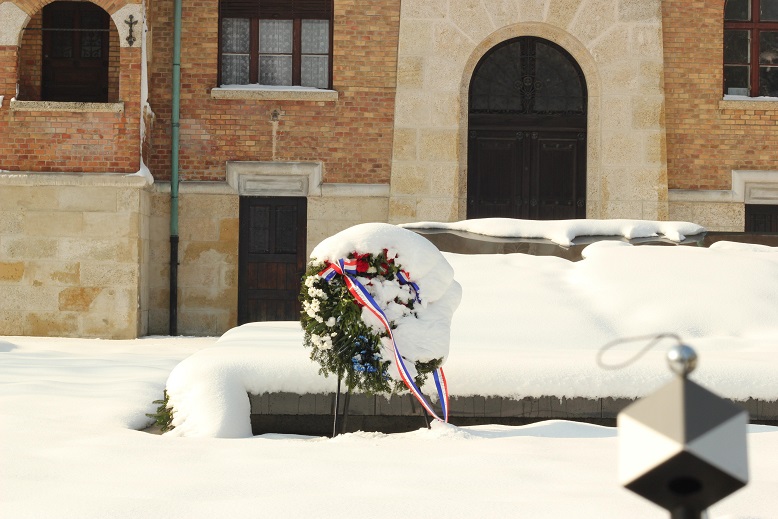}
		\includegraphics[width=0.16\linewidth]{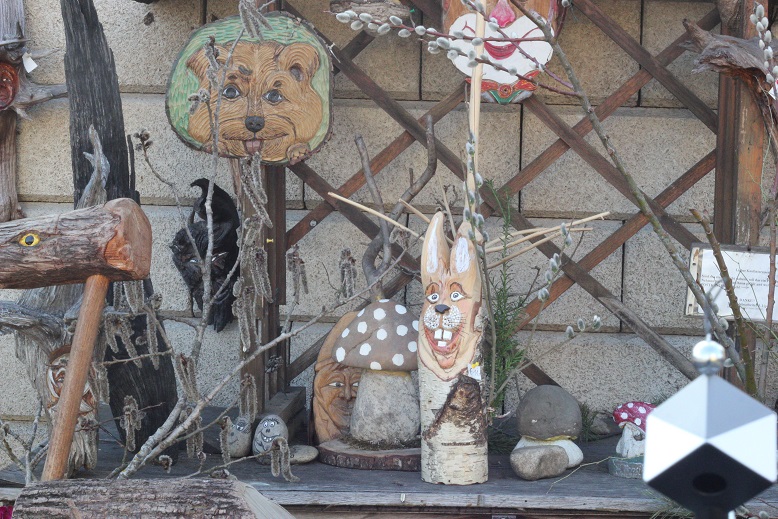}
	\end{subfloat}\\
	
    \caption{Example images from the newly created Cube dataset.}
	\label{fig:cube}
    
\end{figure*}

\section{Experimental results}
\label{sec:results}

\subsection{Experimental setup}
\label{subsec:setup}

The following benchmark datasets have been used to compare the accuracy of the proposed method to the accuracy of other well-known methods: the GreyBall dataset~\cite{ciurea2003large}, its approximated linear version, eight linear NUS dataset~\cite{cheng2014illuminant}, and a newly created dataset, which is presented in more detail in the following subsection. The ColorChecker dataset~\cite{gehler2008bayesian, shi2015online} has not been used to avoid confusion over different results mentioned in numerous publications during ColorChecker's history of various and wrong usage~\cite{lynch2013colour,finlayson2017curious,hemrit2018rehabilitating,banic2019past}. Since in digital devices illumination estimation is usually performed on linear images~\cite{kim2012new} i.e. images that have not been non-linearly processed and that are in compliance with the model described by Eq.~\eqref{eq:image}, datasets with linear images are usually preferred.

Each dataset has images and their ground-truth illuminations, which have been obtained by putting a calibration object in the image scene, e.g. a color checker or a gray ball. Before applying a method to a dataset image during the testing, the calibration object has to be masked out to avoid bias.

Various illumination estimation accuracy measures have been proposed~\cite{gijsenij2009perceptual, finlayson2014reproduction, banic2015perceptual}. The most commonly used one is the angular error i.e. the angle between the illumination estimation vector and the ground-truth illumination. All angular errors obtained for a given method on a chosen dataset are usually summarized by different statistics. Because of the non-symmetry of the angular error distribution, the most important of these statistics is the median angular error~\cite{hordley2004re}.

Cross-validation on the GreyBall and NUS dataset was performed with the same folds as in other publications. For the Cube and Cube+ datasets that are described in next subsections a three-fold cross-validation with folds of equal size was used. The source code for recreating the results given in one of the following subsections is publicly available at \url{http://www.fer.unizg.hr/ipg/resources/color_constancy/}.

\subsection{The Cube dataset}

\begin{figure}[htb]
    \centering
    
	\subfloat[]{
	\includegraphics[width=0.48\linewidth]{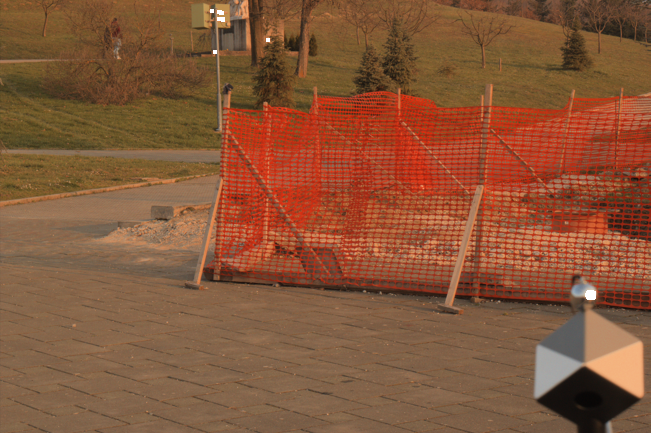}
	\label{fig:left}
	}%
	~%
	\subfloat[]{
	\includegraphics[width=0.48\linewidth]{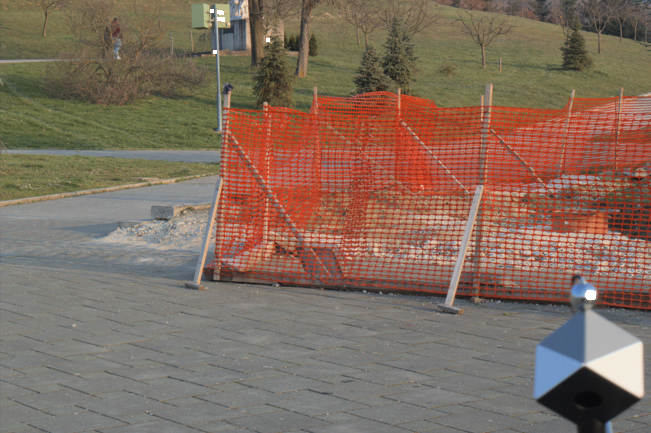}
	\label{fig:right}
	}
	
    \caption{Example of chromatic adaptation based on illumination extracted from a) left and b) right gray face of the SpyderCube calibration object placed in the scene.}
	\label{fig:two}
    
\end{figure}

The newly created dataset contains $1365$ exclusively outdoor images taken with a Canon EOS 550D camera in parts of Croatia, Slovenia, and Austria during various seasons and it is publicly available at \url{http://www.fer.unizg.hr/ipg/resources/color_constancy/}. The image ordering with respect to their creation time has been shuffled. In the lower right corner of each image the SpyderCube calibration object~\cite{cube2017online} is placed. Its two neutral 18\% gray faces were used to determine the ground-truth illumination for each image. Due to the angle between these two faces, for images with two illuminations, e.g. one in the shadow and one under the direct sunlight, it was possible to simultaneously recover both of them and they are provided for each image. This is especially important because it has been reported on multiple occasions~\cite{cheng2016two, qian2018dichromatic} that some of the previous dataset have either incorrectly labelled ground-truth and/or multiple illuminations influencing the images, which can potentially introduce unfair bias towards some methods. For all images in the Cube dataset with two distinct illuminations, one of them is always dominant so that the uniform illumination assumption effectively remains valid. To correctly identify the dominant illumination, for each image its two possible chromatically adapted versions were manually checked and after this has been done for all images, the final ground-truth illumination was created. In this way it was possible to eliminate any images that have problematic illumination conditions that violate the uniform illumination assumption and the accuracy of ground-truth labelling was increased as well. An example of conflicting ground-truths extracted from two gray faces of the SpyderCube calibration object is shown in Fig.~\ref{fig:two} with the angle between them being $9.79^\circ$. This clearly shows that when e.g. a color checker is used to extract the ground-truth illumination from similar scenes, it is very important how it will be placed in the scene i.e. what will be the angle between its achromatic patches and the dominant scene illumination. Its inappropriate placing is one of the reason of the previously mentioned incorrectly extracted ground-truth illuminations. With SpyderCube such potentially problematic scenes can be detected by measuring the angle between the ground-truths extracted from its two gray faces.

The black level, i.e. the intensity that has to be subtracted from all images in order to use them properly, equals $2048$. To make a conclusion about the maximum allowed intensity values of non-clipped pixels in the dataset images, histograms of intensities for various images were observed. If $m$ is the maximum intensity for a given dataset image in any of its channels, then the best practice is to discard all image pixels that have a channel intensity that is greater than or equal to $m-2$. Finally, before an image from the dataset is used to test the accuracy of an illumination estimation method, the calibration object has to be masked out to prevent a biased influence. A simple way to do this is to mask out the lower right rectangle starting at row $1050$ and column $2050$. Because of the used SpyderCube calibration object, the dataset is named Cube. Fig.~\ref{fig:cube} shows some images from the Cube dataset.

\begin{figure}[htb]
    \centering
    
	\includegraphics[width=\linewidth]{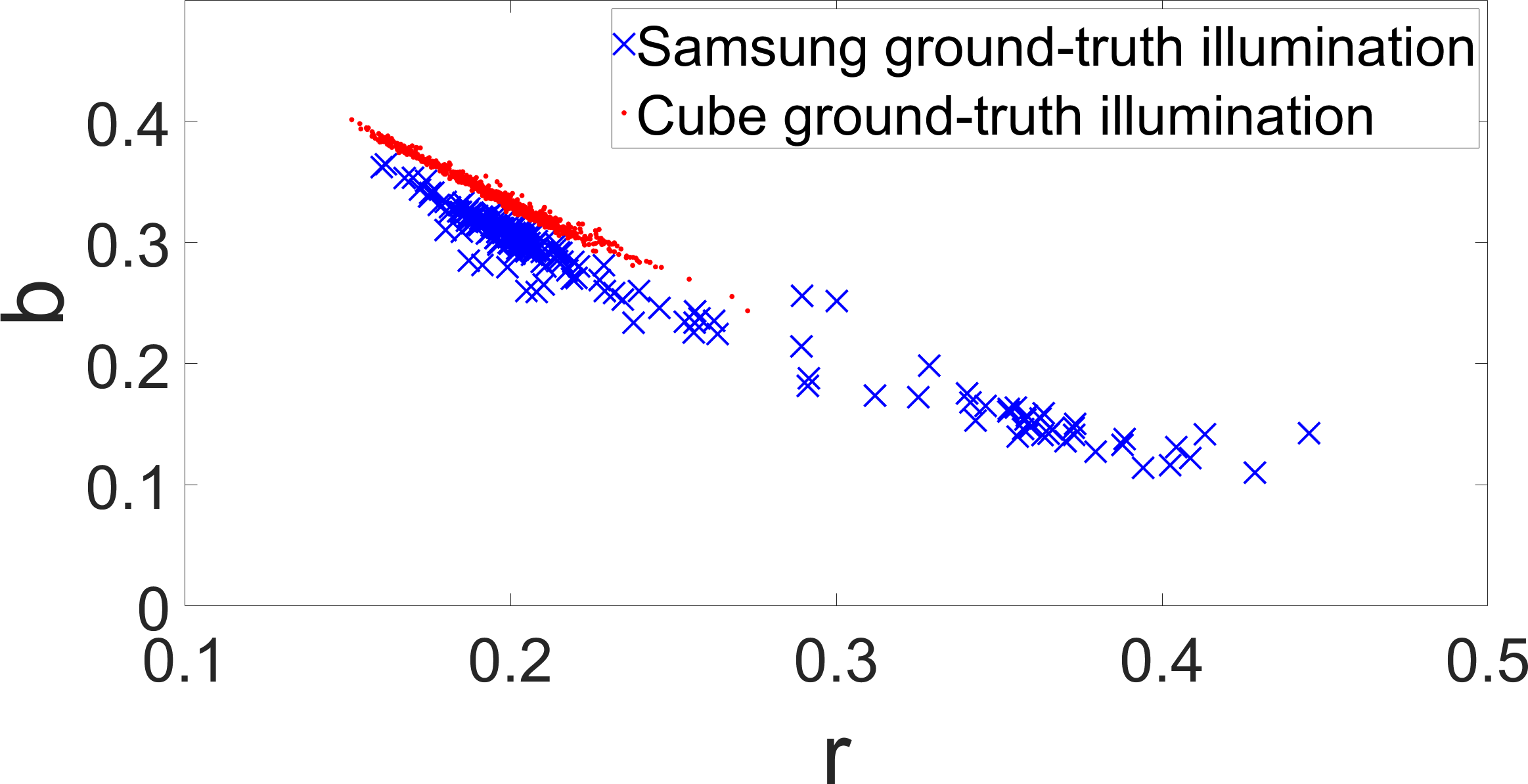}
	
    \caption{$rb$-chromaticities of ground-truth illuminations for Samsung~\cite{cheng2014illuminant} and Cube datasets.}
	\label{fig:samsung_cube}
    
\end{figure}

In addition to having data not only for one, but for two illuminations for each image, another advantage of the Cube dataset is that all its images were created using the same camera, which is very important for methods that require larger amounts of data such as deep learning methods. Namely, a single camera assures the same values for $\boldsymbol{\rho}(\lambda)$ in Eq.~\eqref{eq:e}, which is an important requirement when a method is trained. Additionally, the scenes in the Cube dataset have a much richer content than the ones in e.g. ColorChecker or NUS datasets. This is because they were taken across a much wider area spanning through three countries. On the other hand the diversity of the NUS dataset images is restricted since many images contain the same scene taken with a different camera.

The main disadvantage of the Cube dataset in comparison to other well-known color constancy benchmark datasets in that it contains only outdoor images. This has a significant impact on the distribution of its ground-truth illuminations since they contain only outdoor illuminations. Fig.~\ref{fig:samsung_cube} shows $rb$-chromaticities of ground-truth illuminations for Samsung~\cite{cheng2014illuminant} and Cube datasets. For the Cube dataset there is a clear lack of warmer i.e. reddish indoor illuminations. This is an obvious violation of the two illuminations assumption and when applied to images of the Cube dataset, Color Tiger's two clustering centers will end up dividing what was supposed to be a single well-defined cluster. With such division it is highly probable that the mode of the outdoor illuminations will be missed by both Color Tiger's clustering centers, which will in turn result in lower illumination estimation accuracy. In cases like this when it is known that there is only a single illumination type, the two illumination assumption will fail. A more appropriate method then would be e.g. the Color Mule method that always returns the same single illumination~\cite{banic2015perceptual}.

\begin{figure}[htb]
    \centering
    
	\includegraphics[width=\linewidth]{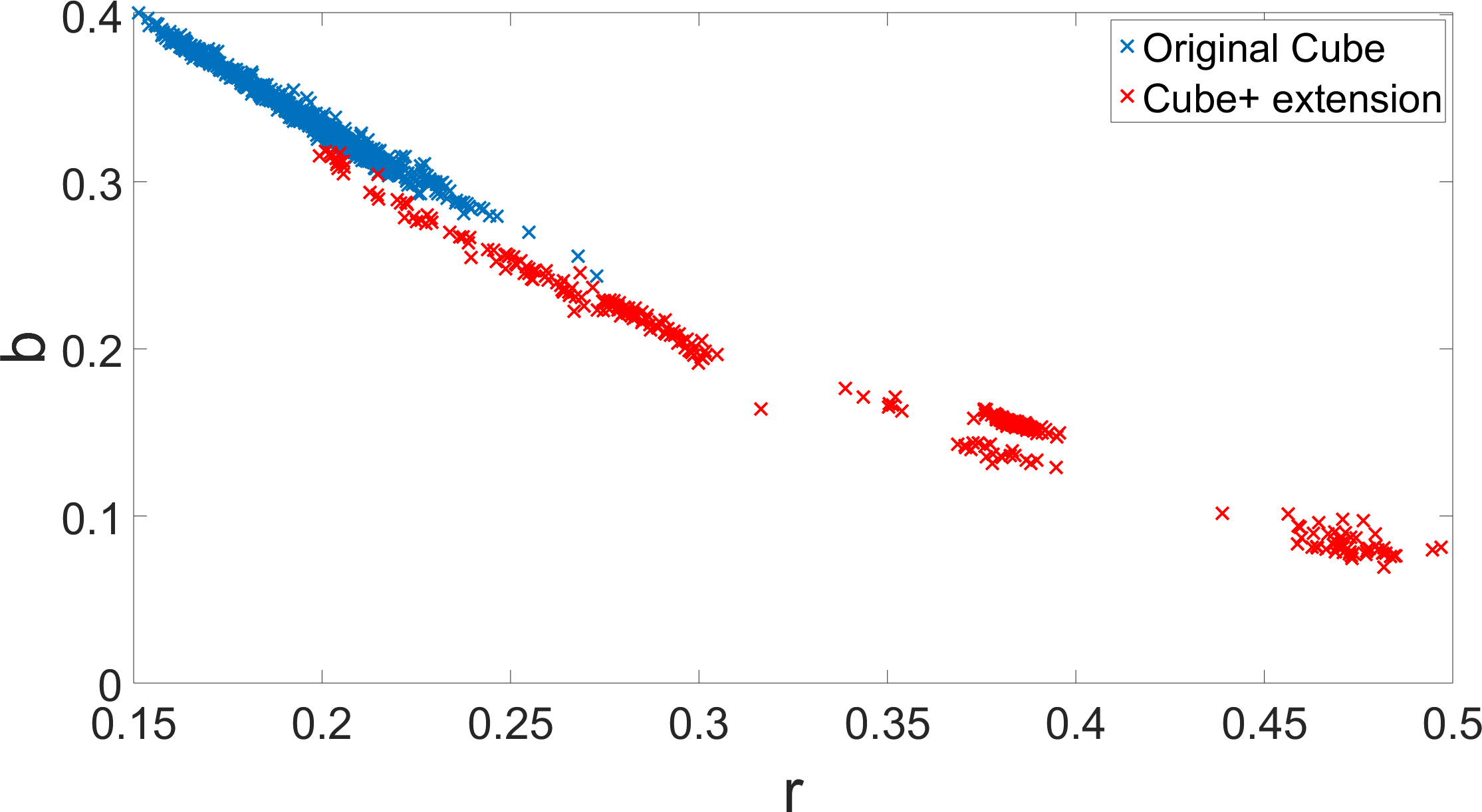}
	
    \caption{The comparison of the ground-truth illuminations in the Cube dataset to the ones newly added in the extended Cube+ dataset shown in the $rb$-chromaticity plane.}
	\label{fig:extension}
    
\end{figure}

\subsection{The Cube+ dataset}
\label{subsec:cube+}

In order to try to alleviate the illumination distribution problem of the Cube dataset, additional $342$ images under much warmer i.e. reddish illuminations were taken with the same camera that was used to create the Cube dataset. The ground-truth extraction procedure was the same as for the Cube dataset. Together with the original Cube dataset images, the combined dataset now consists of $1707$ images with around a fifth of them being the new ones with the described illumination properties. This makes the whole illumination distribution similar to the one in e.g. the NUS datasets.

Since the arXiv paper version where the Cube dataset was first published has already been used by other researchers, the new images were not simply included into the Cube dataset. Instead, the new combination of the Cube dataset images and the newly created images is now the new Cube+ dataset. Beside having a large number of images taken by using a single camera, the Cube+ dataset also includes both indoor and outdoor images taken during the night. The comparison of the ground-truth illuminations in the Cube dataset to the ones newly added in the Cube+ dataset is shown in Fig.~\ref{fig:extension}.

\begin{table}[ht]
%\normalsize
\scriptsize
%\tiny
\caption{Combined performance of different color constancy methods on eight NUS dataset (lower Avg. is better). The used format is the same as in~\cite{barron2015convolutional}.}
\label{tab:nus}
\centering
%\begin{tabular}{|c|p{2mm}p{2mm}p{2mm}p{3mm}p{3mm}|p{2mm}|}
\begin{tabular}{|>{\centering}m{3.75cm}| >{\centering}m{3mm} >{\centering}m{3mm} >{\centering\arraybackslash}m{3mm} >{\centering\arraybackslash}m{3mm} >{\centering\arraybackslash}m{3mm} >{\centering\arraybackslash}m{3mm}|}

    \hline
    \textbf{Algorithm} & \textbf{Mean} & \textbf{Med.} & \textbf{Tri.} & \makecell{\textbf{Best}\\\textbf{25\%}} & \makecell{\textbf{Worst}\\\textbf{25\%}} & \textbf{Avg.}\\
    
    \hline
    
    White-Patch~\cite{funt2010rehabilitation} & 10.62 & 10.58 & 10.49 & 1.86 & 19.45 & 8.43\\
    Edge-based Gamut~\cite{barnard2000improvements} & 8.43 & 7.05 & 7.37 & 2.41 & 16.08 & 7.01\\
    Pixel-based Gamut~\cite{barnard2000improvements} & 7.70 & 6.71 & 6.90 & 2.51 & 14.05 & 6.60\\
    Intersection-based Gamut~\cite{barnard2000improvements} & 7.20 & 5.96 & 6.28 & 2.20 & 13.61 & 6.05\\
    Gray-world~\cite{buchsbaum1980spatial} & 4.14 & 3.20 & 3.39 & 0.90 & 9.00 & 3.25\\
    Color Mule~\cite{banic2015perceptual} & 5.58 & 1.85 & 2.67 & 0.53 & 17.19 & 3.02\\
    Bayesian~\cite{gehler2008bayesian} & 3.67 & 2.73 & 2.91 & 0.82 & 8.21 & 2.88\\
    Natural Image Statistics~\cite{gijsenij2011color} & 3.71 & 2.60 & 2.84 & 0.79 & 8.47 & 2.83\\
    Shades-of-Gray~\cite{finlayson2004shades} & 3.40 & 2.57 & 2.73 & 0.77 & 7.41 & 2.67\\
    %Local Surface Reflectance Statistics~\cite{gao2014efficient} & \\
    Spatio-spectral Statistics~(ML)~\cite{chakrabarti2012color} & 3.11 & 2.49 & 2.60 & 0.82 & 6.59 & 2.55\\
    General Gray-World~\cite{barnard2002comparison} & 3.21 & 2.38 & 2.53 & 0.71 & 7.10 & 2.49\\
    2nd-order Gray-Edge~\cite{van2007edge} & 3.20 & 2.26 & 2.44 & 0.75 & 7.27 & 2.49\\
    Bright Pixels~\cite{joze2012role} & 3.17 & 2.41 & 2.55 & 0.69 & 7.02 & 2.48\\
    1st-order Gray-Edge~\cite{van2007edge} & 3.20 & 2.22 & 2.43 & 0.72 & 7.36 & 2.46\\
    Spatio-spectral Statistics~(GP)~\cite{chakrabarti2012color} & 2.96 & 2.33 & 2.47 & 0.80 & 6.18 & 2.43\\
    Corrected-Moment~(19 Edge)~\cite{finlayson2013corrected} & 3.03 & 2.11 & 2.25 & 0.68 & 7.08 & 2.34\\
    Corrected-Moment~(19 Color)\cite{finlayson2013corrected} & 3.05 & 1.90 & 2.13 & 0.65 & 7.41 & 2.26\\
    Bright-and-dark Colors PCA~\cite{cheng2014illuminant} & 2.92 & 2.04 & 2.24 & 0.62 & 6.61 & 2.23\\
	%Dichromatic Gray Pixel~\cite{qian2018dichromatic} & \\
	\hline
    \textbf{Color Tiger (proposed)} & 2.96 & 1.70 & 1.97 & 0.53 & 7.50 & 2.09\\
    \hline
    Color Dog~\cite{banic2015acolor} & 2.83 & 1.77 & 2.03 & 0.48 & 7.04 & 2.03\\
    \textit{ideal case for Color Tiger} & 2.52 & 1.63 & 1.87 & 0.53 & 5.95 & 1.89\\
    %Shi et al. 2016~\cite{shi2016deep} & 2.24 & 1.46 & 1.68 & 0.48 & 6.08 & 1.74\\
    CCC~\cite{barron2015convolutional} & 2.38 & 1.48 & 1.69 & 0.45 & 5.85 & 1.74\\
    %Cheng 2015~\cite{cheng2015effective} & 2.18 & 1.48 & 1.64 & 0.46 & 5.03 & 1.65\\
    FFCC~\cite{barron2017fast} & 1.99 & 1.31 & 1.43 & 0.35 & 4.75 & 1.44\\
    
    \hline
    
\end{tabular}
\end{table}

\begin{table}[ht]
%\normalsize
\scriptsize
\caption{Performance of different color constancy methods on the original GreyBall dataset (lower median is better).}
\label{tab:gb}
\centering
\begin{tabular}{|c|c|c|c|}
    \hline
    \textbf{method} & \textbf{mean $(^{\circ})$} & \textbf{median $(^{\circ})$} & \textbf{trimean $(^{\circ})$}\\
    
    \hline
    \hline
    
    do nothing & 8.28 & 6.70 & 7.25\\
    \hline
	\multicolumn{4}{|c|}{\textbf{Low-level statistics-based methods}}\\
	\hline
	Gray-world~(GW)~\cite{buchsbaum1980spatial} & 7.87 & 6.97 & 7.14\\
    \hline
    White-Patch~(WP)~\cite{funt2010rehabilitation} & 6.80 & 5.30 & 5.77\\
    \hline
    Shades-of-Gray~\cite{finlayson2004shades} & 6.14 & 5.33 & 5.51\\
    \hline
    General Gray-World~\cite{barnard2002comparison} & 6.14 & 5.33 & 5.51\\
    \hline
    1st-order Gray-Edge~\cite{van2007edge} & 5.88 & 4.65 & 5.11\\
    \hline
    2nd-order Gray-Edge~\cite{van2007edge} & 6.10 & 4.85 & 5.28\\
    \hline
	\multicolumn{4}{|c|}{\textbf{Learning-based methods}}\\
	\hline
    Pixel-based gamut~\cite{finlayson2006gamut} & 7.07 & 5.81 & 6.12\\
    \hline
    Edge-based gamut~\cite{finlayson2006gamut} & 6.81 & 5.81 & 6.03\\
    \hline
    Intersection-based gamut~\cite{finlayson2006gamut} & 6.93 & 5.80 & 6.05\\
    \hline
	Natural Image Statistics~\cite{gijsenij2011color} & 5.19 & 3.93 & 4.31\\
	\hline
	Exemplar-based learning~\cite{joze2012exemplar} & 4.38 & 3.43 & 3.67\\
	\hline
	\hline
	\textbf{Color Tiger (proposed)} & 5.61 & 3.39 & 4.31\\
	\hline
	\hline
	Color Cat~(CC)~\cite{banic2015color} & \textbf{4.22} & 3.17 & \textbf{3.46}\\
	\hline
	Color Dog$_{WP, GW}$~\cite{banic2015acolor} & 5.27 & 3.71 & 4.16\\
	\hline
	Smart Color Cat~(SCC)~\cite{banic2015using} & 4.62 & 3.52 & 3.80\\
	\hline
	Color Dog$_{SCC}$~\cite{banic2015acolor} & 4.80 & 3.08 & 3.71\\
	\hline
	Color Dog$_{CC}$~\cite{banic2015acolor} & 4.50 & \textbf{2.86} & 3.50\\
	\hline
	
\end{tabular}
\end{table}

\begin{table}[ht]
%\normalsize
\scriptsize	
\caption{Performance of different color constancy methods on the linear GreyBall dataset (lower median is better).}
\label{tab:lgb}
\centering
\begin{tabular}{|c|c|c|c|}
    \hline
    \textbf{method} & \textbf{mean $(^{\circ})$} & \textbf{median $(^{\circ})$} & \textbf{trimean $(^{\circ})$}\\
    
    \hline
    \hline
    
    do nothing & 15.62 & 14.00 & 14.56\\
    \hline
	\multicolumn{4}{|c|}{\textbf{Low-level statistics-based methods}}\\
	\hline
	Gray-world~(GW)~\cite{buchsbaum1980spatial} & 13.01 & 10.96 & 11.53\\
    \hline
    White-Patch~(WP)~\cite{funt2010rehabilitation} & 12.68 & 10.50 & 11.25\\
    \hline
    Shades-of-Gray~\cite{finlayson2004shades} & 11.55 & 9.70 & 10.23\\
    \hline
    General Gray-World~\cite{barnard2002comparison} & 11.55 & 9.70 & 10.23\\
    \hline
    1st-order Gray-Edge~\cite{van2007edge} & 10.58 & 8.84 & 9.18\\
    \hline
    2nd-order Gray-Edge~\cite{van2007edge} & 10.68 & 9.02 & 9.40\\
    \hline
	\multicolumn{4}{|c|}{\textbf{Learning-based methods}}\\
	\hline
    Edge-based gamut~\cite{finlayson2006gamut} & 12.78 & 10.88 & 11.38\\
    \hline
    Pixel-based gamut~\cite{finlayson2006gamut} & 11.79 & 8.88 & 9.97\\
    \hline
    Intersection-based gamut~\cite{finlayson2006gamut} & 11.81 & 8.93 & 10.00\\
    \hline
	%HVLI & 9.73 & 7.71 & 8.17\\
	%\hline
	Natural Image Statistics~\cite{gijsenij2011color} & 9.87 & 7.65 & 8.29\\
	\hline
	Color Dog$_{WP, GW}$~\cite{banic2015acolor} & 10.27 & 7.33 & 8.20\\
	\hline
	\hline
	\textbf{Color Tiger (proposed)} & 9.51 & 7.11 & 7.66\\
	\hline
	\hline
	Color Cat~(CC)~\cite{banic2015color} & 8.73 & 7.07 & 7.43\\
	\hline
	Exemplar-based learning~\cite{joze2012exemplar} & \textbf{7.97} & 6.46 & 6.77\\
	\hline
	Smart Color Cat~(SCC)~\cite{banic2015using} & 8.18 & 6.28 & 6.73\\
	\hline
	Color Dog$_{CC}$~\cite{banic2015acolor} & 8.81 & 5.98 & 6.97\\
	\hline
	Color Dog$_{SCC}$~\cite{banic2015acolor} & 8.51 & \textbf{5.55} & \textbf{6.56}\\
	\hline
    
\end{tabular}
\end{table}

\begin{table}[ht]
%\normalsize
\scriptsize
%\tiny
\caption{Performance of different color constancy methods on the new Cube dataset (lower Avg. is better). The used format is the same as in~\cite{barron2015convolutional}.}
\label{tab:cube}
\centering
%\begin{tabular}{|c|p{2mm}p{2mm}p{2mm}p{3mm}p{3mm}|p{2mm}|}
\begin{tabular}{|>{\centering}m{3.9cm}| >{\centering}m{3mm} >{\centering}m{3mm} >{\centering\arraybackslash}m{3mm} >{\centering\arraybackslash}m{3mm} >{\centering\arraybackslash}m{3mm} >{\centering\arraybackslash}m{3mm}|}

    \hline
    \textbf{Algorithm} & \textbf{Mean} & \textbf{Med.} & \textbf{Tri.} & \makecell{\textbf{Best}\\\textbf{25\%}} & \makecell{\textbf{Worst}\\\textbf{25\%}} & \textbf{Avg.}\\
    
    \hline
    
    White-Patch~\cite{funt2010rehabilitation} & 6.58 & 4.48 & 5.27 & 1.18 & 15.23 & 4.88\\
    Gray-world~\cite{buchsbaum1980spatial} & 3.75 & 2.91 & 3.15 & 0.69 & 8.18 & 2.87\\
	\hline
    \textbf{Color Tiger (proposed)} & 2.94 & 2.59 & 2.66 & 0.61 & 5.88 & 2.35\\
	\hline
	Double-opponency (max pooling)~\cite{gao2015color} & 2.66 & 1.69 & 1.89 & 0.47 & 6.60 & 1.92\\
    Shades-of-Gray~\cite{finlayson2004shades} & 2.58 & 1.79 & 1.95 & 0.38 & 6.19 & 1.84\\
    2nd-order Gray-Edge~\cite{van2007edge} & 2.49 & 1.60 & 1.80 & 0.49 & 6.00 & 1.84\\
    1st-order Gray-Edge~\cite{van2007edge} & 2.45 & 1.58 & 1.81 & 0.48 & 5.89 & 1.81\\
    General Gray-World~\cite{barnard2002comparison} & 2.50 & 1.61 & 1.79 & 0.37 & 6.23 & 1.76\\
    Using gray pixels~\cite{yang2015efficient} & 2.40 & 1.45 & 1.65 & 0.39 & 6.05 & 1.69\\
    Color Mule~\cite{banic2015perceptual} & 1.62 & 0.84 & 1.07 & 0.21 & 4.33 & 1.06\\
    Color Dog~\cite{banic2015acolor} & 1.50 & 0.81 & 0.99 & 0.27 & 3.86 & 1.05\\
    Smart Color Cat~\cite{banic2015using} & 1.49 & 0.88 & 1.06 & 0.24 & 3.75 & 1.04\\
	Color Beaver (using Gray-world)~\cite{koscevic2019color} & 1.48 & 0.76 & 0.98 & 0.21 & 3.90 & 0.98\\
    
    \hline
    
\end{tabular}
\end{table}

\begin{table}[ht]
%\normalsize
\scriptsize
%\tiny
\caption{Performance of different color constancy methods on the new Cube+ dataset (lower Avg. is better). The used format is the same as in~\cite{barron2015convolutional}.}
\label{tab:cube_plus}
\centering
%\begin{tabular}{|c|p{2mm}p{2mm}p{2mm}p{3mm}p{3mm}|p{2mm}|}
\begin{tabular}{|>{\centering}m{3.9cm}| >{\centering}m{3mm} >{\centering}m{3mm} >{\centering\arraybackslash}m{3mm} >{\centering\arraybackslash}m{3mm} >{\centering\arraybackslash}m{3mm} >{\centering\arraybackslash}m{3mm}|}

    \hline
    \textbf{Algorithm} & \textbf{Mean} & \textbf{Med.} & \textbf{Tri.} & \makecell{\textbf{Best}\\\textbf{25\%}} & \makecell{\textbf{Worst}\\\textbf{25\%}} & \textbf{Avg.}\\
    
    \hline
    
    White-Patch~\cite{funt2010rehabilitation} & 9.69 & 7.48 & 8.56 & 1.72 & 20.49 & 7.38\\
    Gray-world~\cite{buchsbaum1980spatial} & 7.71 & 4.29 & 4.98 & 1.01 & 20.19 & 5.08\\
	Double-opponency (max pooling)~\cite{gao2015color} & 6.76 & 3.44 & 4.15 & 0.79 & 18.54 & 4.27\\
    Using gray pixels~\cite{yang2015efficient} & 6.65 & 3.26 & 3.95 & 0.68 & 18.75 & 4.05\\
	\hline
    \textbf{Color Tiger (proposed)} & 3.91 & 2.05 & 2.53 & 0.98 & 10.00 & 2.88\\
	\hline
	Double-opponency (max pooling)~\cite{gao2015color} & 5.19 & 1.35 & 2.10 & 0.32 & 16.85 & 2.40\\
    Color Mule~\cite{banic2015perceptual} & 5.16 & 1.30 & 2.03 & 0.25 & 16.93 & 2.25\\
    Shades-of-Gray~\cite{finlayson2004shades} & 2.59 & 1.73 & 1.93 & 0.46 & 6.19 & 1.90\\
    2nd-order Gray-Edge~\cite{van2007edge} & 2.50 & 1.59 & 1.78 & 0.48 & 6.08 & 1.83\\
    1st-order Gray-Edge~\cite{van2007edge} & 2.41 & 1.52 & 1.72 & 0.45 & 5.89 & 1.76\\
    Color Dog~\cite{banic2015acolor} & 3.32 & 1.19 & 1.60 & 0.22 & 10.22 & 1.70\\
    General Gray-World~\cite{barnard2002comparison} & 2.38 & 1.43 & 1.66 & 0.35 & 6.01 & 1.64\\
    Smart Color Cat~\cite{banic2015using} & 2.27 & 1.35 & 1.61 & 0.34 & 5.72 & 1.58\\
	Color Beaver (using Gray-world)~\cite{koscevic2019color} & 1.49 & 0.77 & 0.98 & 0.21 & 3.94 & 0.99\\
    
    \hline
    
\end{tabular}
\end{table}

\begin{figure*}[htb]
    \centering
    
	\subfloat[]{
	\includegraphics[height=2.8cm]{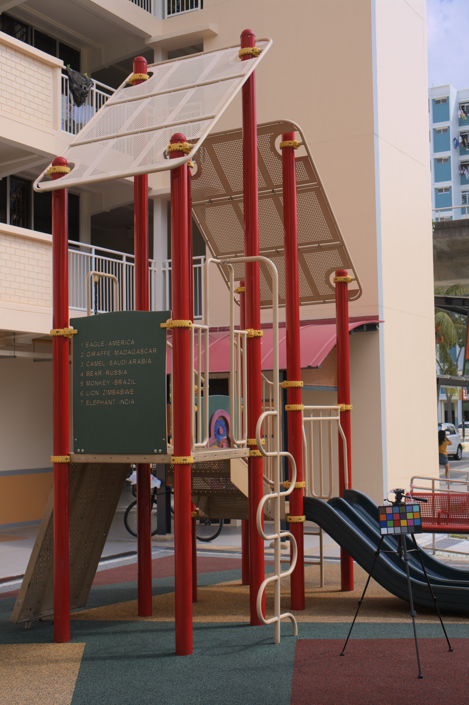}
	\label{fig:failure_3}
	}%
	~%
	\subfloat[]{
	\includegraphics[height=2.8cm]{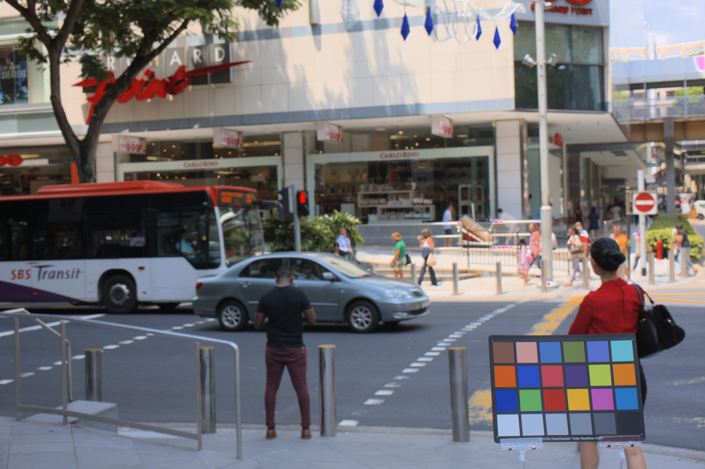}
	\label{fig:failure_5}
	}%
	~%
	\subfloat[]{
	\includegraphics[height=2.8cm]{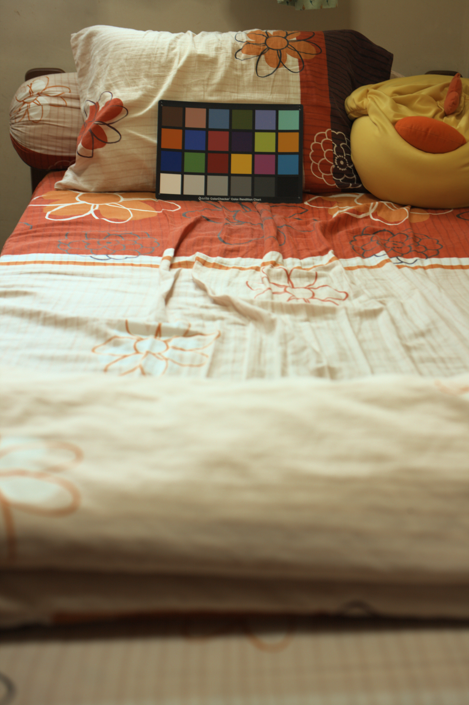}
	\label{fig:failure_7}
	}%
	~%
	\subfloat[]{
	\includegraphics[height=2.8cm]{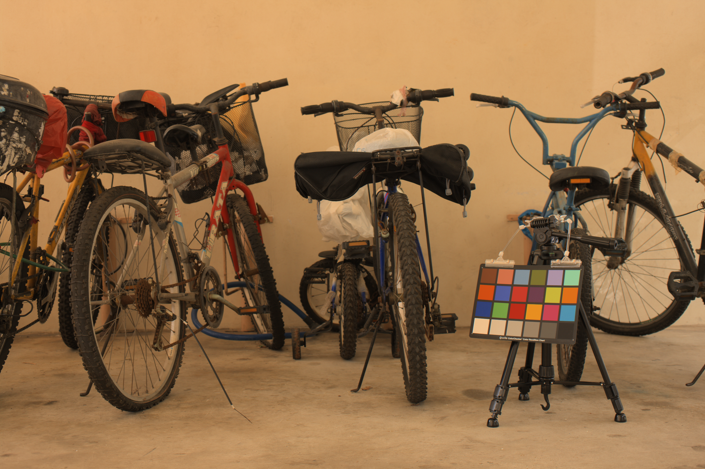}
	\label{fig:failure_11}
	}%
	~%
	\subfloat[]{
	\includegraphics[height=2.8cm]{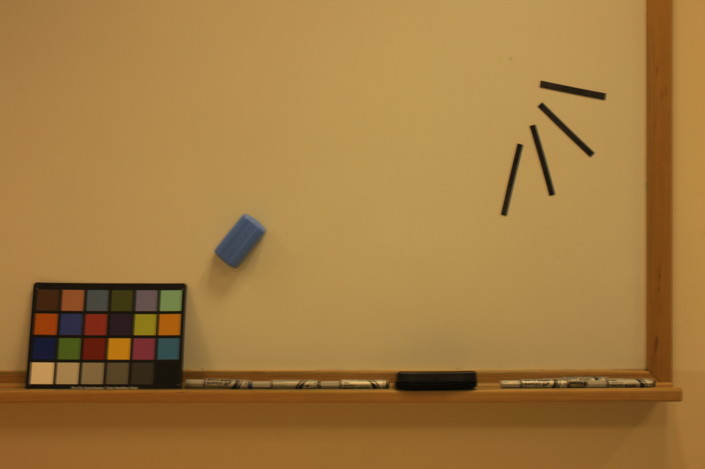}
	\label{fig:failure_15}
	}
	
    \caption{Failure cases for Color Tiger on Canon EOS-1Ds Mark III NUS dataset~\cite{cheng2014illuminant} with chromatic adaptation and Flash tone mapping~\cite{banic2016puma, banic2018flash} applied. The angular errors are a)~$3.03^\circ$, b)~$5^\circ$, c)~$7.07^\circ$, d)~$11^\circ$, and e)~$15.41^\circ$, respectively.}
	\label{fig:failure}
    
\end{figure*}

\subsection{Accuracy}
\label{subsec:accuracy}

Tables~\ref{tab:nus},~\ref{tab:gb},~\ref{tab:lgb},~\ref{tab:cube}, and~\ref{tab:cube_plus} show the comparisons between the accuracies of the proposed method and other illumination estimation methods on various datasets. The Avg. column in Tables~\ref{tab:nus} and~\ref{tab:cube} is the geometric mean of the mean, median, trimean, best 25\% and worst 25\% angular error. This statistics was first introduced in~\cite{barron2015convolutional}. On NUS datasets the proposed method outperforms all statistics-based methods and also many learning-based methods. The mean angular error of the statistics-based bright and dark colors method~\cite{cheng2014illuminant} is slightly lower, but it simultaneously has a higher median angular error, which is more important as mentioned earlier\cite{hordley2004re} and which effectively describes it as less accurate. For all datasets, except for the GreyBall dataset, its median angular error is below $3^\circ$, which was shown to be an acceptable error~\cite{finlayson2005colour, fredembach2008bright}. The methods that outperform the proposed method do so on average by a small, perceptually mostly unnoticeable margin.

While Color Tiger outperforms many methods in terms of median angular error, its higher values for some other statistics, e.g. worst $25$\% error, may be somewhat confusing. For example on the NUS datasets Color Tiger's $1.70^\circ$ median angular error is lower than Spatio-spectral Statistics' $2.58^\circ$, but simultaneously Color Tiger has worst $25$\% angular error of $7.50^\circ$, which is higher than $6.17^\circ$ of Spatio-spectral Statistics and in terms of absolute difference it represents a higher increase in error than in the case of median error. While it may seem that this makes Spatio-spectral Statistics better, this may be misleading. Namely, in accordance with Weber's law~\cite{weber2artikel} the noticeability of the difference between images chromatically adapted by two illumination estimations depends on the ratio, and not the absolute difference, of these estimations' errors~\cite{gijsenij2009perceptual, banic2015perceptual}. That being said, the decrease of the median angular error here outweighs the increase in worst $25$\% angular error. This is seen more clearly in Fig.~\ref{fig:failure}, which shows some cases of Color Tiger's illumination estimation failure. While for errors below $5^\circ$ the impact of the errors on the final image is smaller, it becomes more visible as the error increases. Nevertheless, it may be argued that the difference between the error on Fig.~\ref{fig:failure_7} and Fig.~\ref{fig:failure_11} is more visible than the difference between the error on Fig.~\ref{fig:failure_11} and Fig.~\ref{fig:failure_15} despite the latter having a higher absolute value. This example also demonstrates the importance of the median angular error comparison and the potentially misleading interpretations of values of some other commonly used angular error statistics.

Nevertheless, the question remains whether worst $25$\% error is high due to using only two centers or Color Tiger's inability to properly select the right one for a given image. To answer this question, it is enough to take a look at the accuracy statistics of the \textit{ideal case for Color Tiger} method in Table~\ref{tab:nus}. Its steps are the same as those of Color Tiger, but the two centers are learned on the ground-truth and when an image is processed, the method always chooses the optimal center. The worst $25$\% error of this ideal and unrealistic method is $5.95^\circ$. Since this is also relatively high and the centers are chosen optimally, it can be concluded that the reason for the high worst $25$\% error is the number of centers i.e. the fact that only two centers are used. For example, if the method optimally choose between 3, 4, or 5 ideally learned centers, the worst $25$\% errors would be $4.29^\circ$, $3.76^\circ$, and $3.39^\circ$, respectively. As stated earlier, choosing between more centers is a harder classification problem and also one of the reasons why only two fixed centers are chosen during Color Tiger's training.

Despite all these problems and an accuracy that is below that of the state-of-the-art methods, Color Tiger effectively proves that successful learning-based illumination estimation is indeed possible even without ground-truth illumination, which is not the case for the mentioned state-of-the-art methods that rely heavily on such information. In that sense Color Tiger should also be considered as a successful proof-of-concept method for a new approach and not only just another method that must beat all existing methods. The same goes for Color Bengal Tiger, which also demonstrates a highly successful inter-camera learning without any access to ground-truth.

\begin{figure}[htb]
    \centering
    
	\includegraphics[width=\linewidth]{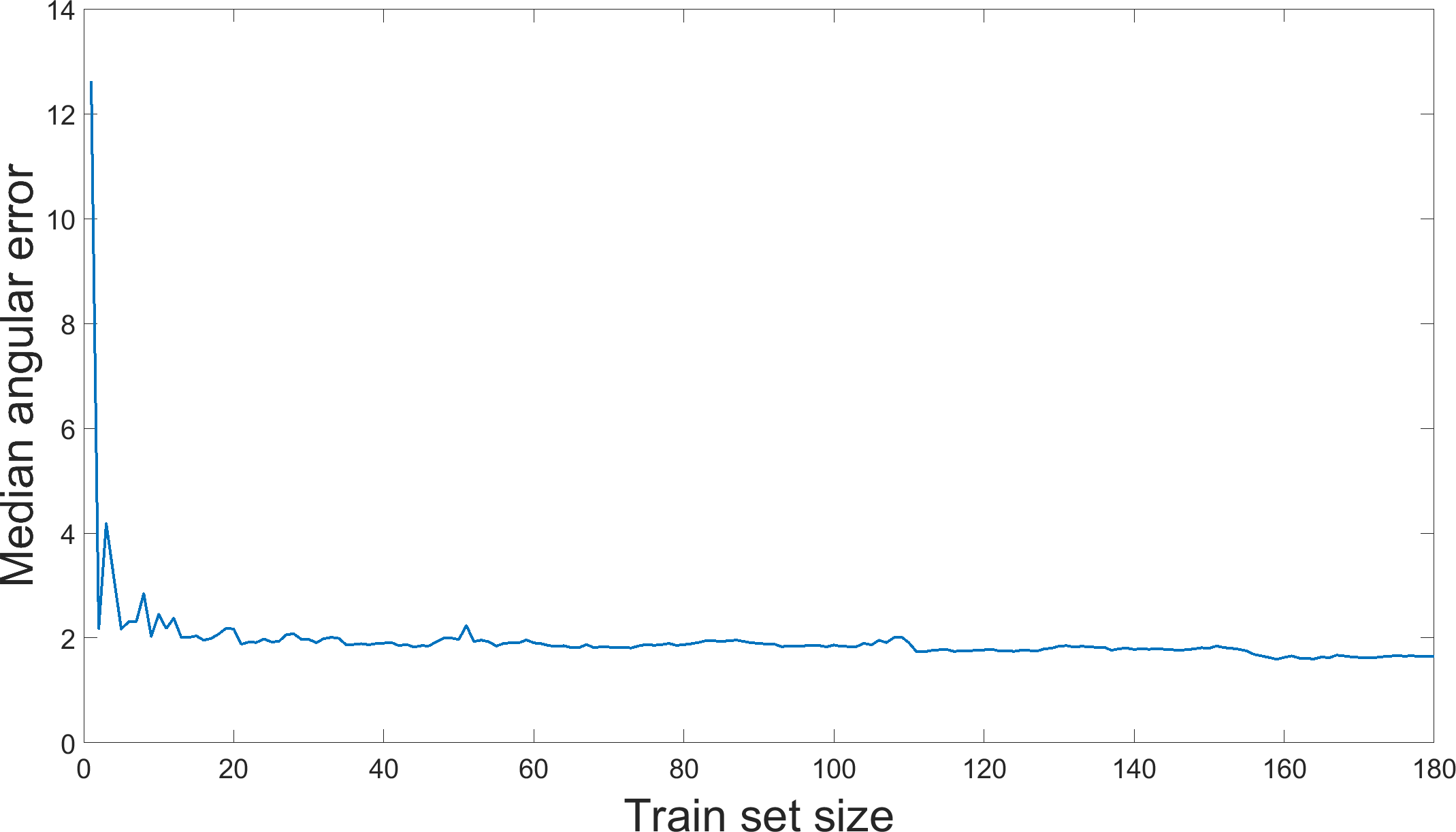}
	
    \caption{Influence of limiting the train fold sizes to only a given number of images on the median angular error of the Color Tiger method achieved on the Sony~\cite{cheng2014illuminant} dataset.}
	\label{fig:train_size}
    
\end{figure}

Another interesting property of the Color Tiger method is that it requires a relatively small number of training images. Fig.~\ref{fig:train_size} shows the influence of limiting the train fold sizes to only a given number of images on the median angular error of the Color Tiger method achieved on the Sony~\cite{cheng2014illuminant} dataset. As the limit rises, the median angular error gets stable relatively quickly and for any train size over $20$ it remains below $2^\circ$.

Although the Cube dataset is a new one, besides the results for the proposed method Table~\ref{tab:cube} additionally includes only the results for some well-known statistics-methods and for some of the precursors of the proposed method. The reason for not including other state-of-the-art methods is that besides the methods' descriptions provided in their respective papers, in too many cases some additional information and even latent parameters are needed to fully reproduce the reported results. Therefore, in order not to report the results based on suboptimal implementations, such methods have been left out.

\subsection{Inter-camera accuracy}
\label{subsec:inter}

Eight NUS datasets~\cite{cheng2014illuminant} were used to check the accuracy of the proposed Color Bengal Tiger method. Each of them was created by using a different camera. These datasets are supposed to contain the same scenes and although this varies from dataset to dataset, the content type distribution is effectively the same among all datasets. Color Bengal Tiger was tested by using all possible ordered pairs of distinct NUS datasets where the first dataset was used for training i.e. learning the illumination centers and the second one for testing and learning matrix $\mathbf{G}$. The obtained results are summarized in Table~\ref{tab:cbt} and for the sake of simplicity only the median angular error is given since it is the most important statistics.

\begin{table}[ht]
%\normalsize
\scriptsize
%\tiny
\caption{Median angular error achieved when training Color Bengal Tiger on all images of a given NUS dataset and testing on images of another NUS datasets (lower is better).}
\label{tab:cbt}
\centering
\begin{tabular}{|c|c|cccccccc|}

	\hline
	\multicolumn{10}{|c|}{\textbf{Test dataset}}\\
	\hline
	\multicolumn{2}{|c|}{} & \textbf{C1} & \textbf{C2} & \textbf{Fuji} &  \textbf{N52} &  \textbf{Oly} &  \textbf{Pan} &  \textbf{Sam} & \textbf{Sony}\\
	\hline
	\parbox[t]{2mm}{\multirow{8}{*}{\rotatebox[origin=c]{90}{\textbf{Train dataset}}}} & \textbf{C1} & - & 1.81 & 1.79 & 1.74 & 1.71 & 1.88 & 1.67 & 1.76\\
	& \textbf{C2} & 1.84 & - & 1.87 & 1.78 & 1.78 & 1.63 & 1.70 & 1.60\\
	& \textbf{Fuji} & 2.12 & 1.69 & - & 1.70 & 1.74 & 1.76 & 1.61 & 1.75\\
	& \textbf{N52} & 1.79 & 1.92 & 1.85 & - & 2.07 & 1.63 & 1.83 & 1.66\\
	& \textbf{Oly} & 2.09 & 1.75 & 1.75 & 1.98 & - & 1.90 & 1.64 & 1.55\\
	& \textbf{Pan} & 1.87 & 1.81 & 1.89 & 1.73 & 2.10 & - & 1.72 & 1.71\\
	& \textbf{Sam} & 1.97 & 1.69 & 1.81 & 1.75 & 1.72 & 1.89 & - & 1.67\\
	& \textbf{Sony} & 1.94 & 1.66 & 1.72 & 1.78 & 1.81 & 1.68 & 1.67 & -\\
    \hline

\end{tabular}
\end{table}

The median angular error is almost always below $2^\circ$, which means that the Color Bengal Tiger's learning and application was conducted successfully in practically all cases. For comparison, Table~\ref{tab:cbt_no_gains} shows the same results, but without neutralizing the camera sensor gains of each dataset during both training and testing. In comparison to Table~\ref{tab:cbt} the medians are significantly higher, which clearly shows the importance of using the estimation of camera sensor gains to bring all images into the same neutral RGB colorspace.

\begin{table}[ht]
%\normalsize
\scriptsize
%\tiny
\caption{Same as Table~\ref{tab:cbt}, but without neutralizing camera sensor gains (lower is better).}
\label{tab:cbt_no_gains}
\centering
\begin{tabular}{|c|c|cccccccc|}

	\hline
	\multicolumn{10}{|c|}{\textbf{Test dataset}}\\
	\hline
	\multicolumn{2}{|c|}{} & \textbf{C1} & \textbf{C2} & \textbf{Fuji} &  \textbf{N52} &  \textbf{Oly} &  \textbf{Pan} &  \textbf{Sam} & \textbf{Sony}\\
	\hline
	\parbox[t]{2mm}{\multirow{8}{*}{\rotatebox[origin=c]{90}{\textbf{Train dataset}}}} & \textbf{C1} & - & 1.76 & 4.19 & 3.83 & 5.13 & 2.69 & 3.63 & 5.13\\
	& \textbf{C2} & 1.79 & - & 4.67 & 3.75 & 5.42 & 2.44 & 3.54 & 4.89\\
	& \textbf{Fuji} & 5.17 & 4.58 & - & 6.38 & 2.26 & 5.62 & 5.57 & 8.16\\
	& \textbf{N52} & 4.14 & 3.55 & 6.24 & - & 6.53 & 2.32 & 1.96 & 2.66\\
	& \textbf{Oly} & 5.65 & 4.94 & 1.98 & 6.22 & - & 5.75 & 5.29 & 8.17\\
	& \textbf{Pan} & 2.52 & 2.18 & 5.41 & 2.52 & 5.93 & - & 2.68 & 3.60\\
	& \textbf{Sam} & 3.69 & 2.97 & 5.26 & 2.05 & 5.47 & 2.08 & - & 3.35\\
	& \textbf{Sony} & 5.30 & 4.96 & 8.72 & 2.86 & 8.78 & 3.65 & 3.60 & -\\
    \hline

\end{tabular}
\end{table}

\begin{table*}[ht]
%\normalsize
%\scriptsize
\tiny
\caption{Data from Table 4 taken from~\cite{cheng2014illuminant} with training and testing time in minutes for the Canon1DsMarkIII dataset; the table has been extended with the approximated (*) data for the proposed Color Tiger and Color Bengal Tiger methods.}
\label{tab:time}
\centering

\begin{tabular}{|c||c|c|c|c|c|c|c|c||c|c|c|c|c|c|c|c|c|}

	\hline
	Method & PCA~\cite{cheng2014illuminant} & GW~\cite{buchsbaum1980spatial} & WP~\cite{funt2010rehabilitation} & SoG~\cite{finlayson2004shades} & GGW~\cite{barnard2002comparison} & BP~\cite{joze2012role} & GE1~\cite{van2007edge} & GE2~\cite{van2007edge} & PG~\cite{barnard2000improvements} & EG~\cite{barnard2000improvements} & IG~\cite{barnard2000improvements} & BL~\cite{gehler2008bayesian} & ML~\cite{chakrabarti2012color} & GP~\cite{chakrabarti2012color} & NIS~\cite{gijsenij2011color} & Proposed\\
	\hline
	Train (min) & 0.0 & 0.0 & 0.0 & 0.0 & 0.0 & 0.0 & 0.0 & 0.0 & 254 & 245 & 251 & 32.2 & 133.2 & 126.9 & 453.2 & 116.8*\\
	\hline
	Test (min) & 9.9 & 7.8 & 8.0 & 14.6 & 27.3 & 13.6 & 29.5 & 34.6 & 254 & 184 & 235 & 2316 & 168.3 & 61.7 & 25.2 & 15.8*\\
    \hline

\end{tabular}
\end{table*}

\begin{figure}[htb]
    \centering
    
	\includegraphics[width=\linewidth]{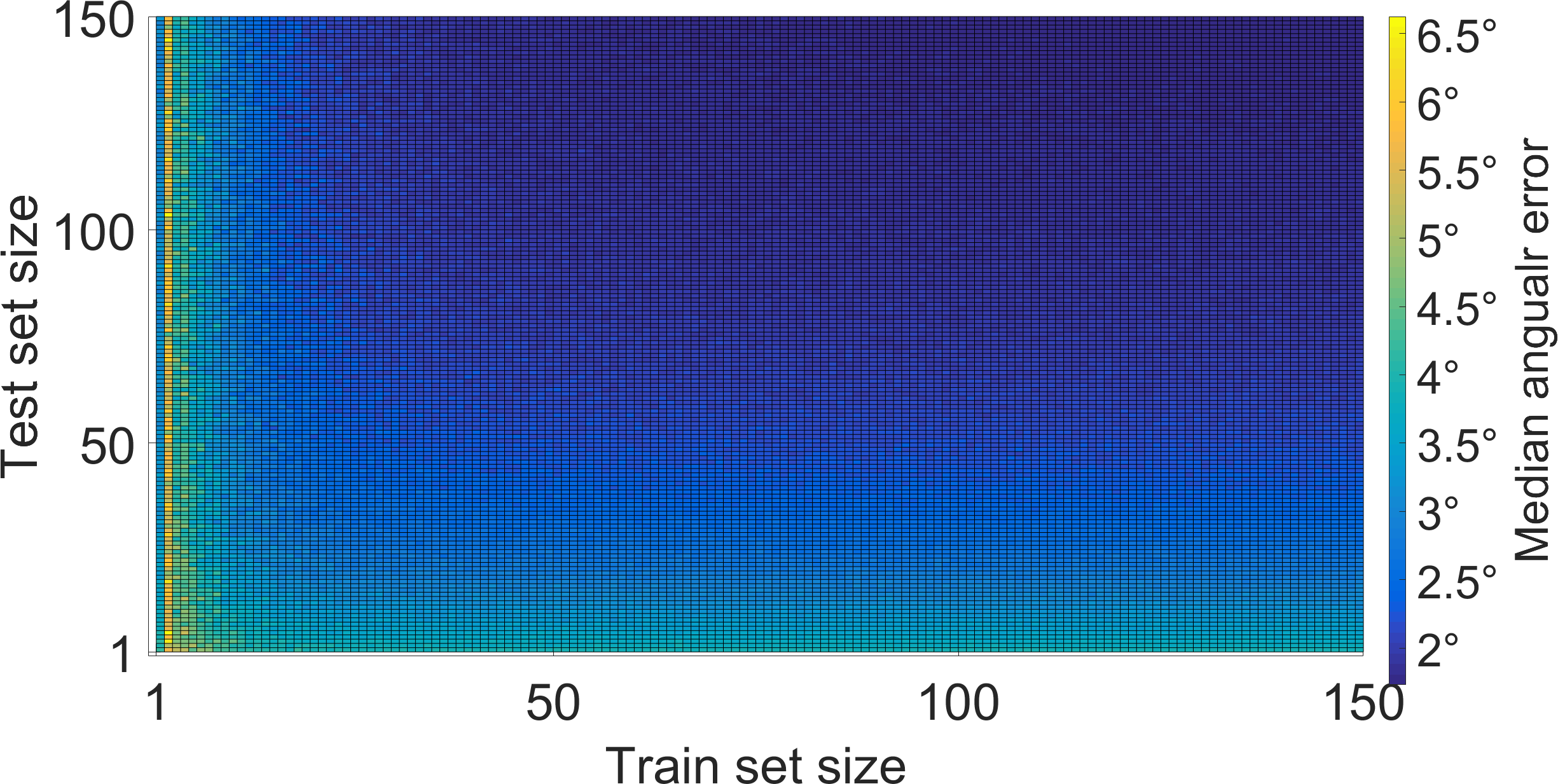}
	
    \caption{Influence of sizes of Fuji and Sony~\cite{cheng2014illuminant} dataset samples on Color Bengal Tigers's median angular error on the whole Sony dataset when Fuji sample is used to learn the cluster centers and matrix $\mathbf{G}$ and Sony sample is used to learn matrix $\mathbf{G}'$; each median angular error was calculated by averaging the results of $50$ random samplings.}
	\label{fig:fuji_sony_cbt}
    
\end{figure}

For the results in Table~\ref{tab:cbt} whole test datasets were used to learn the matrix $\mathbf{G}'$. To examine how the size of the training dataset for learning the cluster centers and matrix $\mathbf{G}$ and the size of the part of the test dataset used to learn matrix $\mathbf{G}'$ simultaneously influence the median angular error obtained by Color Bengal Tiger on the whole test dataset, an experiment was conducted by using the Fuji dataset~\cite{cheng2014illuminant} as the training dataset and the Sony dataset~\cite{cheng2014illuminant} as the test dataset. The results of the experiment are shown in Fig.~\ref{fig:fuji_sony_cbt}. It can be seen that learning of cluster centers is much more negatively affected by smaller dataset sizes than learning of matrix $\mathbf{G}'$. However, as the dataset size rises, the accuracy contribution of learning of cluster centers converges significantly faster. This is in accordance with the results shown in Fig.~\ref{fig:train_size} and it also shows that if only few images from a new camera are given, it is probably better to apply Color Tiger training to them than to apply Color Bengal Tiger learning that relies on clusters learned on another camera. Nevertheless, when full datasets are available, Color Bengal Tiger can still outperform Color Tiger, which can be seen when the results from Table~\ref{tab:cbt} are compared to results of Color Tiger on individual NUS datasets. This makes Color Bengal Tiger a successful proof-of-concept of unsupervised inter-camera learning for color constancy.

The currently closest approach to the one of the proposed method is described in~\cite{gao2017improving}. While the method proposed there is also tested on the NUS datasets, there are no numerical results given explicitly. Instead, there is a summary graph in that paper's Figure 12 that covers only some of the inter-camera training and testing cases that are possible for the NUS datasets. Additionally, the average error mentioned in that paper is not defined, which makes it unclear whether it refers to the mean angular error, to the usually used median angular error, or to the recently introduced definition of the average error as the geometric mean of several other error statistics~\cite{barron2015convolutional}. By visually comparing the results obtained in the other paper's mentioned graph to the ones in Table~\ref{tab:cbt}, it can be concluded that the two methods are on par in terms of accuracy. However, it must be stressed again that the method proposed here is more practical since it does not require the ground-truth illuminations and camera spectral sensitivities.

\subsection{Discussion}
\label{subsec:discussion}

Beyond the fact that the proposed method outperformed all statistics-based methods and many learning-based methods, a far more important thing to stress here is that it did so without having any ground-truth illumination data available. Not only does this show the abundance of information available in even the simplest natural image statistics, but it also opens a simple and effective way of achieving highly accurate illumination estimation for a given sensor by only providing training images without ground-truth illumination data. As demonstrated, by introducing camera sensor gains estimation, such illumination estimation can also be performed for images taken with a sensor that is different from the one that was used to create the images in the training set, which opens the way for effective inter-camera learning. Skipping the calibration of training images can save a significant amount of time and in some cases this can make the proposed method more suitable for practical applications than other learning-based methods. Since in production it only needs to execute Gray-world and White-patch, two of the fastest statistics-based methods~\cite{cheng2014illuminant} with practically no memory requirements, and then perform a small and constant number of calculations for voting, the proposed method is hardware-friendly and thus widely applicable for embedded systems. Another potential benefit of the proposed method is that it avoids problems connected to false ground-truth data when calibration is not performed accurately~\cite{zakizadeh2015hybrid}. Finally, if the assumptions and steps proposed in this paper have led to the described results, it is reasonable to assume that with more sophisticated image statistics, more accurate voters, and better trimming procedures the proposed method could achieve even higher accuracy of illumination estimation.

Because of its nature the proposed method has another great advantage over other illumination estimation methods. Namely, it can be used in conditions where automatic online learning and adjustment of parameter values are required, e.g. when a camera is used in special illumination conditions for a prolonged period of time like in extreme regions or in closed spaces with specific lighting types. In such conditions a given camera's illumination estimation system could definitely improve its accuracy if it was recalibrated. While regularly this would require some kind of ground-truth extraction or manual intervention, the proposed method would simply solve this issue by performing the recalibration automatically based on the statistical properties of some of the recently taken images.

\subsection{Computational cost}
\label{subsec:computational}

The main computational tasks of the proposed methods during test time can be boiled down to executing Gray-world and White-patch methods once, while the influence of the last voting step can be dismissed. During train time the main computational tasks are running the Shades-of-Gray method eight times on each of the images in the training set, while the final clustering step takes only a small percentage of the whole time and so it can also be dismissed. By taking these facts into consideration and by using the train and test times obtained by various methods durign the experiments performed and described in~\cite{cheng2014illuminant}, the train and test times that would have been achieved in the same environment by the proposed methods were approximated and put in Table~\ref{tab:time}. As shown, the proposed methods compare relatively well to other methods.

%------------------------------------------------------------------------

\section{Conclusions and future research}
\label{sec:conclusions}

A fast and hardware-friendly unsupervised learning-based method that learns its parameter values from images with unknown ground-truth illumination has been proposed. In terms of accuracy the method outperforms all statistics-based and many learning-based methods. This demonstrates how to achieve highly accurate color constancy for a given sensor without carrying out the usually time consuming calibration of training images. It has also been shown how to train on images created with one camera sensor and use the learned parameters on images created with another camera sensor by simply estimating and neutralizing the sensor gains for both cameras. The proposed method could possibly also be an important step in color constancy philosophy, especially now when there are large amounts of non-calibrated images available on the Internet. Additionally, a new high quality color constancy benchmark dataset with $1707$ calibrated images has been created, used for testing, and made publicly available. The dataset was named Cube+ and it is currently one of the largest datasets with calibrated linear images that were created by using a single camera. Future research will focus on extracting more useful information from statistics-based illumination estimations obtained on training images without ground-truth illumination and on other ways of outlier removal.

\section*{Acknowledgment}
\label{sec:acknowledgment}

The authors would like to thank the anonymous reviewers for their valuable comments that helped to significantly improve the paper as well as Dr. Tomislav Petkovi{\'{c}} for his help in assembling the handle that connected the calibration object with the camera that was used to create the Cube dataset. 

\balance
\bibliographystyle{IEEEtran}
\bibliography{ct}

\begin{thebibliography}{10}
\providecommand{\url}[1]{#1}
\csname url@samestyle\endcsname
\providecommand{\newblock}{\relax}
\providecommand{\bibinfo}[2]{#2}
\providecommand{\BIBentrySTDinterwordspacing}{\spaceskip=0pt\relax}
\providecommand{\BIBentryALTinterwordstretchfactor}{4}
\providecommand{\BIBentryALTinterwordspacing}{\spaceskip=\fontdimen2\font plus
\BIBentryALTinterwordstretchfactor\fontdimen3\font minus
  \fontdimen4\font\relax}
\providecommand{\BIBforeignlanguage}[2]{{%
\expandafter\ifx\csname l@#1\endcsname\relax
\typeout{** WARNING: IEEEtran.bst: No hyphenation pattern has been}%
\typeout{** loaded for the language `#1'. Using the pattern for}%
\typeout{** the default language instead.}%
\else
\language=\csname l@#1\endcsname
\fi
#2}}
\providecommand{\BIBdecl}{\relax}
\BIBdecl

\bibitem{banic2018unsupervised}
N.~Bani{\'{c}} and S.~Lon{\v{c}}ari{\'{c}}, ``{U}nsupervised {L}earning for
  {C}olor {C}onstancy,'' in \emph{VISAPP}, 2018, pp. 181--188.

\bibitem{ebner2007color}
M.~Ebner, \emph{Color Constancy}, ser. The Wiley-IS\&T Series in Imaging
  Science and Technology.\hskip 1em plus 0.5em minus 0.4em\relax Wiley, 2007.

\bibitem{kim2012new}
S.~J. Kim, H.~T. Lin, Z.~Lu, S.~S{\"u}sstrunk, S.~Lin, and M.~S. Brown, ``A new
  in-camera imaging model for color computer vision and its application,''
  \emph{IEEE Transactions on Pattern Analysis and Machine Intelligence},
  vol.~34, no.~12, pp. 2289--2302, 2012.

\bibitem{gijsenij2011computational}
A.~Gijsenij, T.~Gevers, and J.~Van De~Weijer, ``{C}omputational color
  constancy: {S}urvey and experiments,'' \emph{Image Processing, IEEE
  Transactions on}, vol.~20, no.~9, pp. 2475--2489, 2011.

\bibitem{barnard2002comparison}
K.~Barnard, V.~Cardei, and B.~Funt, ``A comparison of computational color
  constancy algorithms. i: Methodology and experiments with synthesized data,''
  \emph{Image Processing, IEEE Transactions on}, vol.~11, no.~9, pp. 972--984,
  2002.

\bibitem{land1977retinex}
E.~H. Land, \emph{The retinex theory of color vision}.\hskip 1em plus 0.5em
  minus 0.4em\relax Scientific America., 1977.

\bibitem{funt2010rehabilitation}
B.~Funt and L.~Shi, ``The rehabilitation of {MaxRGB},'' in \emph{Color and
  Imaging Conference}, vol. 2010, no.~1.\hskip 1em plus 0.5em minus 0.4em\relax
  Society for Imaging Science and Technology, 2010, pp. 256--259.

\bibitem{banic2013using}
N.~Bani{\'{c}} and S.~Lon{\v{c}}ari{\'{c}}, ``{U}sing the {R}andom {S}prays
  {R}etinex {A}lgorithm for {G}lobal {I}llumination {E}stimation,'' in
  \emph{Proceedings of The Second Croatian Computer Vision Workshopn (CCVW
  2013)}.\hskip 1em plus 0.5em minus 0.4em\relax University of Zagreb Faculty
  of Electrical Engineering and Computing, 2013, pp. 3--7.

\bibitem{banic2014color}
------, ``{C}olor {R}abbit: {G}uiding the {D}istance of {L}ocal {M}aximums in
  {I}llumination {E}stimation,'' in \emph{Digital Signal Processing (DSP), 2014
  19th International Conference on}.\hskip 1em plus 0.5em minus 0.4em\relax
  IEEE, 2014, pp. 345--350.

\bibitem{banic2014improving}
------, ``Improving the {W}hite patch method by subsampling,'' in \emph{Image
  Processing (ICIP), 2014 21st IEEE International Conference on}.\hskip 1em
  plus 0.5em minus 0.4em\relax IEEE, 2014, pp. 605--609.

\bibitem{buchsbaum1980spatial}
G.~Buchsbaum, ``A spatial processor model for object colour perception,''
  \emph{Journal of The Franklin Institute}, vol. 310, no.~1, pp. 1--26, 1980.

\bibitem{finlayson2004shades}
G.~D. Finlayson and E.~Trezzi, ``Shades of gray and colour constancy,'' in
  \emph{Color and Imaging Conference}, vol. 2004, no.~1.\hskip 1em plus 0.5em
  minus 0.4em\relax Society for Imaging Science and Technology, 2004, pp.
  37--41.

\bibitem{van2007edge}
J.~Van De~Weijer, T.~Gevers, and A.~Gijsenij, ``Edge-based color constancy,''
  \emph{Image Processing, IEEE Transactions on}, vol.~16, no.~9, pp.
  2207--2214, 2007.

\bibitem{gijsenij2012improving}
A.~Gijsenij, T.~Gevers, and J.~Van De~Weijer, ``Improving color constancy by
  photometric edge weighting,'' \emph{Pattern Analysis and Machine
  Intelligence, IEEE Transactions on}, vol.~34, no.~5, pp. 918--929, 2012.

\bibitem{joze2012role}
H.~R.~V. Joze, M.~S. Drew, G.~D. Finlayson, and P.~A.~T. Rey, ``{T}he {R}ole of
  {B}right {P}ixels in {I}llumination {E}stimation,'' in \emph{Color and
  Imaging Conference}, vol. 2012, no.~1.\hskip 1em plus 0.5em minus 0.4em\relax
  Society for Imaging Science and Technology, 2012, pp. 41--46.

\bibitem{cheng2014illuminant}
D.~Cheng, D.~K. Prasad, and M.~S. Brown, ``Illuminant estimation for color
  constancy: why spatial-domain methods work and the role of the color
  distribution,'' \emph{JOSA A}, vol.~31, no.~5, pp. 1049--1058, 2014.

\bibitem{banic2019blue}
N.~Bani\'{c} and S.~Lon\v{c}ari\'{c}, ``{B}lue {S}hift {A}ssumption:
  {I}mproving {I}llumination {E}stimation {A}ccuracy for {S}ingle {I}mage from
  {U}nknown {S}ource,'' in \emph{VISAPP}, 2019, pp. 191--197.

\bibitem{quian2019revisiting}
Y.~Qian, S.~Pertuz, J.~Nikkanen, J.-K. K{\:{a}}m{\:{a}}r{\:{a}}inen, and
  J.~Matas, ``{R}evisiting {G}ray {P}ixel for {S}tatistical {I}llumination
  {E}stimation,'' in \emph{VISAPP}, 2019, pp. 36--46.

\bibitem{banic2018green}
N.~Bani{\'c} and S.~Lon{\v{c}}ari{\'c}, ``Green stability assumption:
  Unsupervised learning for statistics-based illumination estimation,''
  \emph{Journal of Imaging}, vol.~4, no.~11, p. 127, 2018.

\bibitem{barnard2000improvements}
K.~Barnard, ``Improvements to gamut mapping colour constancy algorithms,'' in
  \emph{European conference on computer vision}.\hskip 1em plus 0.5em minus
  0.4em\relax Springer, 2000, pp. 390--403.

\bibitem{finlayson2006gamut}
G.~D. Finlayson, S.~D. Hordley, and I.~Tastl, ``Gamut constrained illuminant
  estimation,'' \emph{International Journal of Computer Vision}, vol.~67,
  no.~1, pp. 93--109, 2006.

\bibitem{cardei2002estimating}
V.~C. Cardei, B.~Funt, and K.~Barnard, ``Estimating the scene illumination
  chromaticity by using a neural network,'' \emph{JOSA A}, vol.~19, no.~12, pp.
  2374--2386, 2002.

\bibitem{van2007using}
J.~Van De~Weijer, C.~Schmid, and J.~Verbeek, ``Using high-level visual
  information for color constancy,'' in \emph{Computer Vision, 2007. ICCV 2007.
  IEEE 11th International Conference on}.\hskip 1em plus 0.5em minus
  0.4em\relax IEEE, 2007, pp. 1--8.

\bibitem{gijsenij2007color}
A.~Gijsenij and T.~Gevers, ``{C}olor {C}onstancy using {N}atural {I}mage
  {S}tatistics.'' in \emph{CVPR}, 2007, pp. 1--8.

\bibitem{gehler2008bayesian}
P.~V. Gehler, C.~Rother, A.~Blake, T.~Minka, and T.~Sharp, ``Bayesian color
  constancy revisited,'' in \emph{Computer Vision and Pattern Recognition,
  2008. CVPR 2008. IEEE Conference on}.\hskip 1em plus 0.5em minus 0.4em\relax
  IEEE, 2008, pp. 1--8.

\bibitem{chakrabarti2012color}
A.~Chakrabarti, K.~Hirakawa, and T.~Zickler, ``Color constancy with
  spatio-spectral statistics,'' \emph{Pattern Analysis and Machine
  Intelligence, IEEE Transactions on}, vol.~34, no.~8, pp. 1509--1519, 2012.

\bibitem{banic2015color}
N.~Bani{\'{c}} and S.~Lon{\v{c}}ari{\'{c}}, ``{C}olor {C}at: {R}emembering
  {C}olors for {I}llumination {E}stimation,'' \emph{Signal Processing Letters,
  IEEE}, vol.~22, no.~6, pp. 651--655, 2015.

\bibitem{banic2015using}
------, ``Using the red chromaticity for illumination estimation,'' in
  \emph{Image and Signal Processing and Analysis (ISPA), 2015 9th International
  Symposium on}.\hskip 1em plus 0.5em minus 0.4em\relax IEEE, 2015, pp.
  131--136.

\bibitem{banic2015acolor}
------, ``{C}olor {D}og: {G}uiding the {G}lobal {I}llumination {E}stimation to
  {B}etter {A}ccuracy,'' in \emph{VISAPP}, 2015, pp. 129--135.

\bibitem{finlayson2013corrected}
G.~D. Finlayson, ``Corrected-moment illuminant estimation,'' in
  \emph{Proceedings of the IEEE International Conference on Computer Vision},
  2013, pp. 1904--1911.

\bibitem{cheng2015effective}
D.~Cheng, B.~Price, S.~Cohen, and M.~S. Brown, ``Effective learning-based
  illuminant estimation using simple features,'' in \emph{Proceedings of the
  IEEE Conference on Computer Vision and Pattern Recognition}, 2015, pp.
  1000--1008.

\bibitem{barron2015convolutional}
J.~T. Barron, ``{C}onvolutional {C}olor {C}onstancy,'' in \emph{Proceedings of
  the IEEE International Conference on Computer Vision}, 2015, pp. 379--387.

\bibitem{barron2017fast}
J.~T. Barron and Y.-T. Tsai, ``{F}ast {F}ourier {C}olor {C}onstancy,'' in
  \emph{Computer Vision and Pattern Recognition, 2017. CVPR 2017. IEEE Computer
  Society Conference on}, vol.~1.\hskip 1em plus 0.5em minus 0.4em\relax IEEE,
  2017.

\bibitem{bianco2015color}
S.~Bianco, C.~Cusano, and R.~Schettini, ``{C}olor {C}onstancy {U}sing {CNN}s,''
  in \emph{Proceedings of the IEEE Conference on Computer Vision and Pattern
  Recognition Workshops}, 2015, pp. 81--89.

\bibitem{shi2016deep}
W.~Shi, C.~C. Loy, and X.~Tang, ``{D}eep {S}pecialized {N}etwork for
  {I}lluminant {E}stimation,'' in \emph{European Conference on Computer
  Vision}.\hskip 1em plus 0.5em minus 0.4em\relax Springer, 2016, pp. 371--387.

\bibitem{hu2017fc4}
Y.~Hu, B.~Wang, and S.~Lin, ``{F}ully {C}onvolutional {C}olor {C}onstancy with
  {C}onfidence-weighted {P}ooling,'' in \emph{Computer Vision and Pattern
  Recognition, 2017. CVPR 2017. IEEE Conference on}.\hskip 1em plus 0.5em minus
  0.4em\relax IEEE, 2017, pp. 4085--4094.

\bibitem{oh2017approaching}
S.~W. Oh and S.~J. Kim, ``Approaching the computational color constancy as a
  classification problem through deep learning,'' \emph{Pattern Recognition},
  vol.~61, pp. 405--416, 2017.

\bibitem{koscevic2019color}
K.~Ko{\v{s}}{\v{c}}evi{\'{c}}, N.~Bani\'{c}, and S.~Lon\v{c}ari\'{c}, ``{C}olor
  {B}eaver: {B}ounding {I}llumination {E}stimations for {H}igher {A}ccuracy,''
  in \emph{VISAPP}, 2019, pp. 183--190.

\bibitem{akbarinia2018colour}
A.~Akbarinia and C.~A. Parraga, ``Colour constancy beyond the classical
  receptive field,'' \emph{IEEE transactions on pattern analysis and machine
  intelligence}, vol.~40, no.~9, pp. 2081--2094, 2018.

\bibitem{woo2018improving}
S.-M. Woo, S.-H. Lee, J.-S. Yoo, and J.-O. Kim, ``Improving color constancy in
  an ambient light environment using the phong reflection model,'' \emph{IEEE
  Transactions on Image Processing}, vol.~27, no.~4, pp. 1862--1877, 2018.

\bibitem{yang2015efficient}
K.-F. Yang, S.-B. Gao, and Y.-J. Li, ``Efficient illuminant estimation for
  color constancy using grey pixels,'' in \emph{Proceedings of the IEEE
  Conference on Computer Vision and Pattern Recognition}, 2015, pp. 2254--2263.

\bibitem{gao2015color}
S.-B. Gao, K.-F. Yang, C.-Y. Li, and Y.-J. Li, ``Color constancy using
  double-opponency,'' \emph{IEEE Transactions on Pattern Analysis and Machine
  Intelligence}, vol.~37, no.~10, pp. 1973--1985, 2015.

\bibitem{gao2017improving}
S.-B. Gao, M.~Zhang, C.-Y. Li, and Y.-J. Li, ``Improving color constancy by
  discounting the variation of camera spectral sensitivity,'' \emph{JOSA A},
  vol.~34, no.~8, pp. 1448--1462, 2017.

\bibitem{aytekin2017dataset}
{\c{C}}.~Aytekin, J.~Nikkanen, and M.~Gabbouj, ``{A} {D}ataset for {C}amera
  {I}ndependent {C}olor {C}onstancy,'' \emph{IEEE Transactions on Image
  Processing}, vol.~27, no.~2, pp. 530--544, 2017.

\bibitem{burkard2012assignment}
R.~Burkard, M.~Dell'Amico, and S.~Martello, \emph{Assignment problems: revised
  reprint}.\hskip 1em plus 0.5em minus 0.4em\relax SIAM, 2012.

\bibitem{mazin2015estimation}
B.~Mazin, J.~Delon, and Y.~Gousseau, ``Estimation of illuminants from
  projections on the planckian locus,'' \emph{IEEE Transactions on Image
  Processing}, vol.~24, no.~6, pp. 1944--1955, 2015.

\bibitem{vassilvitskii2007k}
S.~Vassilvitskii, \emph{{K}-means: {A}lgorithms, {A}nalyses,
  {E}xperiments}.\hskip 1em plus 0.5em minus 0.4em\relax Stanford University,
  2007.

\bibitem{japkowicz2011evaluating}
N.~Japkowicz and M.~Shah, \emph{Evaluating learning algorithms: a
  classification perspective}.\hskip 1em plus 0.5em minus 0.4em\relax Cambridge
  University Press, 2011.

\bibitem{banic2015perceptual}
N.~Bani{\'{c}} and S.~Lon{\v{c}}ari{\'{c}}, ``{A} {P}erceptual {M}easure of
  {I}llumination {E}stimation {E}rror,'' in \emph{VISAPP}, 2015, pp. 136--143.

\bibitem{schanda2007colorimetry}
J.~Schanda, \emph{Colorimetry: {U}nderstanding the {CIE} {S}ystem}.\hskip 1em
  plus 0.5em minus 0.4em\relax John Wiley \& Sons, 2007.

\bibitem{cheng2016two}
D.~Cheng, A.~Abdelhamed, B.~Price, S.~Cohen, and M.~S. Brown, ``{T}wo
  {I}lluminant {E}stimation and {U}ser {C}orrection {P}reference,'' in
  \emph{Proceedings of the IEEE Conference on Computer Vision and Pattern
  Recognition}, 2016, pp. 469--477.

\bibitem{ester1996density}
M.~Ester, H.-P. Kriegel, J.~Sander, X.~Xu \emph{et~al.}, ``A density-based
  algorithm for discovering clusters in large spatial databases with noise,''
  in \emph{Kdd}, vol.~96, no.~34, 1996, pp. 226--231.

\bibitem{kries1902theoretische}
J.~v. Kries, ``{T}heoretische {S}tudien {\"u}ber die {U}mstimmung des
  {S}ehorgans,'' \emph{Festschrift der Albrecht-Ludwigs-Universit{\"a}t in
  Freiburg}, pp. 145--148, 1902.

\bibitem{prasad2016strategies}
D.~K. Prasad, ``{S}trategies for {R}esolving {C}amera {M}etamers {U}sing 3+1
  {C}hannel.'' in \emph{CVPR Workshops}, 2016, pp. 954--962.

\bibitem{wug2016yourself}
S.~Wug~Oh, M.~S. Brown, M.~Pollefeys, and S.~Joo~Kim, ``Do it yourself
  hyperspectral imaging with everyday digital cameras,'' in \emph{Proceedings
  of the IEEE Conference on Computer Vision and Pattern Recognition}, 2016, pp.
  2461--2469.

\bibitem{ciurea2003large}
F.~Ciurea and B.~Funt, ``A large image database for color constancy research,''
  in \emph{Color and Imaging Conference}, vol. 2003, no.~1.\hskip 1em plus
  0.5em minus 0.4em\relax Society for Imaging Science and Technology, 2003, pp.
  160--164.

\bibitem{shi2015online}
\BIBentryALTinterwordspacing
B.~F. L.~Shi. (2015, May) {R}e-processed {V}ersion of the {G}ehler {C}olor
  {C}onstancy {D}ataset of 568 {I}mages. [Online]. Available:
  \url{http://www.cs.sfu.ca/ colour/data/}
\BIBentrySTDinterwordspacing

\bibitem{lynch2013colour}
S.~E. Lynch, M.~S. Drew, and k.~G.~D. Finlayson, ``{C}olour {C}onstancy from
  {B}oth {S}ides of the {S}hadow {E}dge,'' in \emph{Color and Photometry in
  Computer Vision Workshop at the International Conference on Computer
  Vision}.\hskip 1em plus 0.5em minus 0.4em\relax IEEE, 2013.

\bibitem{finlayson2017curious}
G.~D. Finlayson, G.~Hemrit, A.~Gijsenij, and P.~Gehler, ``{A} {C}urious
  {P}roblem with {U}sing the {C}olour {C}hecker {D}ataset for {I}lluminant
  {E}stimation,'' in \emph{Color and Imaging Conference}, vol. 2017,
  no.~25.\hskip 1em plus 0.5em minus 0.4em\relax Society for Imaging Science
  and Technology, 2017, pp. 64--69.

\bibitem{hemrit2018rehabilitating}
\BIBentryALTinterwordspacing
G.~Hemrit, G.~D. Finlayson, A.~Gijsenij, P.~V. Gehler, S.~Bianco, and M.~S.
  Drew, ``Rehabilitating the color checker dataset for illuminant estimation,''
  \emph{CoRR}, vol. abs/1805.12262, 2018. [Online]. Available:
  \url{http://arxiv.org/abs/1805.12262}
\BIBentrySTDinterwordspacing

\bibitem{banic2019past}
N.~Bani{\'{c}}, K.~Ko{\v{s}}{\v{c}}evi{\'{c}}, M.~Suba{\v{s}}i{\'{c}}, and
  S.~Lon{\v{c}}ari{\'{c}}, ``{T}he {P}ast and the {P}resent of the {C}olor
  {C}hecker {D}ataset {M}isuse,'' \emph{arXiv preprint arXiv:1903.04473}, 2019.

\bibitem{gijsenij2009perceptual}
A.~Gijsenij, T.~Gevers, and M.~P. Lucassen, ``Perceptual analysis of distance
  measures for color constancy algorithms,'' \emph{JOSA A}, vol.~26, no.~10,
  pp. 2243--2256, 2009.

\bibitem{finlayson2014reproduction}
G.~D. Finlayson and R.~Zakizadeh, ``Reproduction angular error: An improved
  performance metric for illuminant estimation,'' \emph{perception}, vol. 310,
  no.~1, pp. 1--26, 2014.

\bibitem{hordley2004re}
S.~D. Hordley and G.~D. Finlayson, ``Re-evaluating colour constancy
  algorithms,'' in \emph{Pattern Recognition, 2004. ICPR 2004. Proceedings of
  the 17th International Conference on}, vol.~1.\hskip 1em plus 0.5em minus
  0.4em\relax IEEE, 2004, pp. 76--79.

\bibitem{cube2017online}
\BIBentryALTinterwordspacing
(2017, Nov.) {D}atacolor {S}pyder{C}ube. [Online]. Available:
  \url{http://www.datacolor.com/photography-design/product-overview/spydercube/}
\BIBentrySTDinterwordspacing

\bibitem{qian2018dichromatic}
Y.~Qian, K.~Chen, J.~Nikkanen, J.-K. K{\"a}m{\"a}r{\"a}inen, and J.~Matas,
  ``{D}ichromatic {G}ray {P}ixel for {C}amera-agnostic {C}olor {C}onstancy,''
  \emph{arXiv preprint arXiv:1803.08326}, 2018.

\bibitem{gijsenij2011color}
A.~Gijsenij and T.~Gevers, ``Color constancy using natural image statistics and
  scene semantics,'' \emph{IEEE Transactions on Pattern Analysis and Machine
  Intelligence}, vol.~33, no.~4, pp. 687--698, 2011.

\bibitem{joze2012exemplar}
H.~R.~V. Joze and M.~S. Drew, ``{E}xemplar-{B}ased {C}olour {C}onstancy.'' in
  \emph{British Machine Vision Conference}, 2012, pp. 1--12.

\bibitem{banic2016puma}
N.~Bani{\'c} and S.~Lon{\v{c}}ari{\'c}, ``{P}uma: {A} high-quality
  retinex-based tone mapping operator,'' in \emph{Signal Processing Conference
  (EUSIPCO), 2016 24th European}.\hskip 1em plus 0.5em minus 0.4em\relax IEEE,
  2016, pp. 943--947.

\bibitem{banic2018flash}
------, ``{F}lash and {S}torm: {F}ast and {H}ighly {P}ractical {T}one {M}apping
  based on {N}aka-{R}ushton {E}quation,'' in \emph{International Conference on
  Computer Vision Theory and Applications}, 2018, pp. 47--53.

\bibitem{finlayson2005colour}
G.~D. Finlayson, S.~D. Hordley, and P.~Morovic, ``Colour constancy using the
  chromagenic constraint,'' in \emph{Computer Vision and Pattern Recognition,
  2005. CVPR 2005. IEEE Computer Society Conference on}, vol.~1.\hskip 1em plus
  0.5em minus 0.4em\relax IEEE, 2005, pp. 1079--1086.

\bibitem{fredembach2008bright}
C.~Fredembach and G.~Finlayson, ``Bright chromagenic algorithm for illuminant
  estimation,'' \emph{Journal of Imaging Science and Technology}, vol.~52,
  no.~4, pp. 40\,906--1, 2008.

\bibitem{weber2artikel}
E.~Weber, ``{D}er {T}astsinn und das {G}emeingef{\"u}hl,''
  \emph{Handw{\"o}rterbuch der Physiologie}, vol.~3, no.~2, pp. 481--588, 1846.

\bibitem{zakizadeh2015hybrid}
R.~Zakizadeh, M.~S. Brown, and G.~D. Finlayson, ``{A} {H}ybrid {S}trategy {F}or
  {I}lluminant {E}stimation {T}argeting {H}ard {I}mages,'' in \emph{Proceedings
  of the IEEE International Conference on Computer Vision Workshops}, 2015, pp.
  16--23.

\end{thebibliography}

\end{document}